\documentclass[pdflatex,sn-mathphys-ay]{sn-jnl}% Math and Physical Sciences Author Year Reference Style
\usepackage{hyperref}
\usepackage{graphicx}%
\usepackage{amsmath,amssymb,amsfonts}%
\usepackage{epsfig}
\usepackage{subfigure}
\usepackage{amsthm}%
\usepackage[title]{appendix}%
\usepackage{xcolor}%
\usepackage{colortbl}
\usepackage{textcomp}%
\usepackage{manyfoot}%
\usepackage{textcomp,booktabs}
\usepackage{algorithm}
\usepackage{algorithmic}
\usepackage{listings}%
\usepackage[version=4]{mhchem}\usepackage{arydshln}
\usepackage{enumerate}
\usepackage{mathtools}
\usepackage[switch]{lineno}
\usepackage{stfloats}
\usepackage{float}
\usepackage{bm}
\graphicspath{{Figures/}}

\theoremstyle{thmstyleone}%
\theoremstyle{thmstyletwo}%
\newtheorem{remark}{Remark}%

\theoremstyle{thmstylethree}%

\begin{document}

\title[Mitigating intermittent faults in robot swarms]{Proactive--reactive detection and mitigation of intermittent faults in robot swarms}

\author[1,2]{\fnm{Sinan} \sur{O{\u g}uz}}\email{sinan.oguz@ulb.be}

\author[2]{\fnm{Emanuele} \sur{Garone}}\email{emanuele.garone@ulb.be}

\author[1]{\fnm{Marco} \sur{Dorigo}}\email{mdorigo@ulb.ac.be}

\author*[1]{\fnm{Mary Katherine} \sur{Heinrich}}\email{mary.katherine.heinrich@ulb.be}

\affil*[1]{\orgdiv{%Institut de Recherches Interdisciplinaires et de D\'{e}veloppements en Intelligence Artificielle (IRIDIA)
IRIDIA}, \orgname{Universit\'{e} Libre de Bruxelles}, \orgaddress{\city{Brussels}, \country{Belgium}}}

\affil[2]{\orgdiv{%Unit\'{e} d'enseignement en Automatique et Analyse des Syst\'{e}mes (SAAS)
SAAS}, \orgname{Universit\'{e} Libre de Bruxelles}, \orgaddress{\city{Brussels}, \country{Belgium}}}

\abstract{Intermittent faults are transient errors that sporadically appear and disappear. Although intermittent faults pose substantial challenges to reliability and coordination, existing studies of fault tolerance in robot swarms focus instead on permanent faults. To address intermittent faults in robot swarms that have persistent networks, we propose a novel proactive–reactive strategy to detection and mitigation, based on self-organized backup layers and distributed consensus in a multiplex network. Proactively, robots self-organize dynamic backup paths before faults occur, adapting to changes in the primary network topology and the robots' relative positions. Reactively, robots use one-shot likelihood ratio tests to compare information received along different paths in the multiplex network, enabling early fault detection. Upon detection, communication is temporarily rerouted in a self-organized way, until the detected fault resolves. We validate the approach in representative formation control scenarios, demonstrating that intermittent faults are prevented from disrupting convergence to desired formations, with high detection accuracy and low rates of false positives.}

%Fault tolerance, formation control, intermittent faults, robot swarm, multi-robot systems, shortest path problem, network routing, multiplex networks, consensus, likelihood ratio test


\maketitle

\section{Introduction}
\label{sec:Introduction}

Reliability in networked systems requires consistently accurate information exchange among components, often under dynamic and uncertain conditions~\citep{kirst2016dynamic, zavlanos2011graph}. If communication links fail or become unreliable during multi-hop communication, system convergence and performance guarantees can be compromised~\citep{cortes2008distributed, de2006decentralized, moreau2005stability, olfati2007consensus}. In self-organized robot swarms, this challenge is exacerbated by asynchronous ad-hoc communication and decentralized coordination of actuation and decision making. Robots in a self-organized swarm rely solely on local information and communication with nearby robots, without any estimation of the global state of the swarm or its environment, often leading to prolonged convergence times and vulnerability to the spread of incorrect information~\citep{valentini2015efficient}. Frequent communication between robots can cause faulty information to spread quickly and potentially degrade overall swarm performance or lead to permanent failures.

Self-organized robot swarms exhibit some inherent fault tolerance, through redundancy and a lack of single points of failure~\citep{DorBirBra2014:sch-sr,hamann2018swarm}. 
However, many fault types are not mitigated by this passive tolerance and instead require dedicated mechanisms for detection and mitigation~\citep{bjerknes2013fault,tarapore2017generic,dorigo2021swarm,heinrich2022swarm}. Somewhat counter-intuitively, self-organized robot swarms are inherently much more tolerant to complete robot failures than to partial ones~\citep{winfield2006safety}. For example, a single robot producing faulty or malicious information has been shown to be capable of severe disruption to overall swarm behavior~\citep{winfield2006safety, strobel2018managing}. Faulty robots can also physically obstruct the rest of the swarm, and this interference can paradoxically be worsened by the redundancy that provides swarms with some types of inherent fault tolerance~\citep{pini2011task}.  

Other faults to which self-organized robot swarms are vulnerable and which require dedicated mechanisms for detection and mitigation are \textit{intermittent faults} (IFs). IFs are temporary faults that can appear, disappear, and reappear~\citep{breitfelder2000ieee}, potentially caused by communication interference, sensor malfunctions, or software bugs~\citep{zhou2019review}. IFs are difficult to detect and diagnose due to their transience~\citep{niu2021distributed} and can cause significant disruptions without leaving an easily detectable trace~\citep{zhou2019review}. 
A representative example involves intermittent GPS signal degradation in cluttered environments, which can induce sporadic localization errors. These errors propagate through decentralized state estimation protocols, gradually undermining coordination mechanisms without generating explicit failure indicators.
In real applications, e.g., in robot swarms deployed in inaccessible or dangerous environments~\citep{dorigo2020reflections,dorigo2021swarm}, the consequences of IFs to mission performance and to safety can be severe and in some cases could be irreversible. Detecting and resolving IFs before they escalate is key to minimizing disruption: early detection can prevent cascading failures leading to erroneous execution of tasks and can prevent culmination in permanent failures, either of individual robots or the swarm as a whole~\citep{o2023predictive}.

IFs are difficult to detect in robot swarms with fully self-organized ad-hoc networks, because the network topology is transient and often unpredictable. IFs are much more straightforward to detect in fully centralized systems and in networks with static structures, for example in sensor networks~\citep{sheng2021intermittent, zhang2021intermittent,syed2016novel}. However, for multi-robot systems, full centralization and fully static networks also present downsides, such as single points of failure and limited scalability. 

The recently introduced {\it self-organizing nervous systems (SoNS)}~\citep{zhu2024self} approach combines aspects of centralization and decentralization through self-organized hierarchy. Using the SoNS approach, robot swarms are coordinated via temporary logical networks that are hierarchical and culminate in a dynamic ``brain'' robot (i.e., leader), but which are not imposed from the outside, being instead established and maintained in a self-organized manner. This provides robot swarms with persistent and predictable network structures that are more amenable to detecting IFs, without introducing any single points of failure.
In short, the SoNS approach enables, for the first time, the application of
centralized fault detection and mitigation strategies to robot swarms without sacrificing their oft-cited benefits of scalability, flexibility, and a lack of single points of failure.

\subsection{Related work}
\label{sec:Related_work}

Swarm robotics usually studies passive tolerance to \textit{permanent faults}~\citep{IEC192-04-04}---that is, faults such as electromechanical failures that will remain unless they are actively repaired.
When relying on passive fault tolerance, studies have usually demonstrated that a swarm continues its mission after some or many robots have failed, either by continuing with fewer robots  ~\citep{bjerknes2013fault,rubenstein2014programmable,khadidos2015exogenous} or by replacing/repairing the failed robots without pausing the mission~\citep{christensen2009fireflies,  mathews2017mergeable, varadharajan2020swarm, zhu2024self}.

Swarm robotics studies that focus specifically on fault tolerance do not typically rely on passive tolerance, instead developing dedicated mechanisms to handle permanent faults.
The majority of these methods detect and react to permanent electromechanical failures after they have occurred~\citep{o2023predictive,millard2016exogenous}, often relying on time-out mechanisms in which a robot is considered non-operational if it does not respond to a message within a certain time. Existing methods for detecting permanent faults include LED synchronization~\citep{christensen2009fireflies}, simulation comparison~\citep{millard2014towards}, shared sensor data analysis~\citep{khadidos2015exogenous}, and behavioral feature vectors (BFVs)~\citep{tarapore2017generic}. 
These methods often focus on detection, assuming that once a fault is detected, a repair or other intervention is possible during normal operation (e.g.,~\citep{oladiran2019fault, christensen2009fireflies,o2023predictive}). Although such repairs might be unrealistic in inaccessible, hazardous, or congested environments~\citep{tarapore2019fault,o2023predictive,dorigo2021swarm}, future methods for autonomous repair could be developed to complement detection. In short, the existing reactive methods can be considered effective for many types of permanent faults~\citep{bjerknes2013fault}. 
However, the above-mentioned detection approaches are unlikely to be applicable to the transience of IFs and their long response times~\citep{khaldi2017monitoring} would likely be too slow for the early detection and recovery that IFs require. Methods to detect and repair IFs in robot swarms still need to be developed.

To the best of our knowledge, there are no existing swarm robotics methods focused on IF detection and recovery. Strategies developed for IFs in other types of systems, such as model-based analysis (e.g., discrete-event-system models~\citep{carvalho2017diagnosability}, causal models~\citep{abdelwahed2008practical}) and quantitative analysis (e.g., parameter estimation~\citep{zhang2020intermittent}, geometric approaches~\citep{yan2018detection}, Kalman-like filtering~\citep{zhang2018event}), provide valuable insights but primarily target single-unit systems with static and known system models~\citep{syed2016novel, yaramasu2015aircraft}, which is incompatible with self-organized systems such as robot swarms. Likewise, IF strategies developed for sensor networks~\citep{sheng2021intermittent, zhang2021intermittent} typically use fully centralized architectures to correct information transmission and reception~\citep{syed2016novel}, and are therefore incompatible with self-organized systems.

Furthermore, although fully centralized monitoring is highly effective for detecting and correcting IFs, it can present problems of inflexibility, limited scalability, and single points of failure (e.g., at the point where monitoring is centralized). Fully self-organized approaches, by contrast, would be highly flexible and offer greater scalability and a lack of single points of failure, but would present problems of limited accuracy and potentially slow reaction times.
In this paper, we aim to combine elements of each system type to get the benefits of both. Using our proposed \textit{proactive--reactive} approach, robots can monitor each other using self-organizing hierarchy, detecting IFs accurately and remedying them proactively. 

To demonstrate our proposed \textit{proactive--reactive} approach, we build upon the SoNS approach to self-organizing hierarchy in a robot swarm, which was shown to incorporate temporarily centralized structures into an otherwise self-organized robot swarm without introducing single points of failure or inherently limiting scalability~\citep{zhu2024self,mathews2017mergeable,zhu2020formation,jamshidpey2023reducing,jamshidpey2020multi,jam2025sweep}. We also build upon recent theoretical foundations for self-organizing hierarchical frameworks: hierarchical Henneberg construction (HHC)~\citep{zhang2023self}, which preserves rigidity and hierarchy using only local information. HHC was demonstrated~\citep{zhang2023self} for key self-reconfiguration problems (framework merging, robot departure, and framework splitting) and the mathematical conditions of those problems were derived. 

In the remainder of this paper, we assume all graphs are constructed using existing HHC algorithms~\citep{zhang2023self}, and refer to such graphs as {\it HHC-constructed} graphs. See Appendix~\ref{App:HHC} for details on how we assume HHC and SoNS to be related when used together in one system.

\subsection{Approach and contributions}
\label{sec:Approach_and_contributions}

In fault tolerance for multi-robot systems, both proactive and reactive mechanisms are important~\citep{ghedini2017toward}.
In this paper, we propose a novel \textit{proactive--reactive} method to detect and mitigate IFs in robot swarms.
In the proposed \textit{proactive--reactive} method, the robots first use distributed consensus to preemptively self-organize dynamic backup communication paths before IFs are detected. Then, the robots compare information received via primary and backup paths to detect IFs, using a one-shot likelihood ratio test. When IFs are detected, the robots react by rerouting communication through the dynamic backup paths. In this paper, we apply the proposed \textit{proactive--reactive} method to a scenario of intermittently faulty relative positional information within multi-robot formations that have a hierarchical structure towards a fault-free leader, and demonstrate that the method mitigates IFs and robots are able to continue with the desired formations.

The main technical contributions of this paper can be summarized as follows:
\begin{enumerate}
    \item We address a current gap in robot swarm networking, specifically how to establish back-up communication paths for leader--follower formation control in a self-organized robot swarm. We address this gap by extending the biased minimum consensus (BMC)~\citep{zhang2017distributed} protocol for shortest path planning in static graphs. We introduce the \textit{adaptive biased minimum consensus} (ABMC) protocol for dynamic graphs---addressing time-varying topologies, node neighborhoods, and costs. We demonstrate that our ABMC protocol addresses the minimum-cost path problem, with two objectives integrated into a single cost function: to minimize the number of hops to the destination (the leader robot) and to minimize the degree of network congestion (by minimizing the occurrence of parallel edges).We provide the mathematical properties and stability analysis of the ABMC protocol as a distributed consensus mechanism in dynamic graphs with piecewise constancy, including providing the necessary and sufficient conditions to uniquely determine an equilibrium point representing a minimum-cost backup path.
    \item We address a current gap in robot swarm fault tolerance, specifically tolerance against intermittent faults (IFs). We address this gap by proposing a novel \textit{proactive--reactive} fault-tolerance strategy for detection and mitigation of IFs in robot swarms. Our proposed strategy uses the ABMC protocol to construct backup network layers and combines it with a distributed likelihood ratio (LR) protocol to dynamically reroute traffic in the constructed multiplex network. We propose the mathematical conditions and design the distributed algorithms for backup layer construction and for execution of the \textit{proactive--reactive} strategy for IF detection and mitigation. We also provide the time and space complexity and efficiency properties of both distributed algorithms. Finally, we demonstrate the \textit{proactive--reactive} fault-tolerance strategy in formations of 20 robots with moving leaders.
\end{enumerate}

The rest of the paper is organized as follows. In Sec.~\ref{sec:Preliminaries}, the foundational concepts regarding hierarchical frameworks are presented, along with the existing BMC protocol. In Sec.~\ref{sec:Problem_statement}, we formulate three key problems addressed in this paper: construction of dynamic minimum-cost backup paths, detection of IFs using the constructed backup paths, and mitigation of the detected IFs using the constructed backup paths. The first problem is addressed in Secs.~\ref{sec:Backup_paths} and~\ref{sec:Backup_layers}, and the second and third problem are addressed in Sec.~VI. Finally, in Sec.~VII we validate our contributions in experiments of representative scenarios. The conclusions are summarized in Sec. VIII.

\section{Preliminaries}
\label{sec:Preliminaries}

In this section, we introduce notations and graph-theoretic definitions used in the paper, as well as the biased minimum consensus (BMC) protocol for distributed path planning in static undirected networks~\citep{zhang2017distributed}.

\subsection{Directed graphs and hierarchical frameworks}
\label{sec:Graph_theory_and_hierarchical_swarm_architecture}

\textbf{Notation:} Consider a swarm of \( n \ge 2 \) robots operating in the \( d \)-dimensional Euclidean space \(\mathbb{R}^d\) (with \( d = 2 \) or \( 3 \)). {The robots are capable of establishing directed logical connections. 
The resulting graph \( \mathcal{G} = \left( \mathcal{V}, \mathcal{E} \right) \) is a rooted directed graph where the vertex set \(\mathcal{V} = \{ v_1, v_2, \ldots, v_n \}\) represents the robots and the edge set \( \mathcal{E} = \{ e_{ij} = (v_i,~v_j) \mid v_i, v_j \in \mathcal{V},\, v_i \neq v_j \} \) models the logical connections between them, where an edge \( e_{ij} \) indicates that the child $v_i$ can receive information from the parent $v_j$.}

\begin{figure}[t]
\centering
\subfigure[]{
\includegraphics[width=0.19\textwidth]{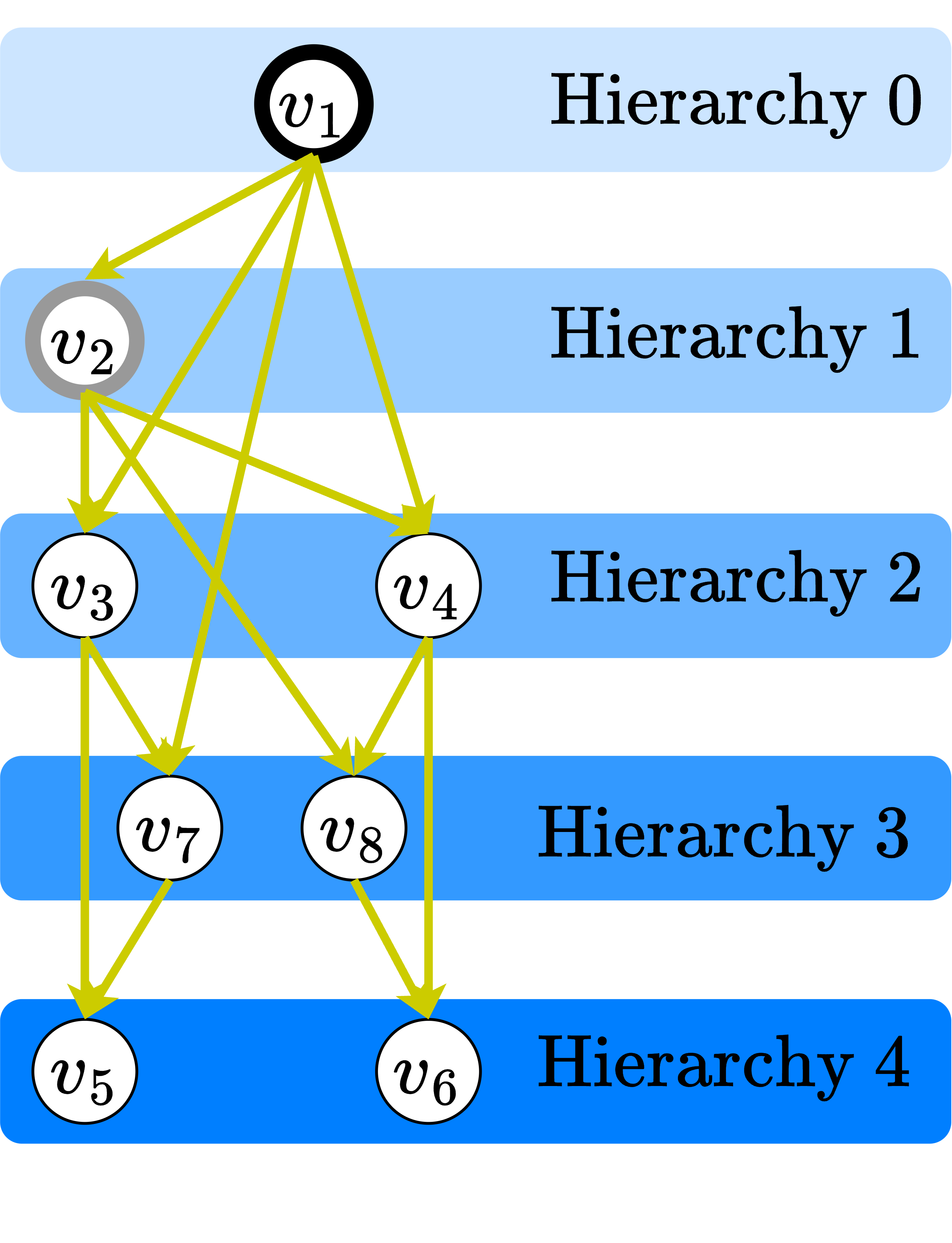}
}
\subfigure[]{
\includegraphics[width=0.17\textwidth]{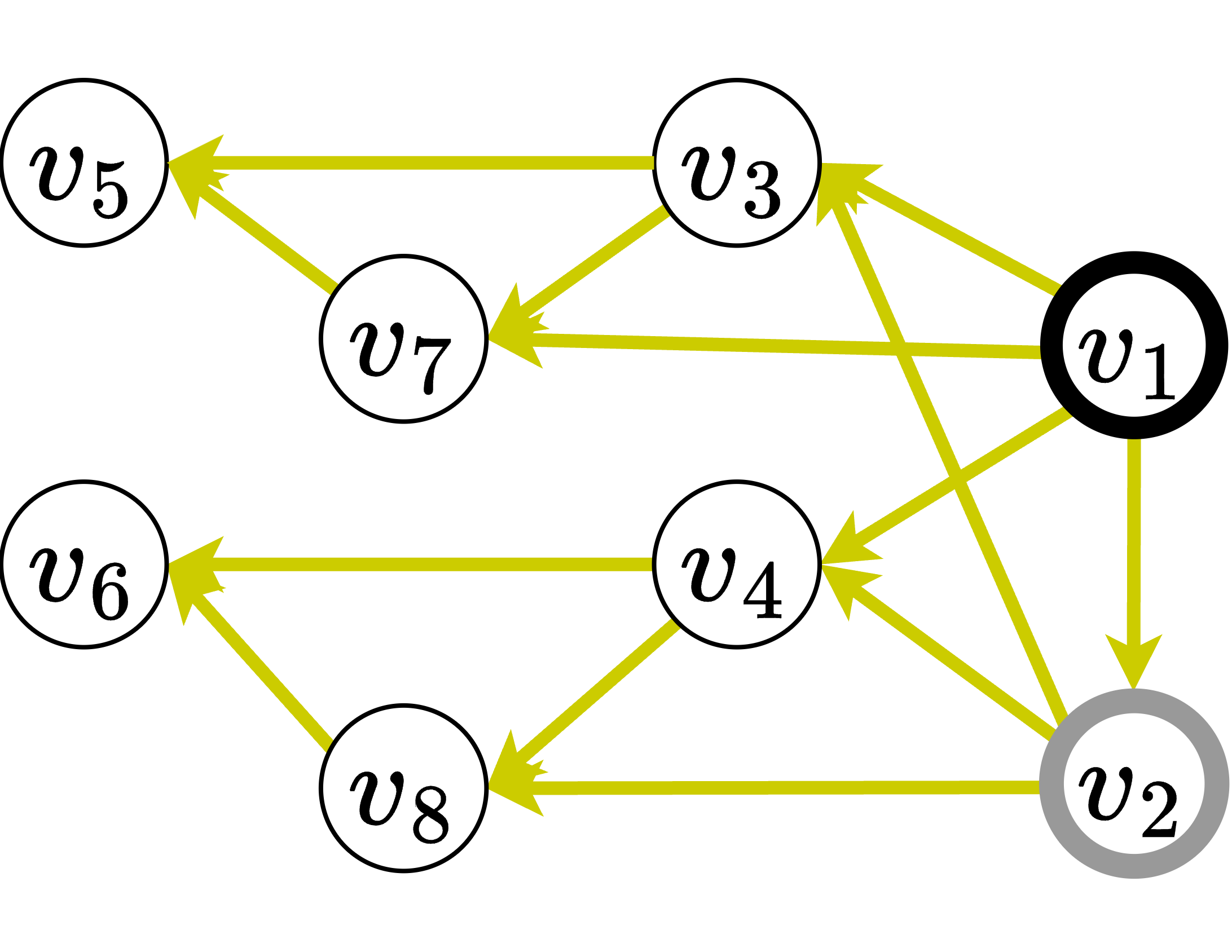}
}
\subfigure[]{
\includegraphics[width=0.10\textwidth]{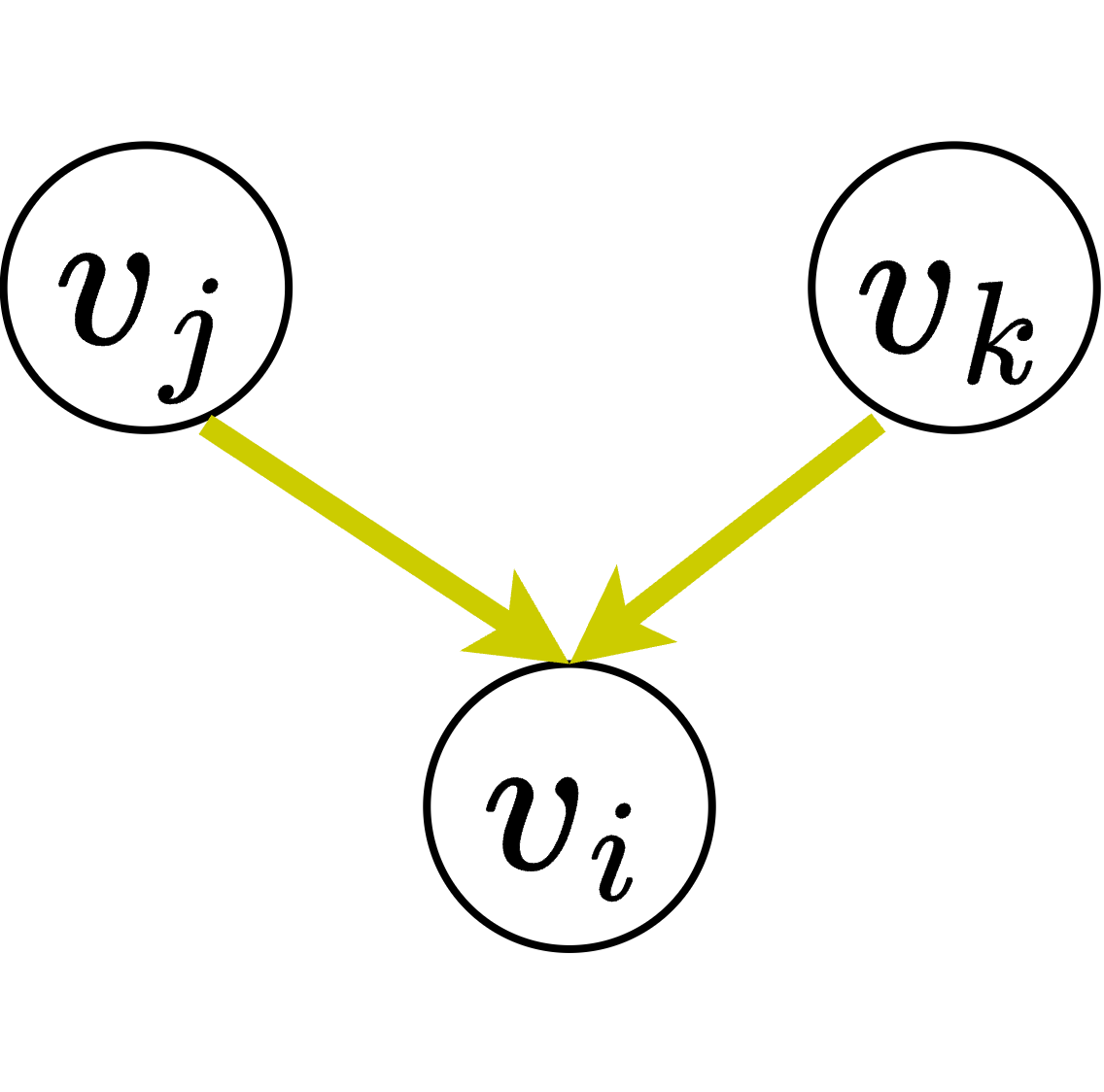}
}
\caption{Adapted from \cite{zhang2023self}. Illustration of an example HHC-constructed graph for a group of eight robots. Robots $v_1$ and $v_2$ are the leader and the first follower, respectively. (a) Directed graph $\mathcal{G}$, with green arrows indicating the directed connections (b) The robots in an example position configuration, $\mathbf{Q}(t)$. (c) The minimal structure of an HHC-constructed graph, where $v_j$ and $v_k$ are parent nodes and $v_i$ is the child node.} 
\label{fig:Example_HC}
\end{figure}

Because the graph \(\mathcal{G}\) is HHC-constructed~\citep[see][and Fig.~\ref{fig:Example_HC}]{zhang2023self}, it is constructed incrementally from a two-robot kernel. The two starting robots are designated as the {\it leader} \(v_1\) (the root) and the {\it first follower} \(v_2\) (a unique child of the leader). This designation is not preassigned; any two robots can assume \(v_1\) and \(v_2\). Every other robot added to the graph is a {\it follower} with exactly two parents.
For each robot \(v_i\) in $\mathcal{G}$, we define the following:
\begin{itemize}
    \item parent set $\mathcal{P}_i$,
    \item neighbor set $\mathcal{N}_i$ (i.e., both the parents and children),
    \item in-degree \( \delta_i^{\mathrm{in}}\),
    \item out-degree $\delta_i^{\mathrm{out}}$, and
    \item {\it hierarchy} level $\mathcal{H}_i$, which is calculated as the hop count of the longest directed path from \( v_i \) to $v_1$.
\end{itemize}

The graph \(\mathcal{G}\) is paired with the position configuration \(\mathbf{Q}\), which describes the physical formation of the robots using pairwise relative positions. The configuration at time \( t \) is given by \( \mathbf{Q}(t) \). For each robot pair \((v_i, v_j)\) in the graph, we denote the true relative position of robot \( v_i \) with respect to \( v_j \) by \(\mathbf{q}_{ij}\).
The true relative position \(\mathbf{q}_{ij}\) is supposed to be transmitted from robot \( v_j \) to \( v_i \), however, the information might be corrupted by, e.g., sensor faults, communication errors, or malicious interference. 
Therefore, the (potentially compromised) relative position that is actually received by robot \( v_i \) is denoted as \(\mathbf{\tilde{q}}_{ij}\).

\subsection{The biased minimum consensus (BMC) protocol}
\label{sec:The_biased_minimum_consensus_protocol}

The biased minimum consensus (BMC) protocol~\citep{zhang2017distributed} is a distributed mechanism designed for static undirected graphs in which nodes update local states through interactions with neighbors, for the purpose of constructing paths to destination nodes. Each node $v_i$ maintains a scalar state $s_i(t) \in \mathbb{R}$: its current estimate of the quantity of interest (such as Euclidean distance). By relying only on information exchanged with neighbors (i.e., nodes directly connected to $v_i$ by an edge in graph $G$), consensus on a path can be reached without requiring centralized control or monitoring.

The BMC protocol operates by first partitioning the node set $\mathcal{V}$ into two: $\mathcal{V}_1$, a set of destination nodes for paths, and the remainder set $\mathcal{V}_2 = \mathcal{V} \setminus \mathcal{V}_1$. All nodes $v_i$ in $\mathcal{V}_2$ iteratively update their scalar state $s_i(t)$ to seek the minimum cost available through their respective neighbor sets \( \mathcal{N}_i\) and thus collectively construct minimum-cost paths to destinations in $\mathcal{V}_1$. Each node $v_i$ assesses costs according to weights $a_{ij} = a_{ji}>0$ biasing its edges $e_{ij}$. 
The dynamics of this process are governed by:
\begin{equation}
    \eta \dot{s}_i(t) = 
    \begin{cases}
    0, & v_i \in \mathcal{V}_1 \\
    -s_i(t) + \underset{v_j \in {\mathcal{N}_i}}{\min}\{s_j(t) + a_{ij}\}, & v_i \in \mathcal{V}_2
    \end{cases}
    \label{eq:BMC}
\end{equation}
where parameter $\eta > 0$ is a rate factor that influences the speed of convergence.

The states \( s_i(t) \) gradually converge to a shared steady-state value $s^*_i$. In this equilibrium state, {destination nodes in $\mathcal{V}_1$ retain their initial states, while the remaining nodes in $\mathcal{V}_2$} settle on the minimum biased state value among their neighbors, as follows~\citep{zhang2017distributed}:
\begin{equation}
s_i^*(t) = \begin{cases} s_i(0), & v_i \in {\cal V}_1 \\ 
\underset{v_j \in {\cal N}_i}{\min}\{s_j(t) + a_{ij}\}, & v_i \in {\cal V}_2 \end{cases}
\label{eq:BMC_equil}
\end{equation}

{
When applying the BMC protocol to the shortest path problem based on Euclidean distance~\citep{zhang2017distributed, mo2021global},} a node's state \( s_i(t) \) can be interpreted as the distance from node $v_i$ to a destination, and the bias term \( a_{ij} \) as the distance from $v_i$ to $v_j$. Through iterative updates, guided by Bellman's optimality principle, the protocol can establish shortest-distance paths from any source node to the given destination node(s)~\citep{zhang2017distributed}.

\section{Problem statement}
\label{sec:Problem_statement}

In a self-organizing hierarchical swarm~\citep{zhu2024self}, a robot occupying the leader position uses information accumulated from its downstream robots to make decisions and then issues instructions to its downstream robots. However, the leader interacts directly only with its direct children, using multi-hop communication to interact with the rest of the robots in the swarm. Therefore, maintaining reliable multi-hop communication paths between the leader and all other robots in the swarm is crucial for effective coordination. These communication paths can become disrupted or inefficient when intermittent faults (IFs) are present. The objective of this paper is to develop a proactive-reactive fault tolerance mechanism to mitigate the effect of IFs on a swarm's ability to maintain accurate positional information for performing formation tasks, in a robot swarm with a self-organizing hierarchical architecture~\citep[HHC-constructed,][]{zhang2023self}.
The following three questions will be addressed:

\begin{enumerate}
\item Given a swarm of $n$ robots in an HHC-constructed formation, how can the robots collectively self-organize dynamic minimum-cost backup paths to the leader that maintain the hierarchy conditions of the original graph and also adapt to its reconfigurations, using only local information from nearby robots?
\item Given the constructed backup paths, how can the robots use them to detect the presence of IFs in their original paths to the leader?
\item Given some detected IFs, how can the robots use the constructed backup paths to mitigate the effect of those IFs while present, and to switch back to their original paths to the leader once the respective IFs have stabilized?
\end{enumerate}

\section{Adaptive minimum-cost backup paths} 
\label{sec:Backup_paths}

In a scenario in which some robots in a swarm are subject to IFs, a robot $v_i$ can circumvent faulty information being transmitted by an intermediary robot (i.e., one lying between it and the leader $v_1$) by constructing a new ``backup" path to $v_1$ that circumvents the faulty robot.
This section presents our distributed method to construct backup paths that adapt to dynamic networks, by allowing follower robots to independently determine their own upstream connections. 

Using locally available information, $v_i$ selects a robot from among those in its communication range to become its preferred backup parent (that is, its backup next hop towards the leader $v_1$). As follower robots in a swarm repeatedly update their preferred backup parents at each step, the resulting chains of distributed parent choices form a {\it minimum-cost} backup path \(\mathcal{B}_i = \{v_i, \dots, v_1\}\) for each follower robot $v_i$. 

\subsection{Adaptive biased minimum consensus protocol (ABMC)}

To address the minimum-cost path problem in rooted directed graphs, we propose our adaptive biased minimum consensus (ABMC) protocol. 
Because the scenario we consider is that of communication paths among networked robots, we aim to construct paths that are both 1) efficient (minimizing the number of hops to the root) and 2) maximally disjoint (minimizing communication congestion and bottlenecks from different paths sharing common edges and vertices), using a single composite cost.
For this aim, our ABMC protocol leverages the structure of the graph by restricting next-hop candidates to upstream nodes and by considering the outdegree $\delta^{\mathrm{out}}$ of the robots.

The ABMC protocol extends the BMC protocol by introducing a dynamic neighbor set and dynamic bias term.
The original BMC was designed for selecting pre-existing edges from a static network based on static biases. 
The ABMC, by contrast, is designed to construct new edges that might not be present in the original network, based on the dynamic topology of the original network, the dynamic positions of the robots, and the dynamic biases associated to potential new paths.
The ABMC is designed as follows: 
\begin{equation} 
\eta \dot{s}_i(t)= 
\begin{cases}
0, & v_i \in {\cal V}_1 \\
-s_i(t) + \underset{v_j \in {\cal P}_i^{\mathrm{cand}}(t)}{\min} \left\{ s_j(t) + a_{ij}(t) \right\}, & v_i \in {\cal V}_2
\end{cases}
\label{eq:ABMC}
\end{equation}
where, \(s_i(t)\) is the state value representing the estimated number of hops at time \(t\) from robot \(v_i\) to the leader, {\({\cal P}_i^{\mathrm{cand}}(t)\) is the candidate parent set} of robot \(v_i\) at time \(t\), \(a_{ij}(t)\) is the bias against selecting robot \(v_j\) as the next hop for communication at time \(t\), and $a_{ij} \neq a_{ji}$. The parameter \(\eta\) is the convergence rate factor, determining how quickly the hop count estimates converge to the final values. \({\cal V}_1\) is the set of the leader and first follower, 
for which the hop count estimate remains constant, and \({\cal V}_2\) is the set of all other robots in the network. 

The candidate parent set ${\cal P}_i^{\mathrm{cand}}(t)$ in Eq.~\eqref{eq:ABMC} is a departure from the neighbor set $N_i$ in the existing BMC in Eq.~\eqref{eq:BMC}, because it is dynamic, includes nodes that are not connected to $v_i$ in the original graph, and leverages graph directionality towards the destination. The candidate parent set ${\cal P}_i^{\mathrm{cand}}(t)$ includes any node $v_j$ that meets the following criteria at time $t$:
\begin{enumerate}
  \item {\bf In-range:} $v_j$ is within communication range $r$ of $v_i$.
  \item {\bf Non-adjacent:} \(v_i\) and \(v_j\) are not connected by an edge in the primary network $G$).
  \item {\bf Leader-proximate:} $v_j$ is fewer hops than $v_i$ from the leader $v_1$ (i.e., ${\cal H}_j<{\cal H}_i$).
\end{enumerate}
Formally, these criteria are given by:
\begin{equation}
\label{eq:ABMC_neighbor}
\begin{split}
{\cal P}^{\mathrm{cand}}_i(t)
=\Bigl\{&\, v_j \in \mathcal V \setminus \{v_i\}\ \Big|\ (v_i,v_j)\notin{\cal E},\ {\cal H}_j < {\cal H}_i, \\
& |v_i,v_j| \leq r \Bigr\}
\end{split}
\end{equation}
where $|i,j|$ denotes the Euclidean distance between $i$ and $j$, and $r>0$ is the communication range.

In BMC, the bias term is static and typically represents Euclidean distances between fixed positions associated with nodes of the original (static) network. By contrast, in ABMC, the bias term is dynamic and accounts for both hierarchy differences and the potential network congestion at each node. The dynamic bias \(a_{ij}(t)\) in Eq.~\eqref{eq:ABMC} is defined as follows:
\begin{equation}
\label{eq:ABMC_weight}
\begin{split}
a_{ij}(t) = \max\Bigl\{
&\, 1 - \rho\bigl(\mathcal{H}_{i}(t) - \mathcal{H}_{j}(t)\bigr)\\
 &\ + \psi\,\phi\bigl(\delta_{j}^{\mathrm{out}}(t),\,\kappa_{d}\bigr), \gamma
\Bigr\}
\end{split}
\end{equation}
where \(\rho > 0\) is a weighting factor adjusting the influence of hierarchy differences on the bias term, and \(\psi \geq 0\) serves as a penalty weight that scales the impact of node congestion on the dynamic bias. We define the congestion penalty function as
$
  \phi\bigl(\delta_{j}^{\mathrm{out}}(t), ~\kappa_{d}\bigr)
    \;=\;
  \max\left\{0,\;\delta_{j}^{\mathrm{out}}(t) - \kappa_{d}\right\}
$
where \(\delta_{j}^{\mathrm{out}}(t)\) is the outdegree of robot \(v_{j}\) at time \(t\), \(\kappa_{d}\) is the outdegree threshold, and \(\gamma > 0\) is a small positive value ensuring \(a_{ij}(t)\) never falls below a specified minimum cost.

\remark\label{rm:2}{
The parameter \(\rho\) will usually be close to 1, ensuring moderate hierarchy differences do not excessively lower the term \(1 - \rho (\mathcal{H}_{i}(t) - \mathcal{H}_{j}(t))\). This prevents the bias term from frequently saturating at its lower bound, allowing it to remain responsive to meaningful hierarchical variations without overly penalizing larger hierarchy gaps.}

\remark\label{rm:3}{
The parameter \(\psi\) is used to design how strongly the penalty term discourages node usage once the outdegree \(\delta_{j}^{\mathrm{out}}(t)\) surpasses a threshold \(\kappa_{d}\). A sufficiently large value of \(\psi\) can simulate a hard constraint, virtually eliminating paths through overloaded nodes, or a moderate \(\psi\) can permit a balance between hop minimization and congestion management. Simultaneously, a higher \(\kappa_d\) value increases the number of next-hop candidates, but also increases the likelihood of bottlenecks, while a lower \(\kappa_d\) reduces the search space, thus lowering the likelihood of bottlenecks but also reducing the availability of next-hop candidates.}

\remark\label{rm:4}{
The relationship between hierarchy difference and hop count is non-monotonic because the topology of the primary network \(\mathcal{G}\) determines the hierarchy levels, while the position configuration \(\mathbf{Q}\) determines which robots are in range $r$ for robot $v_i$ and therefore are candidates to be in the set \(\mathcal{P}^{\mathrm{cand}}_i\). In short, larger hierarchy differences do not necessarily correspond to fewer hops in the backup path. }

Hierarchy levels depend on the dynamic reconfiguration of the primary network \(\mathcal{G}\). In practice, a stable topology is often maintained over extended periods, allowing us to model the hierarchy levels as piecewise constant functions over time intervals between reconfiguration events.
Let \(\{t_k\}_{k=0}^{N} \subset [0, T]\) denote the discrete time instants at which reconfiguration events occur, where \(N\) is the total number of events. These instants satisfy \(0 = t_0 < t_1 < \dots < t_N \le T\), with \(T\) representing the total operational time. Between these reconfiguration events, the hierarchy levels remain constant:
\begin{equation}
\mathcal{H}_{i}(t) = \mathcal{H}_{i}(t_k), \quad \forall t \in [t_k, t_{k+1}), \quad \forall v_i \in \mathcal{V}
\label{eq:hierarchy_piecewise_constant}
\end{equation}
This implies that the hierarchy difference \(\mathcal{H}_{i}(t) - \mathcal{H}_{j}(t)\) and, consequently, the bias term \(a_{ji}(t)\) defined in Eq.~\eqref{eq:ABMC_weight}, remain constant within each interval \([t_k, t_{k+1})\):
\begin{equation}
a_{ij}(t) = a_{ij}(t_k), \quad \forall t \in [t_k, t_{k+1}), \quad \forall v_i, v_j \in \mathcal{V}
\label{eq:cost_piecewise_constant}
\end{equation}
When a reconfiguration of \(\mathcal{G}\) occurs, hierarchy levels are recalculated as:
\begin{equation}
\mathcal{H}_{i}(t_k^+) = \underset{v_j \in \mathcal{P}_i}{\max}\{\mathcal{H}_j(t_k^+)\} + 1, \quad \forall v_i \in \mathcal{V}
\label{eq:hierarchy_recalculation}
\end{equation}
where \(t_k^+\) denotes the time immediately after \(t_k\).

{\remark\label{rm:5}{We distinguish between two different categories of topology changes in our scenario: transient disruptions versus desired semi-permanent changes. On one hand, there can be transient interruptions of some edges, for example because of temporary sensor occlusion or communication disruption. These transient topology interruptions because of minor edge disturbances are negligible to the ABMC consensus time scale, because such disturbances are much shorter-lived than the ABMC convergence process. 
On the other hand, there can be semi-permanent reconfigurations of the desired topology, which trigger HHC-reconstruction events~\citep[see][]{zhang2023self} to rebuild the graph and the hierarchy states. After such a reconfiguration event, the ABMC protocol then adapts to the resulting semi-permanent graph, re-converging on new backup paths (until the next reconfiguration event). Accordingly, between two reconfiguration instants \(t_k\) and \(t_{k+1}\), both the hierarchy states and the ABMC candidate parent sets are treated as piecewise constant; the topology is fixed within each interval \([t_k,t_{k+1})\). This permits the use of standard convergence analysis for consensus on each interval. The overall stability of the protocol is preserved provided the dwell time \(t_{k+1}-t_k\) is sufficiently large relative to the convergence rate set by \(\eta\)~\citep{olfati2004consensus,ren2005consensus,moreau2005stability}.
}}

\begin{figure}[t]
\centering
\includegraphics[width=0.65\textwidth]{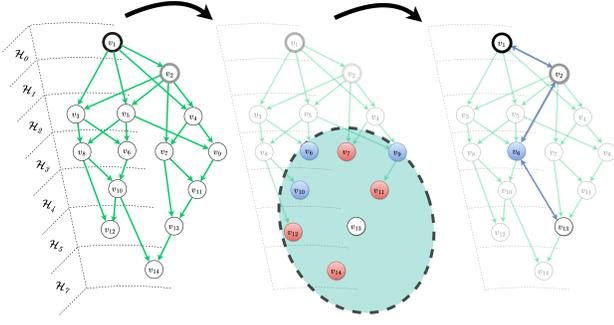}
\caption{Illustration of constructing a minimum-cost backup path using the ABMC protocol, in an example hierarchical swarm of 14 robots with seven hierarchy levels \(\mathcal{H}_0\) to \(\mathcal{H}_6\). The left panel shows the primary network $\mathcal{G}$. The middle panel shows how robot \(v_{13}\) checks the robots in its communication range (dashed circle) and determines its potential next hops (blue) toward the leader \(v_{1}\), by excluding direct parents (\(v_{7}\) and \(v_{11}\); red) and any robots that fail to meet hierarchy requirement (\(v_{12}\) and \(v_{14}\); also red). The right panel depicts the minimum-cost backup path (blue edges) from \(v_{13}\) to \(v_{1}\) that results from each robot in the swarm executing several iterations of the ABMC protocol, in a fully decentralized way.}
\label{fig:Example_SPP}
\end{figure}

The operation of the ABMC protocol on an example HHC-constructed primary network \(\mathcal{G}\) and formation \(\mathbf{Q}\) is illustrated in Fig. \ref{fig:Example_SPP}. The figure shows how a robot identifies potential next hops within its communication range $r$ and creates a backup edge towards the leader. 

\subsection{Mathematical properties and stability analysis of ABMC protocol}

Standard consensus analysis using graph Laplacians relies on symmetric interactions~\citep{bullo2018lectures}.
The ABMC introduces directional bias terms, breaking symmetry and invalidating spectral methods to understand convergence behavior (i.e., methods that rely on the eigenvalues and eigenvectors of the graph Laplacian).  
We therefore provide specialized mathematical foundations and stability analysis for the asymmetric interactions of the ABMC protocol.

Let \(\beta_i(t)\) represent the right-hand side of Eq.~\eqref{eq:ABMC}.
The upper bound of \(\beta_i\), denoted by \(\bar{\beta}\), is defined as the maximum value of \(\beta_i\) across all robots, given by \(\bar{\beta} = \underset{v_i \in {\cal V}}{\max} \{ \beta_i \}\). The set of robots that attain this maximum is denoted by \(\overline{{\cal S}} = \arg \underset{v_i \in {\cal V}}{\max} \{ \beta_i \}\). Similarly, the lower bound \(\underline{\beta}\) and the corresponding set \(\underline{{\cal S}}\) are defined analogously as \(\underline{\beta} = \underset{v_i \in {\cal V}}{\min} \{ \beta_i \}\) and \(\underline{{\cal S}} = \arg \underset{v_i \in {\cal V}}{\min} \{ \beta_i \}\), respectively.

Let ${\cal P}^{\mathrm{back}}_i$ represent the backup parents that $v_i$ selects from among its candidate parents ${\cal P}^{\mathrm{cand}}_i$. ${\cal P}^{\mathrm{back}}_i$ is defined as a subset of ${\cal P}^{\mathrm{cand}}_i$ that minimizes the sum of the state value and the bias term, as follows:
\begin{equation}
{\cal P}^{\mathrm{back}}_i := \arg \min_{v_j \in {\cal P}^{\mathrm{cand}}_i(t)} \left\{ s_j(t) + a_{ij}(t) \right\}
\label{eq:minimazers}
\end{equation}
Note that, because the leader robot $v_1$ originates the positional information in the swarm, its backup parent set is empty, \({\cal P}^{\mathrm{back}}_1 = \emptyset\): it does not require an alternative information path.

We now present a series of four lemmas that establish the properties of the ABMC protocol. The proofs of all four lemmas are reported in Appendix~\ref{App:Proofs}. The first lemma demonstrates that the robots' hop count estimates evolve in a controlled manner. Specifically, it establish that the maximum and minimum values of \(\beta_i(t)\)---which are directly related to the rates of change of the robots' hop count estimates---are monotonically non-increasing and non-decreasing, respectively. This ensures that the updates neither accelerate nor decelerate in an unbounded manner.

\textbf{Lemma 1:}  
The upper bound \(\bar{\beta}(t)=\max_{v_i \in \mathcal{V}} \beta_i(t)\) is monotonically non-increasing, and the lower bound \(\underline{\beta}(t)=\min_{v_i \in \mathcal{V}} \beta_i(t)\) is monotonically non-decreasing.
\smallskip

The second lemma describes the long-term interaction dynamics among the robots. It shows that, over time, the robots influencing the state of those achieving the maximum and minimum rate of change are themselves among the robots achieving the maximum and minimum rate of change, respectively.

\textbf{Lemma 2:}  
As \(t \rightarrow +\infty\), for every robot \(v_i\) that attains the upper bound \(\bar{\beta}(t)\) (i.e., \(v_i \in \overline{\mathcal{S}}\)), its backup parent set satisfies \({\cal P}^{\mathrm{back}}_i \subset \overline{\mathcal{S}}\). Similarly, for every robot \(v_i\) that attains the lower bound \(\underline{\beta}(t)\) (i.e., \(v_i \in \underline{\mathcal{S}}\)), we have \({\cal P}^{\mathrm{back}}_i \subset \underline{\mathcal{S}}\).
\smallskip

The third lemma shows the boundedness of the protocol states. This ensures that the state values of the robots do not increase indefinitely and thus protocol stability is maintained.

\textbf{Lemma 3:} $s_i(t) \leq S_{\max},~ t > 0,~ \forall v_i \in \mathcal{V}_2$.
\smallskip

Finally, the fourth lemma demonstrates that as time approaches infinity, the leader will be included in the set of robots with the minimum state value. By aligning its state with the minimum, the leader follows the same protocol as other robots, promoting network stability and preventing divergence. 

\textbf{Lemma 4:} As $t \rightarrow +\infty$, $\underline{{\cal S}} \cap \mathcal{V}_1\neq \emptyset$, where $\underline{{\cal S}}$.
\smallskip

Having established the properties of the ABMC protocol through lemmas 1--4, we now present two theorems that demonstrate its stability---that is, the ability of the consensus protocol to reliably reach a state of equilibrium~\citep{ren2005survey}. The proofs of both theorems are reported in Appendix~\ref{App:Proofs}.

\textbf{Theorem 1:}
Fix $[t_k,t_{k+1})$ on which $\mathcal{H}$, $\mathcal{N}_i$, and $a_{ij}$ are constant. 
Let $\mathcal{G}=(\mathcal{V},\mathcal{E})$ be an HHC-constructed graph with the leader robot $v_1$ and first follower $v_2$; destination set $\mathcal{V}_1=\{v_1,v_2\}$, remainder set $\mathcal{V}_2=\mathcal{V}\setminus\mathcal{V}_1$. 
For each $v_i\in\mathcal{V}_2$, the candidate parent set $\mathcal{P}^{\mathrm{cand}}_i$ is defined in Eq.~\eqref{eq:ABMC_neighbor} and \emph{reachability} of $\mathcal{V}_1$ via the candidate parents holds: 
$\forall\,v_i\in\mathcal{V}_2\,\exists\,m$ and a sequence $v_{k_0},\dots,v_{k_m}$ with $v_{k_0}=v_i$, $v_{k_{\ell+1}}\in\mathcal{P}^{\mathrm{cand}}_{k_\ell}$, and $v_{k_m}\in\mathcal{V}_1$.
Each robot $v_i$ maintains a scalar state $s_i(t)\in\mathbb{R}$, interpreted as its current estimate of the minimum cumulative bias from $v_i$ to $\mathcal{V}_1$ along converged paths.
Each robot adheres to the ABMC protocol given in Eq.~\eqref{eq:ABMC}. Then on $[t_k,t_{k+1})$,
there exists a unique $s^*\in\mathbb{R}^{|\mathcal{V}|}$ for each $v_i\in\mathcal{V}$ with:
\begin{equation}
s_i^*(t) = \begin{cases}
~~s_i(0), & v_i \in {\cal V}_1 \\
\underset{v_j \in {\cal P}^{\mathrm{cand}}_i(t)}{\min}\left\{s_j^*(t) + a_{ij}(t)\right\}, & v_i \in {\cal V}_2
\end{cases}
\label{eq:ABMC_equil}
\end{equation}
For any $s(0)\in\mathbb{R}^{|\mathcal{V}|}$, the solution $s(t)$ converges globally and asymptotically to $s^*$.
\smallskip

Building upon the global convergence of the ABMC protocol, we now examine the relationship between the equilibrium state and the minimum-cost backup path. The following theorem establishes the equivalence between the equilibrium point of the protocol and the solution to the minimum-cost path problem.

\textbf{Theorem 2:} If the initial state of the leader robot is $s_1(0)=0$, then the equilibrium point of the ABMC protocol serves as a solution to the minimum-cost variant of the shortest‐path problem.

\remark\label{rm:7}{Upon convergence of each robot's state to the solution of the aforementioned nonlinear equation, the backup path from any robot in \({\cal V}_2\) to the leader robot can be constructed by recursively tracing the sequence of backup parent robots.}

\remark\label{rm:8}{Theorems 1 and 2 guarantee that all robots' state values converge to their respective backup paths, regardless of their initial states. }

\section{Backup network layers}
\label{sec:Backup_layers}

\begin{figure}[ht]
\centering
\includegraphics[width=0.45\textwidth]{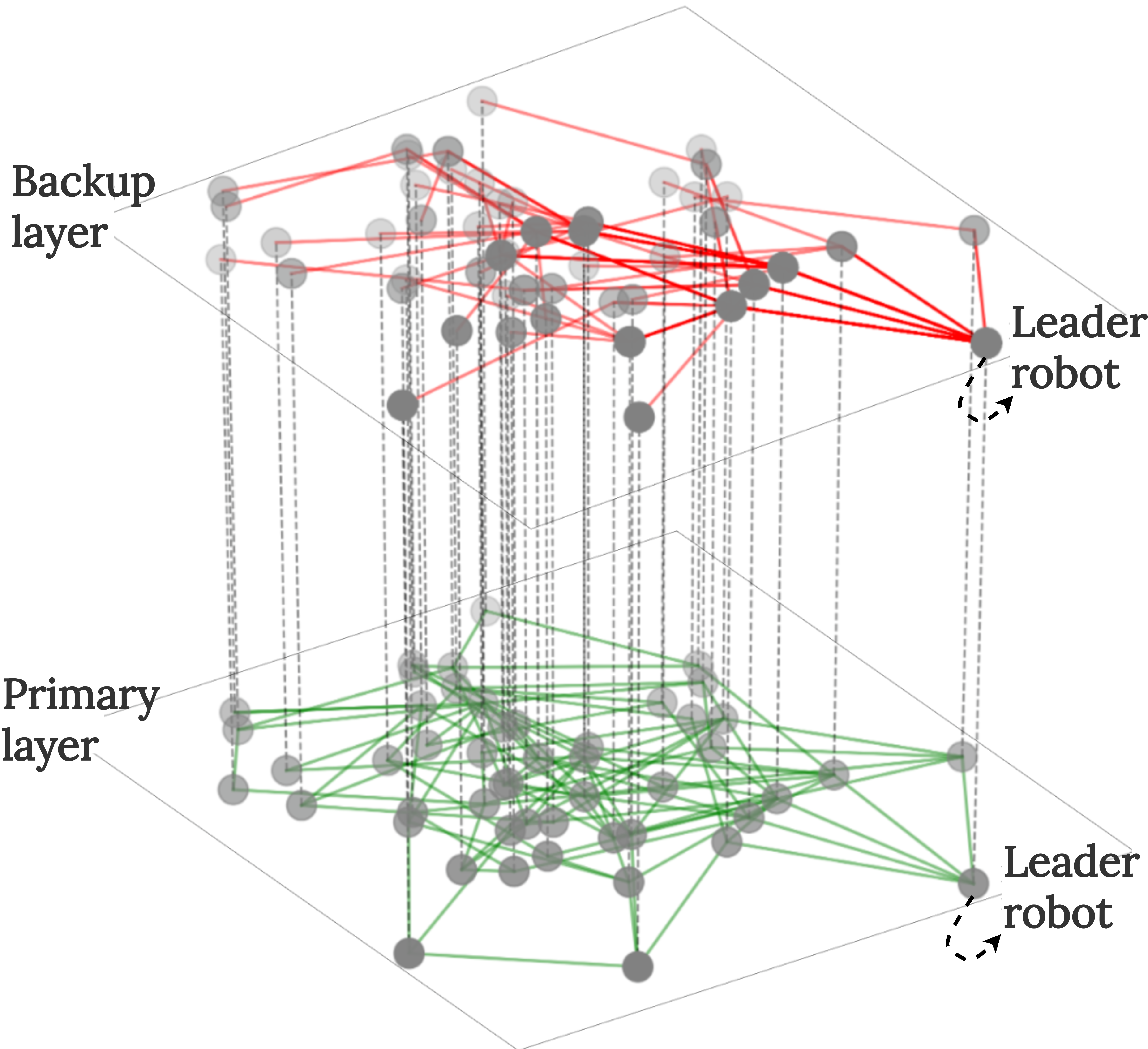}
\caption{Visualization of a backup network layer established by the ABMC protocol for a 50-robot swarm. Each robot (grey circles) independently establishes minimum-cost backup edges (red lines; darker red indicates a higher number of edges between the two respective nodes) that collectively form backup paths to the leader robot. These backup paths collectively form the backup network layer (red) that complements the primary network (green).}
\label{fig:ABMC2}
\end{figure}

Collectively, the backup paths generated by the ABMC protocol form the backup network layer (Fig.~\ref{fig:ABMC2}). The backup network layer allows each robot to receive positional information along multiple paths from the leader, via its primary parents and its backup parents. The backup layer thus provides the structure for a feedback mechanism, enabling independent verification of the accuracy of information flowing along different paths through the swarm. The backup network layer is self-organized using local information from nearby robots and it adaptive to changes in the original graph, thus maintaining the adaptability to current conditions that is crucial for addressing intermittent faults.

\subsection{Algorithm for backup layer construction}
Algorithm~\ref{alg:ABMC-Single} details the process of constructing backup paths for an individual robot, for a robot swarm deployed in  $\mathbb{R}^2$. The paths constructed throughout a swarm are aggregated to form the backup network layer. 

First, for each follower robot \( v_i \in \mathcal{V}_2 \), the algorithm begins by initializing two sets: one for the minimum-cost backup edge (\( {\cal E}^{\text{min}} \)) and one for alternative backup edges (\( {\cal E}^{\text{alt}} \)). Note that while the ABMC protocol is originally formulated in continuous time, see Eq.~\eqref{eq:ABMC}, in practical implementations the differential equation is discretized. Here, the iteration index \( k \) corresponds to discrete time instants \( t = k\Delta t \) for a chosen sampling interval \(\Delta t\), and the update \( s_i[k+1] \) approximates the state \( s_i(t+\Delta t) \). 
{Define the step size \( \mathcal{s} := \Delta t/\eta \in (0,1] \). We use the forward-Euler update $ s_i[k+1] = (1-\mathcal{s})\,s_i[k] + \mathcal{s}\bigl(s_{j^*}[k] + a_{ij^*}[k]\bigr)$, where \(j^*\) indexes the minimum-cost candidate parent at step \(k\). This convex-combination form preserves nonnegativity for hop-like states when \(s_i[0]\ge 0\). }
Consequently, the convergence criterion $\left| s_i[k+1] - s_i[k] \right| < \zeta,$ with a predefined threshold \( \zeta > 0 \), serves as a discrete-time analogue to the continuous-time derivative approaching zero. 

Then, the minimum-cost backup path is reconstructed by backtracking from the candidate parent \( v_{j^*} \) using the edges stored in \( {\cal E}^{\text{min}} \). 
{Here \( v_{j^*} := \underset{v_j \in {\cal P}^{\mathrm{cand}}_i[k]}{\arg\min}\bigl( s_j[k] + a_{ij}[k] \bigr) \) is the minimum-cost candidate parent, and it is thus selected as the backup parent for \(v_i\) at iteration \(k\); equivalently, the directed edge \((v_i, v_{j^*})\) is the link stored for path construction.} 
The algorithm ensures that each robot in the backtracked path is connected to its predecessor via a valid edge from the minimum-cost edge set, \({\cal E}^{\text{min}}\). In the final iteration, for each robot \(v_j \in {\cal P}^{\mathrm{cand}}_i[k]\), if \(v_j \neq v_{j^*}\) and the difference between the alternative candidate's cost and the minimum-cost path's cost is within an acceptable range, i.e., 
\begin{equation}
    \left| \bigl( s_j[k] + a_{ij}[k] \bigr) - s_i[k+1] \right| < \tau
    \label{eq:alt_crt}
\end{equation}
then the edge \((v_i, v_j)\) is added to the alternative edge set.

The threshold \(\tau > 0\) is a predefined parameter that allows slight variations in the path costs, ensuring that an alternative path is considered viable only if its cost is only marginally higher than the minimum-cost path's cost. Subsequently, the alternative paths are obtained by backtracking from each eligible robot \(v_j\) using the edges stored in \({\cal E}^{\text{alt}}\). Upon completion of the backtracking processes, the algorithm stores the resulting minimum-cost backup path \({\cal B}_i^{\text{min}}\) and the set of alternative backup paths \({\cal B}_i^{\text{alt}}\) (along with their associated costs and edge sets) for robot \(v_i\) in the set $\bm{\mathcal{B}}_i$.

\begin{algorithm}
\caption{\small Construction of backup paths for Robot \(v_i\)}
\label{alg:ABMC-Single}
\begin{algorithmic}[1]
\small
\REQUIRE 
\(\eta\): Convergence rate factor, \(\rho\): Weight adjusting the hierarchy’s impact on the bias term, $\psi$: Weight scaling the contribution of congestion penalty to the overall bias, \(\kappa_d\): Maximum robot outdegree, \({\cal H}_{i}\): Robot's hierarchy, \(K\): Maximum iterations, \(\tau\): Alternative path cost threshold, \(\zeta\): Convergence tolerance, \(r\): Communication range, {$\Delta t$: Sampling interval with $0<\Delta t/\eta\le 1$}
\ENSURE 
$\bm{\mathcal{B}}_i$: Set of backup paths (minimum-cost ${\cal B}_i^{\text{min}}$ and alternatives ${\cal B}_i^{\text{alt}}$) for $v_i$.
\STATE \textbf{Initialize:} 
\STATE \quad \({\cal E}^{\text{min}} \leftarrow \emptyset\), \({\cal E}^{\text{alt}} \leftarrow \emptyset\), \({\cal B}_i^{\text{min}} \leftarrow \emptyset\), \({\cal B}_i^{\text{alt}} \leftarrow \emptyset\)
\STATE \quad \(s_i[0] \leftarrow \infty\), \(\text{converged} \leftarrow \text{False}\), \(k \leftarrow 0\)
\WHILE{\(k \leq K\) \textbf{and} not converged}
    \STATE Compute the candidate parent set \({\cal P}^{\mathrm{cand}}_i[k]\)  according to Eq.~\eqref{eq:ABMC_neighbor}
    \FORALL{\(v_j \in {\cal P}^{\mathrm{cand}}_i[k]\)}
        \STATE Compute \(a_{ij}[k]\) using Eq.~\eqref{eq:ABMC_weight}
    \ENDFOR
    \STATE \(v_{j^*} \leftarrow \underset{v_j \in {\cal P}^{\mathrm{cand}}_i[k]}{\arg\min} \Bigl( s_j[k] + a_{ij}[k] \Bigr)\)
    {\STATE $s_i[k+1] \leftarrow (1-\mathcal{s})\,s_i[k] + \mathcal{s}\bigl(s_{j^*}[k] + a_{ij^*}[k]\bigr)$ ~~~$\triangleright$ Euler step; preserves $s_i\ge 0$}
    \STATE Set backup parent: \({\cal P}^{\mathrm{back}}_i \leftarrow v_j^*\)
     \STATE Let edge \(e_{ij^*} \leftarrow (v_i, v_{j^*})\)
    \STATE \({\cal E}^{\text{min}} \leftarrow {\cal E}^{\text{min}} \cup e_{ij^*}\)
    \STATE \(\text{converged} \leftarrow \Bigl(|s_i[k+1]-s_i[k]| < \zeta\Bigr)\)
    \STATE \(k \leftarrow k + 1\)
\ENDWHILE
\IF{\text{converged}}
    \STATE \( {\cal B}_i^{\text{min}} \leftarrow\) Backtrack from \(v_i\) using the edges in \({\cal E}^{\text{min}}\) ~~~$\triangleright$ cf. Remark~\ref{rm:7}
\ENDIF
\FORALL{\(v_j \in {\cal P}^{\mathrm{cand}}_i[k]\) such that \(v_j \neq v_{j^*}\) and \(\bigl| (s_j[k] + a_{ij}[k]) - s_i[k+1] \bigr| < \tau\)}
    \STATE Let edge \(e_{ij} \leftarrow (v_i, v_j)\)
    \STATE \({\cal E}^{\text{alt}} \leftarrow {\cal E}^{alt} \cup e_{ij}\)
    \STATE \( {\cal B}_i^{\text{alt}} \leftarrow\) Backtrack from \(v_i\) using the edges in \({\cal E}^{\text{alt}}\) 
\ENDFOR
\STATE \( \bm{\mathcal{B}}_i \leftarrow \{ {\cal B}_i^{\text{min}},\, {\cal B}_i^{\text{alt}}\}\) ~~~$\triangleright$ {minimum-cost backup paths ${\cal B}_i^{\text{min}}$ and alternatives ${\cal B}_i^{\text{alt}}$}
\RETURN $\bm{\mathcal{B}}_i$
\end{algorithmic}
\end{algorithm}

{\remark\label{rm:9}{The discrete next-hop selection {\mbox{ $v_{j^\star} \in \underset{v_j\in\mathcal P^{\mathrm{cand}}_i[k]}{\arg\min}\bigl(s_j[k]+a_{ij}[k]\bigr)$}} is set-valued under ties or near-ties, which can lead to chattering even when $s_i[k]$ has settled. This can be suppressed using a hysteresis policy, often used to suppress chattering and limit rapid switching~\citep{lin2009stability,hespanha2003hysteresis}: in short, retain the current parent unless a candidate provides a clearly better local cost, or the improvement is consistently observed over several updates.}}

\subsection{Properties of backup layer construction}

{\bf Runtime:} The time complexity of Algorithm~\ref{alg:ABMC-Single} for a single robot \(v_i\) is primarily determined by the main loop, which iterates up to \(K\) times, where \(K\) is a constant representing the maximum number of iterations. Each iteration involves several key steps: candidate parent selection, computation of bias terms and state updates, identification of the minimum-cost parent, and a convergence check.

Robot selection is efficiently handled using a grid-based method~\citep{thrun2002probabilistic}, which runs in \(O(1 + |{\cal P}^{\mathrm{cand}}_i|)\) time. For each node \(v_j \in {\cal P}^{\mathrm{cand}}_i\), the computation of the bias \(a_{ij}[k]\) and the update of the state value \(s_i[k+1]\) are constant-time operations, yielding a total per-iteration cost of \(O(|{\cal P}^{\mathrm{cand}}_i|)\) for these steps. Identifying the minimum-cost parent \(v_j^*\) (i.e., the one that minimizes the cost \(s_j[k] + a_{ij}[k]\)) requires scanning through all candidate parents, which also takes \(O(|{\cal P}^{\mathrm{cand}}_i|)\). The convergence check---determining whether \(|s_i[k+1] - s_i[k]|\) falls below a predefined threshold \(\zeta\)---is an \(O(1)\) operation. Hence, each iteration runs in \(O(1 + |{\cal P}^{\mathrm{cand}}_i|)\) time. Since \(K\) is constant, the cumulative time complexity over the main loop is \(O(K(1+|{\cal P}^{\mathrm{cand}}_i|)) = O(|{\cal P}^{\mathrm{cand}}_i|)\).

Once convergence is achieved, the algorithm backtracks the minimum-cost path from the selected parent \(v_j^*\) using the stored edges \({\cal E}^{\text{min}}\). This backtracking process has a time complexity of \(O(L)\), where \(L\) is the length of the path (i.e., the number of hops) to the leader robot. Additionally, the algorithm backtracks alternative paths from each eligible robot \(v_j \in {\cal P}^{\mathrm{cand}}_i\). For each such robot—if it satisfies the condition in Eq.~\eqref{eq:alt_crt}, an alternative path is backtracked using the stored edges \({\cal E}^{\text{alt}}\). As there may be up to \(|{\cal P}^{\mathrm{cand}}_i|\) alternative paths, each requiring \(O(L)\) time, this step has a complexity of \(O(|{\cal P}^{\mathrm{cand}}_i| L)\).

Combining all steps, the overall time complexity---including the main loop and the backtracking processes---is
$O(1 + |{\cal P}^{\mathrm{cand}}_i| + L + |{\cal P}^{\mathrm{cand}}_i| L)$. Since constant terms and lower-order terms are absorbed by the dominant factor, and given that \(L\) is generally small compared to the total number of robots, the final complexity simplifies to \(O(|{\cal P}^{\mathrm{cand}}_i| L)\).

{\bf Memory space requirements:} In terms of the memory space required for a single robot to run Algorithm~\ref{alg:ABMC-Single}, storing the minimum-cost path \({\cal B}_i^{\text{min}}\) requires \(O(L)\) space, while storing all alternative paths \({\cal B}_i^{\text{alt}}\) requires up to \(O(|{\cal P}^{\mathrm{cand}}_i| L)\) space, since there may be as many as \(|{\cal P}^{\mathrm{cand}}_i|\) alternative paths (each of length \(L\)). Thus, the total space complexity for storing all paths is \(O(|{\cal P}^{\mathrm{cand}}_i| L)\).

{\bf Efficiency:} The efficiency of Algorithm~\ref{alg:ABMC-Single} for a single robot can be influenced by the spatial layout of the robots in the formation. In sparser areas, where the number of robots in in the communication range $r$ of robot $v_i$ is much smaller than the total number of robots in the swarm, the algorithm performs more efficiently in both time and space, making it suitable for real-time applications. In denser formations, where the number of robots in \(|{\cal P}^{\mathrm{cand}}_i|\) increases w.r.t. the total number of robots in the swarm, performance will be more affected by the number of candidate parents.

\section{Proactive--reactive detection and mitigation of intermittent faults}
\label{sec:Proactive-reactive}

Intermittent faults cause transient deviations (biases, offset errors, noise), and thus can perturb the statistical properties of transmitted data, particularly the mean and variance~\citep{muhammed2017analysis}; see Appendix~\ref{APP:Data} for details of the fault and noise model used in this study. In our scenario, in which backup paths to the leader have already been constructed, the detection task can be reduced to checking whether the distribution of primary data differs significantly from that of the backup data.
To check this, we adopt a Likelihood Ratio (LR) test for the data from the primary parents compared to the data from the backup parents, contrasting two hypotheses:
\begin{itemize}
    \item \textbf{Null hypothesis $H_0$}: the data are consistent with the backup (no fault present).
    \item \textbf{Alternative hypothesis $H_1$}: the data distribution is perturbed (fault present).
\end{itemize}
Note that we use the \emph{log}-Likelihood Ratio (LLR) for numerical stability and additivity, which yields the same decisions as the LR because $\log(\cdot)$ is monotone. Each robot computes LLR values for its parents and uses them to decide whether a fault is present and, if so, whether to switch to a backup path.

\subsection{Algorithm for fault detection and mitigation}
\label{sec:Algorithm_for_fault_detection}

\begin{algorithm}[t!]
\caption{\small\textit{Proactive--reactive} fault detection and mitigation}
\label{alg:FDMM}
\begin{algorithmic}[1]
\small
\REQUIRE{
    $\bm{\mathcal{B}}_i$: Set of backup paths for robot $v_i$ to the leader robot (from Algorithm \ref{alg:ABMC-Single}), including:
         ${\cal B}_i^{\text{min}}$: Minimum-cost backup path (a list of (robot, cost) tuples),
         and ${\cal B}_i^{\text{alt}}$: Alternative backup paths (each a list of (robot, cost) tuples);
    $\mathbf{\tilde{Q}}_{ij} \triangleq \{\mathbf{\tilde{q}}_{ij}[k] : k = 1, 2, \dots, N\}$: Potentially faulty data set includes $N$ number $\mathbf{\tilde{q}}_{ij}$ data;
    $\mathbf{Q}_{ij}[b] \triangleq \{\mathbf{q}_{ij}[k] : k = 1, 2, \dots, N\}$: Fault-free data set includes $N$ number $\mathbf{{q}}_{ij}$ data from backup path $\mathcal{B}_i[b] \in \bm{\mathcal{B}}_i$;
    $\text{use\_backup}$: Boolean flag indicating if the backup path is in use;
    $T^{\mathrm{lock}}$: Timer to prevent rapid switching; $T^{\mathrm{dur}}$: A positive real number defining the minimum time between decision changes;
    $\Delta t$: Sampling interval
}
\ENSURE{Fault detection decision for robot $v_i$}

\FORALL{parent robots $v_j \in \mathcal{P}_i$ }
    \STATE $\mathbf{LLR}_{ij} \leftarrow \emptyset$
    \FORALL{backup path \(\mathcal{B}_i[b] \in \bm{\mathcal{B}}_i\)}
        \STATE  \(\boldsymbol{\mu}_{\mathbf{\tilde{Q}}_{ij}} \leftarrow \)Eq.\eqref{eq:sample_means_l1},  \(\boldsymbol{\mu}_{\mathbf{Q}_{ij}[b]} \leftarrow \) Eq.\eqref{eq:sample_means_l2}, 
        \STATE \({\boldsymbol{\Sigma}}_{\mathbf{\tilde{Q}}_{ij}} \leftarrow \) Eq.\eqref{eq:covariance_matrice_parent}, and \({\boldsymbol{\Sigma}}_{\mathbf{{Q}}_{ij}[b]} \leftarrow \) Eq.\eqref{eq:covariance_matrice_backup}
        \STATE  \(\text{LLR}_{ij}[b] \leftarrow \) Eq. \eqref{eq:LLR}
        \STATE \(\mathbf{LLR}_{ij} \leftarrow \mathbf{LLR}_{ij} \cup \{\text{LLR}_{ij}[b]\}\)
    \ENDFOR
    \STATE  \(\mathcal{Q}_1 \leftarrow\)  Eq.\eqref{eq:Q1_il} and \(\mathcal{Q}_3 \leftarrow \) Eq.\eqref{eq:Q3_il} \vspace{1mm}
    \STATE  \(\lambda^{\text{IQR}}_{ij} \leftarrow \)  Eq.\eqref{eq:IQR_il} and \(\lambda_{ij} \leftarrow \) Eq.\eqref{eq:dynamic_detection_threshold} \vspace{1mm}
    \STATE  \(\lambda_{ij}^{\text{recover}} \leftarrow \) Eq.\eqref{eq:recovery_threshold} \vspace{1mm}
    \STATE \(n^{\text{faulty}} \leftarrow \sum_{\mathcal{B}_i[b] \in \mathcal{B}_i} \mathcal{I}[b]\)
    \IF{\(n^{\text{faulty}} \geq \Gamma\) \textbf{and} \(T^{\mathrm{lock}} \leq 0\)}
        \STATE \(\bm{\mathcal{B}}_i^{\text{cand}} \leftarrow \{ \mathcal{B}_i[b] \in \bm{\mathcal{B}}_i \mid \text{LLR}_{ij}[b] > \lambda_{ij} \}\)
        \IF{\({\cal B}_i^{\text{min}} \in \bm{\mathcal{B}}_i^{\text{cand}}\)}
            \STATE \(\mathcal{B}_i[b^*] \leftarrow {\cal B}_i^{\text{min}}\)
        \ELSE
            \STATE \(\mathcal{B}_i[b^*] \leftarrow \arg \max_{\mathcal{B}_i[b] \in \bm{\mathcal{B}}_i^{\text{cand}}} \text{LLR}_{ij}[b]\)
        \ENDIF
        \STATE \(\text{use\_backup} \leftarrow \textbf{True}\)
        \STATE \(T^{\mathrm{lock}} \leftarrow T^{\mathrm{dur}} \)
    \ELSIF{\(\text{use\_backup}\) \textbf{and} \(\lambda^{\text{med}}_{ij} < \lambda^{\text{recover}}_{ij}\) \textbf{and} \(T^{\mathrm{lock}} \leq 0\)}
        \STATE \(\text{use\_backup} \leftarrow \textbf{False}\)
        \STATE \(T^{\mathrm{lock}}\leftarrow T^{\mathrm{dur}}\)
    \ENDIF
    \STATE \(T^{\mathrm{lock}}\leftarrow \max(0,\, T^{\mathrm{lock}} - \Delta t)\)
\ENDFOR
\RETURN Fault detected or no fault -- Continue with chosen data source
\end{algorithmic}
\end{algorithm}

First, Algorithm \ref{alg:FDMM} collects \(N\) data points from both the primary parents $\mathcal{P}_i$ and the backup parents $\mathcal{P}_i^{\mathrm{back}}$ of robot $v_i$, and the sample means and covariances are calculated for these data points. Then, \emph{log}-Likelihood Ratio (LLR) tests (see Appendix~\ref{APP:LR} for details of the calculation the LR statistic) are computed for robot $v_i$ across its all backup paths $\mathcal{B}_i[b] \in \bm{\mathcal{B}}_i$, comparing the likelihood of the potentially faulty data under the Alternative hypothesis (i.e., faulty hypothesis) against the Null hypothesis (i.e., fault-free hypothesis). After calculating the LLRs, they are stored in a set, \(\mathbf{LLR}_{ij} = \{ \text{LLR}_{ij}[b] \mid \mathcal{B}_i[b] \in \bm{\mathcal{B}}_i\}.\)
Next, the algorithm calculates two thresholds.

{\bf Detection threshold:} \(\lambda_{ij}\) is computed based on the median and interquartile range (IQR) of \(N\) LLR values sorted in ascending order, denoted as \(\mathbf{LLR}^{\text{sorted}}\). This threshold ensures robust error detection and is defined as
    \begin{equation}
    \lambda_{ij} = \lambda_{ij}^{\text{med}} + 1.5 \times \lambda_{ij}^{\text{IQR}},
    \label{eq:dynamic_detection_threshold}
    \end{equation}
where \(\lambda^{\text{med}}\) is the median value and \(\lambda^{\text{IQR}}\) is the interquartile range computed as the difference between the 75th percentile (\(\mathcal{Q}_3\)) and the 25th percentile (\(\mathcal{Q}_1\)) of the LLR values. Here, let $\mathrm{m} = |\mathbf{LLR}_{ij}|$ denote the total number of LLR values. Then, the 25th and 75th percentiles are given by
    \begin{equation}
    \mathcal{Q}_1 = \mathbf{LLR}_{ij}^{\text{sorted}}\Bigl(\lceil 0.25 \times \mathrm{m} \rceil\Bigr),
    \label{eq:Q1_il}
    \end{equation}
    \begin{equation}
    \mathcal{Q}_3 = \mathbf{LLR}_{ij}^{\text{sorted}}\Bigl(\lceil 0.75 \times \mathrm{m} \rceil\Bigr).
    \label{eq:Q3_il}
    \end{equation}
Finally, the interquartile range is
    \begin{equation}
    \lambda_{ij}^{\text{IQR}} = \mathcal{Q}_3 - \mathcal{Q}_1.
    \label{eq:IQR_il}
    \end{equation}

{\bf Recovery threshold:} \(\lambda^{\text{recover}}_{ij}\) is computed using the minimum and maximum LLR values within the recent \(N\) LLR values, with a tunable factor \(\theta\). This ensures that the system only switches back to the primary path when the LLR values have stabilized across a narrow range. The recovery threshold is defined as
    \begin{equation}
    \lambda^{\text{recover}}_{ij} = \lambda^{\text{min}}_{ij} + \theta \times (\lambda^{\text{max}}_{ij} - \lambda^{\text{min}}_{ij})
    \label{eq:recovery_threshold}
    \end{equation}
where \(\lambda^{\text{min}}\) is the minimum LLR value and \(\lambda^{\text{max}}\) is the maximum LLR value within the recent \(N\) LLR values. This formulation takes into account the spread of the LLR values, ensuring that switching back to the primary path occurs only when the system has sufficiently stabilized.

After calculating the thresholds, the presence of a fault is confirmed based on the number of backup paths whose LLR exceeds the detection threshold \(\lambda_{ij}\). A binary indicator variable \(\mathcal{I}\) is set for each backup path based on whether the corresponding LLR exceeds \(\lambda_{ij}\):
\begin{equation}
\mathcal{I}[b] = \begin{cases}
1, & \text{if } \text{LLR}_{ij}[b] > \lambda_{ij} \\
0, & \text{otherwise}
\end{cases}
\label{eq:fault_declaration}
\end{equation}
A fault is declared if the number of paths indicating a fault, \(n^{\text{faulty}}\), meets or exceeds a majority threshold \(\Gamma\), typically set to the ceiling of half the number of backup paths:
\[
n^{\text{faulty}} \geq \Gamma
\]

Upon detecting a fault, the algorithm identifies a set of candidate backup paths, $\bm{\mathcal{B}}_i^{\text{cand}}$, that have LLRs exceeding the detection threshold \(\lambda_{ij}\). The minimum-cost backup path \({\cal B}_i^\text{{min}}\) is selected if it is among the candidates; otherwise, an alternative backup path \({\cal B}_i^{\text{alt}}\) with the highest LLR is chosen.

Switching back to the primary path occurs when the fault condition resolves. Detection of fault resolution is based on comparing the median LLR value, \(\lambda^{\text{\text{med}}}_{ij}\), with the recovery threshold \(\lambda^{\text{recover}}_{ij}\). If the median LLR value drops below the recovery threshold, it indicates that the system has stabilized, and it is safe to revert to the primary path. This mechanism ensures that the switch back to the primary path happens only when the fault has truly subsided, preventing premature switching in the presence of temporary fluctuations in the LLR values.

To ensure stability, a lock-in timer $T^{\mathrm{lock}}$ prevents frequent switching between the primary and backup paths. After switching to a path, the system enforces a waiting period (lock-in time) before reevaluating the conditions for switching back, stabilizing the decision-making process.

\begin{figure*}[hbt]
\centering
\subfigure[]{
\includegraphics[width=0.40\textwidth]{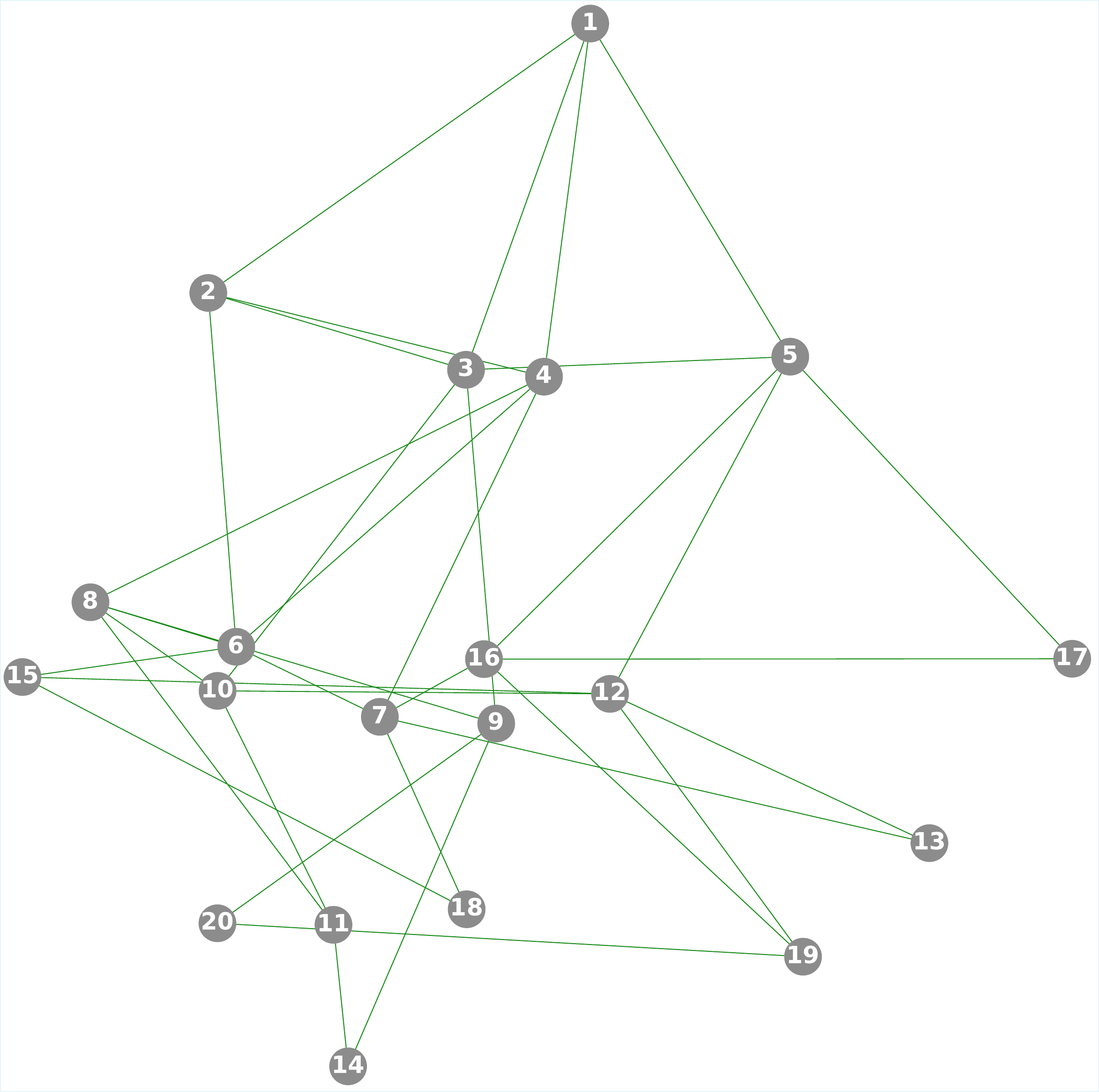}
}
\subfigure[]{
\includegraphics[width=0.36\textwidth]{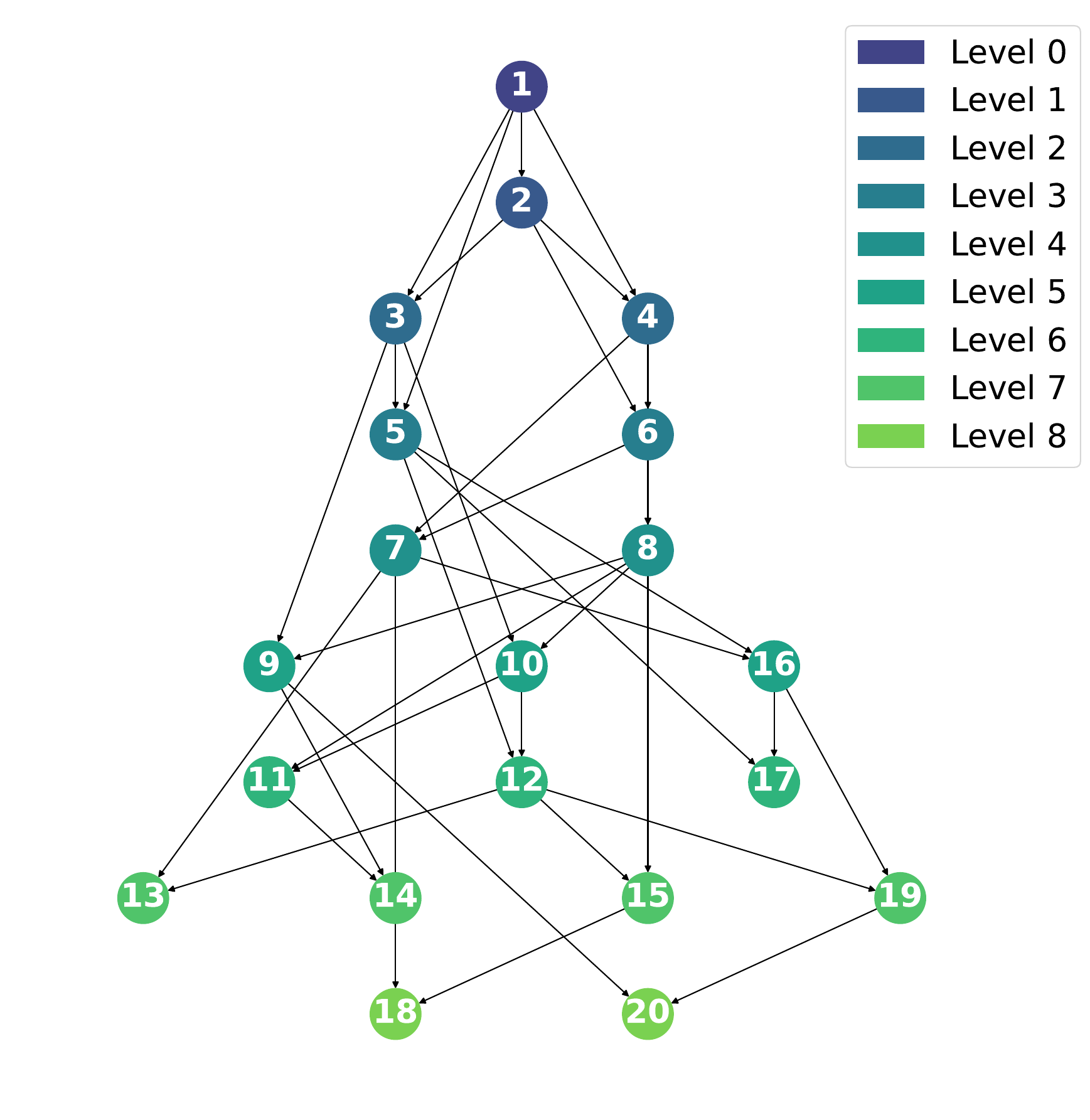}
}
\caption{{\bf Example scenario 1: Example primary network.} (a) A randomly formed HHC-constructed graph of 20 robots. Node 1 is the leader robot. (b) The hierarchical structure underlying the formation in (a), where robots are arranged in levels based on their hop count from the leader (Level 0). Each robot is assigned a level that is one higher than the highest level among its parent robots, and information flows from higher levels (starting from Level 0) to lower levels.}
\label{fig:Swarm20}
\end{figure*}

\subsection{Properties of fault detection and mitigation}

\textbf{Runtime:} The time complexity of Algorithm~\ref{alg:FDMM} is primarily influenced by the computation of sample means, covariance matrices, the LLR values, and sorting the LLRs. For each parent robot, these calculations are performed for every backup path and involve operations over the \(N\) data points in each path. Computing the sample mean \(\boldsymbol{\mu}\) of a dataset with \(N\) points in \(d\)-dimensional space (\(d=2\) here) requires \(O(Nd)\) time; since \(d\) is constant, this is \(O(N)\) per dataset. Similarly, computing the sample covariance \(\boldsymbol{\Sigma}\) takes \(O(Nd^2)=O(N)\) (again \(d\) is constant). Thus, for each backup path, the total to compute mean+covariance is \(O(N)\).

The per-path LLR can be evaluated in \(O(1)\) \emph{if} we use the already computed sufficient statistics (mean, covariance, and \(N\)) rather than re-summing over samples (cf. Eq.~\eqref{eq:LLR} written in terms of sufficient statistics). Sorting the \(|\bm{\mathcal{B}}_i|\) LLR values costs \(O(|\bm{\mathcal{B}}_i|\log|\bm{\mathcal{B}}_i|)\). Therefore, for a single parent \(v_j\), the total time is $O\!\big(N|\bm{\mathcal{B}}_i|\big) \;+\; O\!\big(|\bm{\mathcal{B}}_i|\log|\bm{\mathcal{B}}_i|\big).$

If the statistics of the potentially faulty parent stream \(\mathbf{\tilde{Q}}_{ij}\) are reused across all backup paths (computed once), the bound remains the same. For completeness, multiplying by the in-degree \(\delta_i^{\mathrm{in}}\) gives a per-robot bound \(O\!\big(\delta_i^{\mathrm{in}}(N|\bm{\mathcal{B}}_i| + |\bm{\mathcal{B}}_i|\log|\bm{\mathcal{B}}_i|)\big)\); in the HHC setting \(\delta_i^{\mathrm{in}}\in\{1,2\}\).

{\bf Memory space requirements:} The space complexity is dominated by the storage requirements for the data points, covariance matrices, and LLR values. Each backup path requires storage for \(N\) data points of dimension \(d\), totaling \(O(N  d)\) space per path. For all backup paths, this amounts to \(O(N d |\bm{\mathcal{B}}_i|)\). With \(d\) being constant, this simplifies to \(O(N |\bm{\mathcal{B}}_i|)\). Each covariance matrix and mean vector requires \(O(d^2)\) and \(O(d)\) space, respectively, per dataset. Since \(d\) is constant, the total space for all backup paths is \(O(|\bm{\mathcal{B}}_i|)\), which is negligible compared to the data storage. Storing the LLR values for all backup paths requires \(O(|\bm{\mathcal{B}}_i|)\) space. Therefore, the overall space complexity is: $O(N |\bm{\mathcal{B}}_i|)$

{\bf Efficiency:} Given a limited number of backup paths \(|\bm{\mathcal{B}}_i|\) as well as reasonable values of \(N\), the algorithm remains efficient and suitable for real-time fault detection. The linear scaling with \(N\) in both time and space complexities indicates that the algorithm can handle larger datasets without significant performance penalties.

\section{Validation}
\label{sec:validation}

Using experimental results in simulation, we first demonstrate the adaptability of the backup network layer to reconfigurations in the original graph. We demonstrate this because we assume that in all scenarios in which tolerance to IFs is desired, the underlying network will be time-varying in some manner. For this demonstration, we choose the following scenario: after a backup layer has been constructed in a swarm with a non-moving leader, there is a substantial reconfiguration of the primary HHC-constructed graph because a high proportion of the swarm is removed, thus requiring the backup layer to adapt. Second, we demonstrate our {\it proactive-reactive} fault tolerance strategy for detecting and mitigating intermittent faults in an HHC-constructed formation with a moving leader, and compare its detection of simple and complex IFs to that of a centralized benchmark. Finally, we demonstrate IF detection and mitigation using our {\it proactive-reactive} method in a proof-of-concept swarm of 200 robots.

\subsection{Adaptability of backup layer to reconfigurations in an example swarm}
\label{sec:Illustrative numerical example_1}

We consider a group of 20 robots in a 2-dimensional space, in an example HHC-constructed graph with eight hierarchy levels. In Fig.~\ref{fig:Swarm20}(a), the HHC-constructed graph and the spatial layout of the formation are shown, with the robots represented as nodes. The HHC-constructed graph was generated randomly, but with the leader only allowed to have up to four children and each follower up to three. The hierarchical relationships among the robots are illustrated in Fig.~\ref{fig:Swarm20}(b).

\begin{figure*}[h!]
\centering
\subfigure[]{\includegraphics[width=0.43\textwidth]{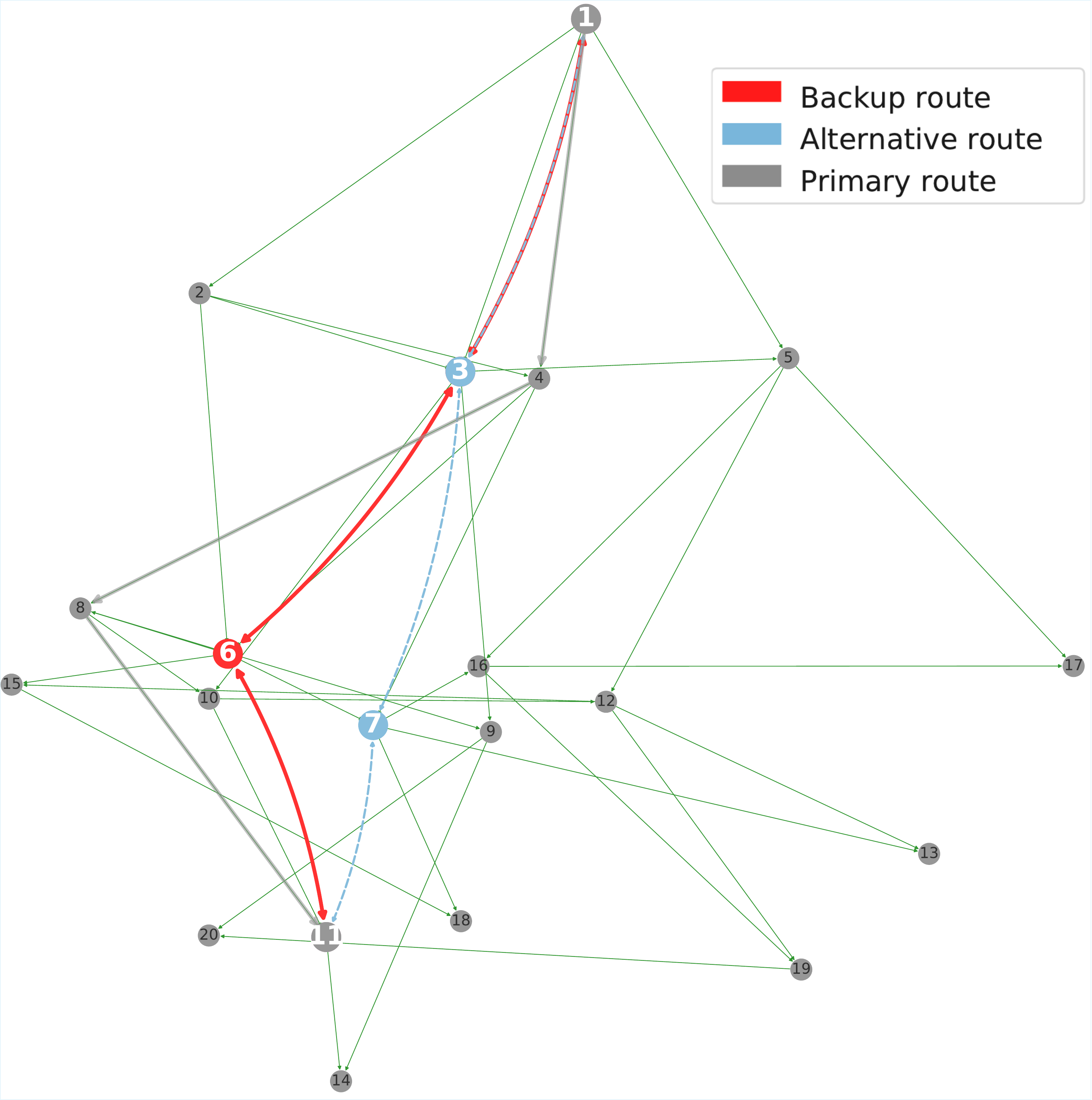}}
\subfigure[]{\includegraphics[width=0.23\textwidth]{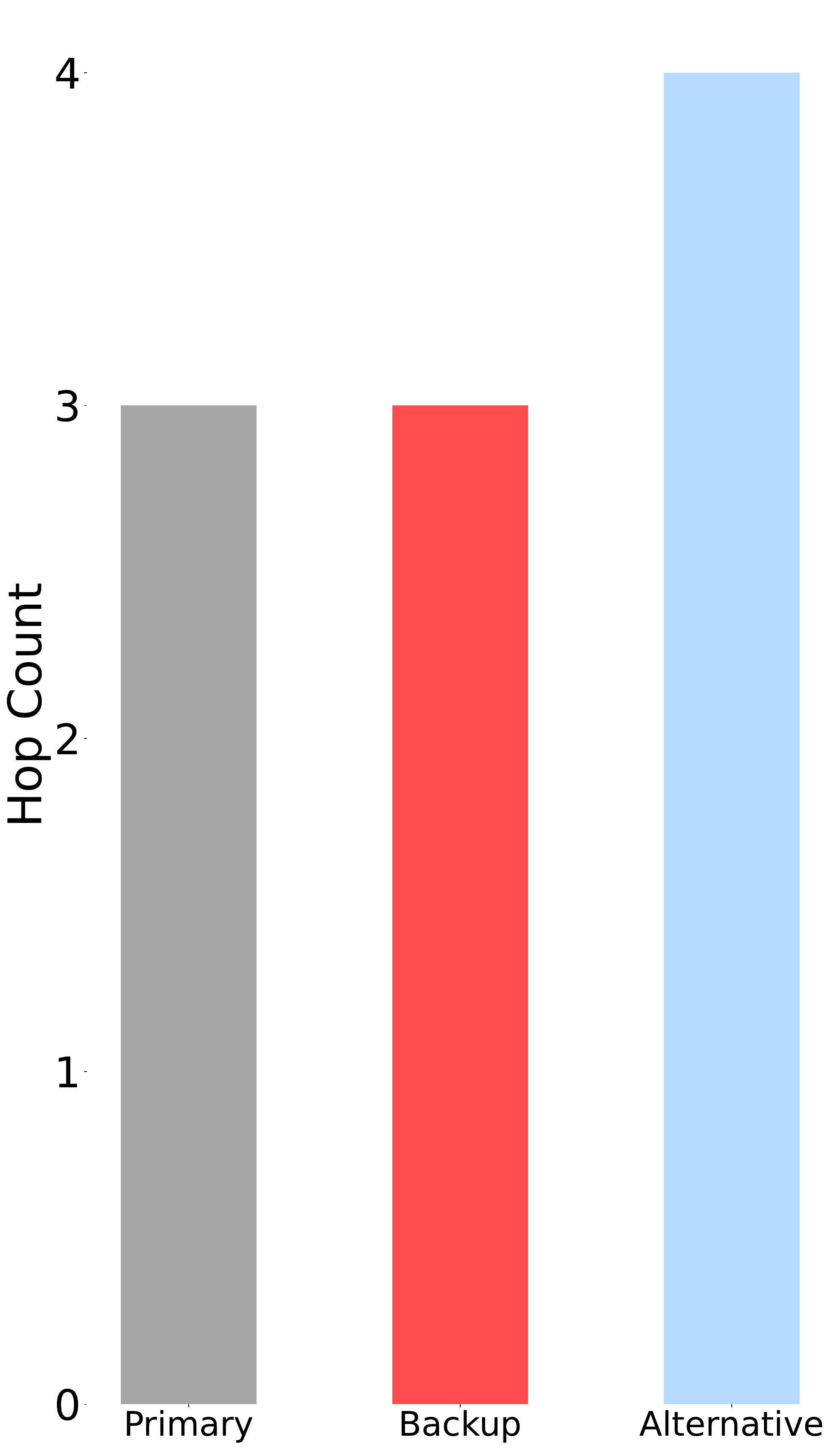}}
\raisebox{0.0\height}{
\raisebox{-0.1cm}{
  \subfigure[]{\includegraphics[width=0.55\textwidth]{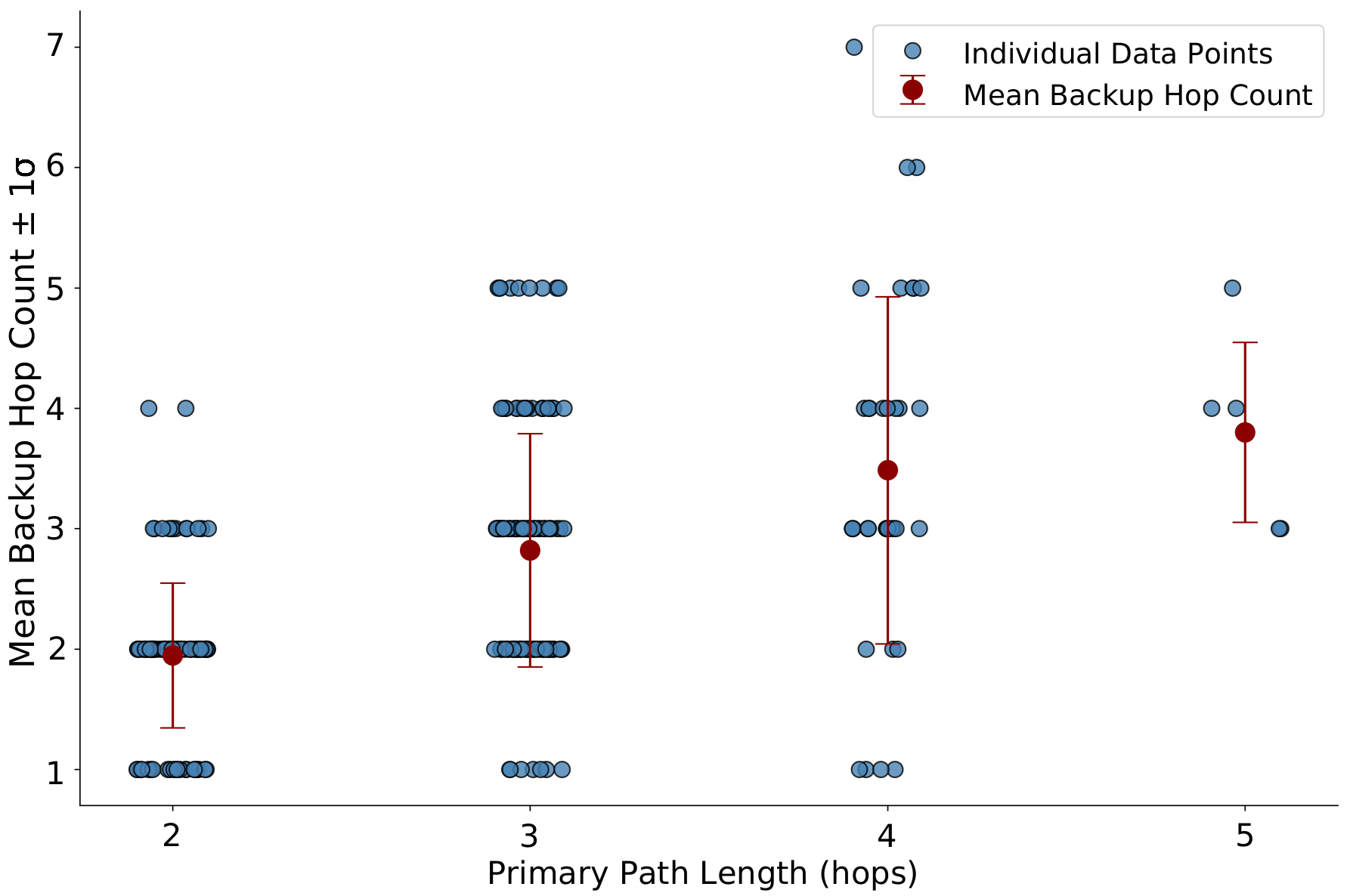}}
}
}
\caption{{\bf Construction of an example backup layer.} (a) ABMC protocol performance for the formation shown in Fig.~\ref{fig:Swarm20}. The shortest path to robot 11 (chosen as a representative example robot) in the primary network is highlighted in gray. The minimum-cost backup path to the leader robot generated by the ABMC protocol for robot 11 is highlighted in red, while the alternative backup path is highlighted in blue. The shortest primary path follows the sequence of robots 1 \(\rightarrow\) 4 \(\rightarrow\) 8 \(\rightarrow\) 11. In this notation, \(\rightarrow\) the single arrow indicates a directed link, specifying the intended direction of communication along the primary path. The minimum-cost backup path is 1 \(\rightarrow\) 3 \(\rightarrow\) 6 \(\rightarrow\) 11, and an alternative path is 1 \(\rightarrow\) 3 \(\rightarrow\) 7 \(\rightarrow\) 11. (b) Comparison of the shortest path in the primary network for robot 11 and the number of hops for the minimum-cost and alternative backup paths. (c) Backup path performance aggregated by primary path length across 20 independent ABMC trials on distinct graph realizations. For each primary‑path hop count (2--5 hops), the mean backup-path hop count (of all follower robots, all 20 trials) is shown in red, with error bars denoting \(\pm\,1\sigma\). The individual backup-path hop counts are shown in blue. Note that path lengths of 1 hop are omitted, since only follower faults are considered in this study, and thus direct connections to the leader do not require a backup path.}
\label{fig:ABMC1}
\end{figure*}

\begin{figure*}[h!]
\centering
%\subfigure[]{
%\raisebox{0.0\height}{
%\includegraphics[width=0.27\textwidth]{11.pdf}
%}
%}
\subfigure[]{
\includegraphics[width=0.34\textwidth]{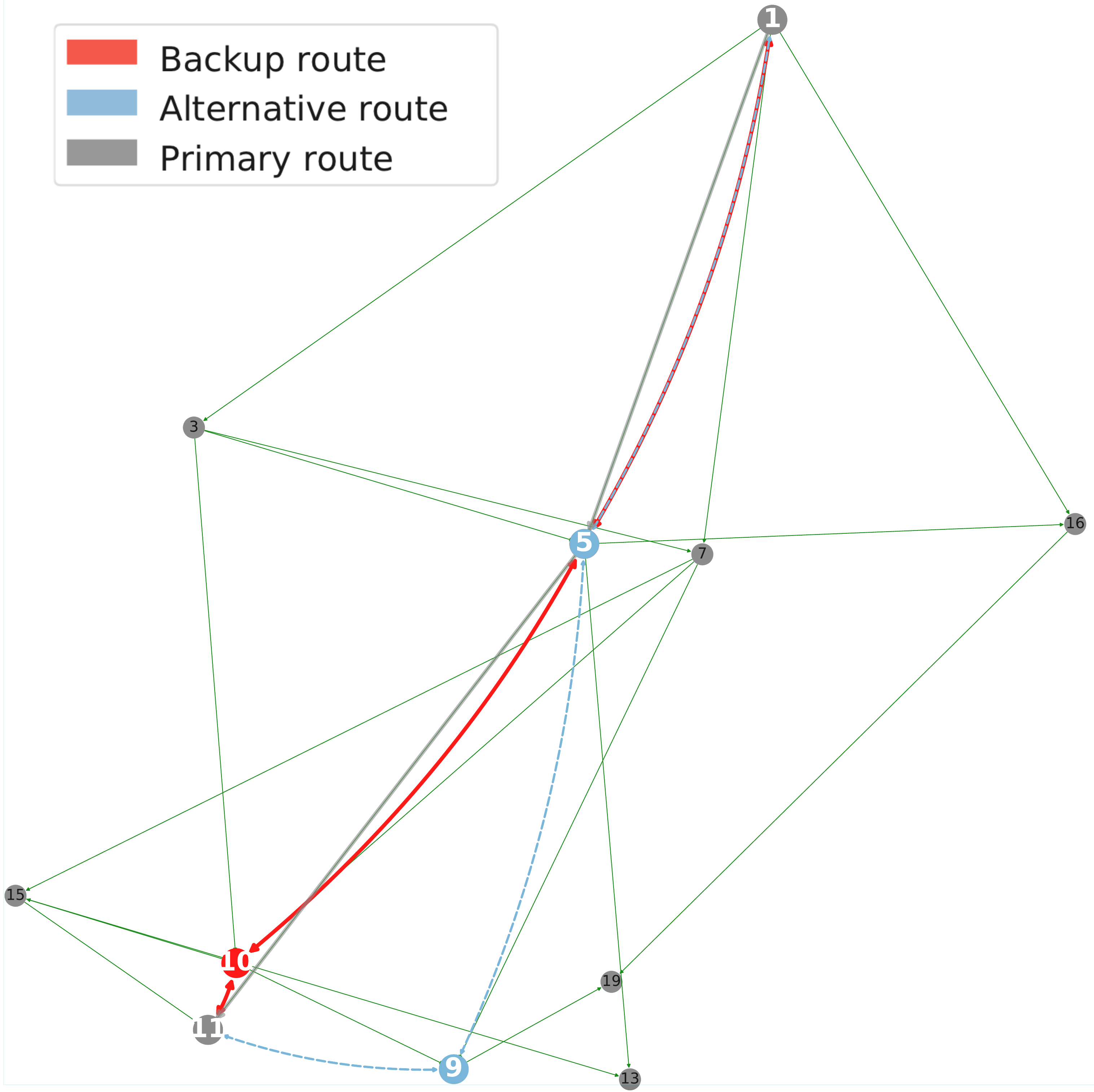}
}
\subfigure[]{
\includegraphics[width=0.58\textwidth]{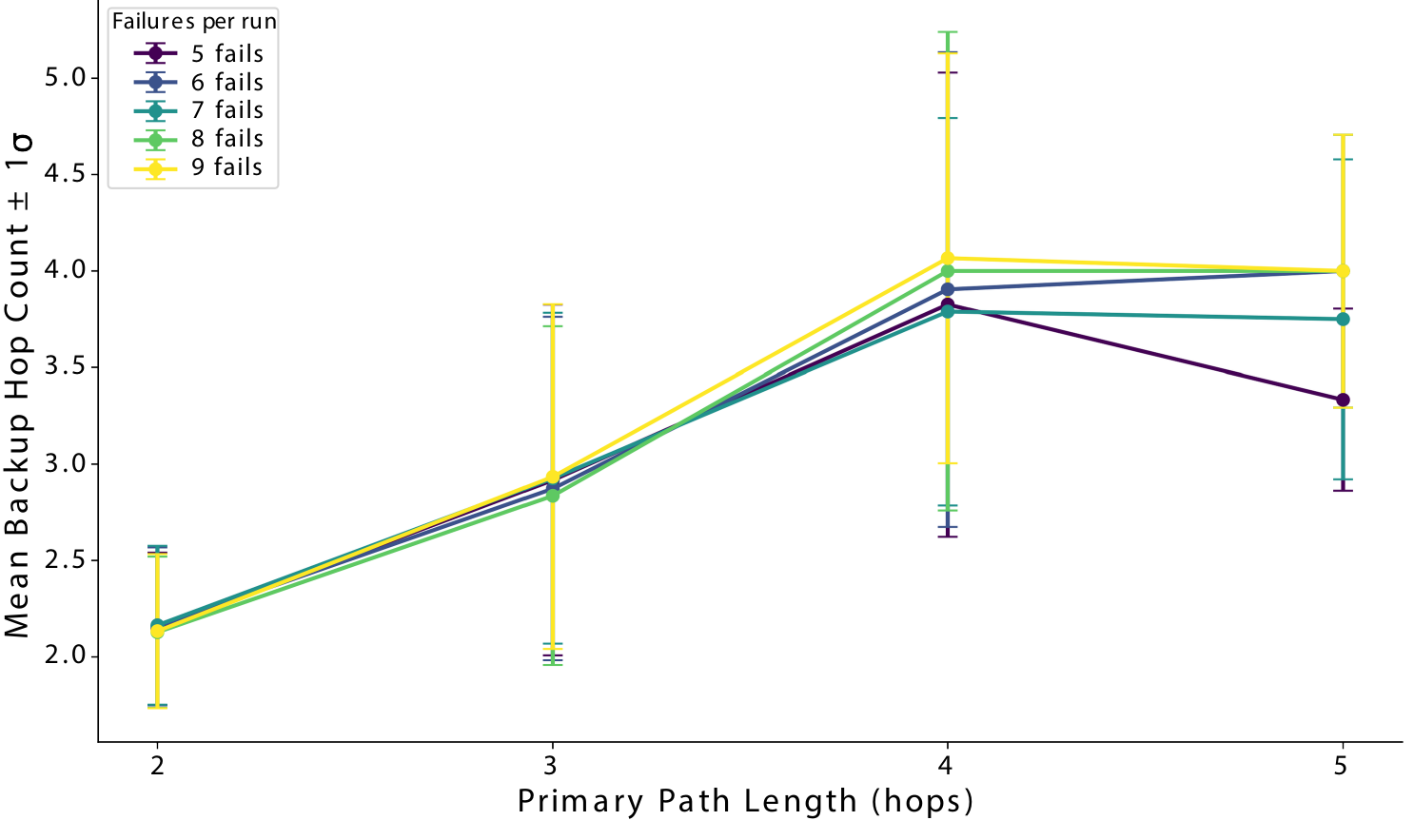}
}
\caption{{\bf Example backup layer after adapting to the reconfiguration of the original graph.} (a) Reconfigured robot formation following the robot removals shown in Fig.~\ref{fig:Swarm20}. Robots 12, 17, 8, 2, 4, 18, 14, 6, and 20 have been removed in this representative example case. %(b) 
In the reconfigured formation, the new shortest path for robot 11 is shown in gray, while the minimum-cost backup and an alternative path are depicted in red and blue, respectively. The revised primary path now follows the sequence 1 $\rightarrow$ 5 $\rightarrow$ 11, with the backup path as 1 $\rightarrow$ 5 $\rightarrow$ 10 $\rightarrow$ 11 and the alternative path as 1 $\rightarrow$ 5 $\rightarrow$ 9 $\rightarrow$ 11. 
(b) Aggregated backup‐path performance under random robot removals in 20 trials, with 5--9 robots (25\%--45\%) removed per trial. For each removal rate, the mean backup-path hop count (of all surviving follower robots, all 20 trials) according to primary‑path hop count is shown, with one‐sigma error bars (no interpolation).}
\label{fig:ABMC4}
\end{figure*}

In Fig.~\ref{fig:ABMC1}(a), the simulation results for node 11 (with parameters configured as $\eta = 0.1$, $\kappa_d = 6$, and $r = 2$ meters) are shown as a representative example, alongside its corresponding path in the primary network.
The simulation results established one minimum-cost and one alternative backup path for node 11. In Fig.~\ref{fig:ABMC1}(b), it is shown that both the primary path and the minimum-cost backup path have a hop count of 3, while the alternative backup path has a hop count of 4.
Thus, in the event of a fault in the primary path, the minimum-cost backup path can be used without increasing the number of hops, minimizing any potential latency increase or disruption to network performance. Although the alternative backup path involves one additional hop, it can still serve as a fail-safe to maintain communication.
In Fig.~\ref{fig:ABMC1}(c) we group the backup-path hop counts according to their respective primary‐path hop counts. For primary paths of 2 and 3 hops, the mean backup-path hop count was $\approx2.0 \pm 0.5$ and $\approx3.0 \pm 0.8$, respectively, showing that approximately 90\% of robots found an equally short or only one‐hop‐longer backup path. For primary paths of 4 and 5 hops, the mean was $\approx3.5 \pm 1.5$ and $\approx4 \pm 0.7$ hops, respectively, showing larger variability in longer paths.

The adaptability of the ABMC protocol to network changes is shown in Fig.~\ref{fig:ABMC4}. 
Following the random removal of 9 out of 20 robots in a representative example trial, the primary network reconfigures using HHC-reconstruction (see Appendix~\ref{APP:Reconfig} for reconfiguration details of the HHC implementation). The new primary network is shown in Fig.~\ref{fig:ABMC4}(a). Then, to construct new backup paths following the reconfiguration, each robot re-executes the ABMC protocol to update its backup edges. In Fig.~\ref{fig:ABMC4}(b), the simulation results for the new HHC-reconfigured graph are presented, comparing the minimum-cost and alternative backup paths for node 11 to its path in the primary network.  In this case, the primary path had a hop count of 2, while both the minimum-cost and alternative backup paths had a hop count of 3. 
In Fig.~\ref{fig:ABMC4}(c), we plot the utilized backup-path hop counts according to their respective primary‐path hop counts, grouped by removal rate (5--9 robots, or 25--45\% of the swarm, are randomly removed in each trial). 
For primary paths of 2 hops, the mean backup-path hop count increases only slightly from $\approx2.1$ hops at 5 removals to $\approx2.2$ hops at 9 removals, with $\sigma$ only increasing from $\approx0.2$ to $\approx0.35$. 
The removal rate has a moderately increasing effect on primary paths of 3 and 4 hops, with an occasional 1 or 2 hop increase occurring for 3-hop primary paths and an occasional 2 or 3 hop increase for 4-hop primary paths.
For primary paths of 5 hops, the removal rate has a much more marked effect on the backup-path hop count, but the relatively small number of samples at this length might contribute to the spread.  
Overall, even with up to 45\% of robots removed, the ABMC protocol always finds a backup path, and the backup paths are usually within 1--2 hops of the primary path (increases of 3 hops occur only occasionally). Mean hop counts increase only modestly with the removal rate, until reaching primary paths of 5 hops (for which there are few samples).

\begin{figure*}[ht]
\centering
\includegraphics[width=0.99\textwidth]{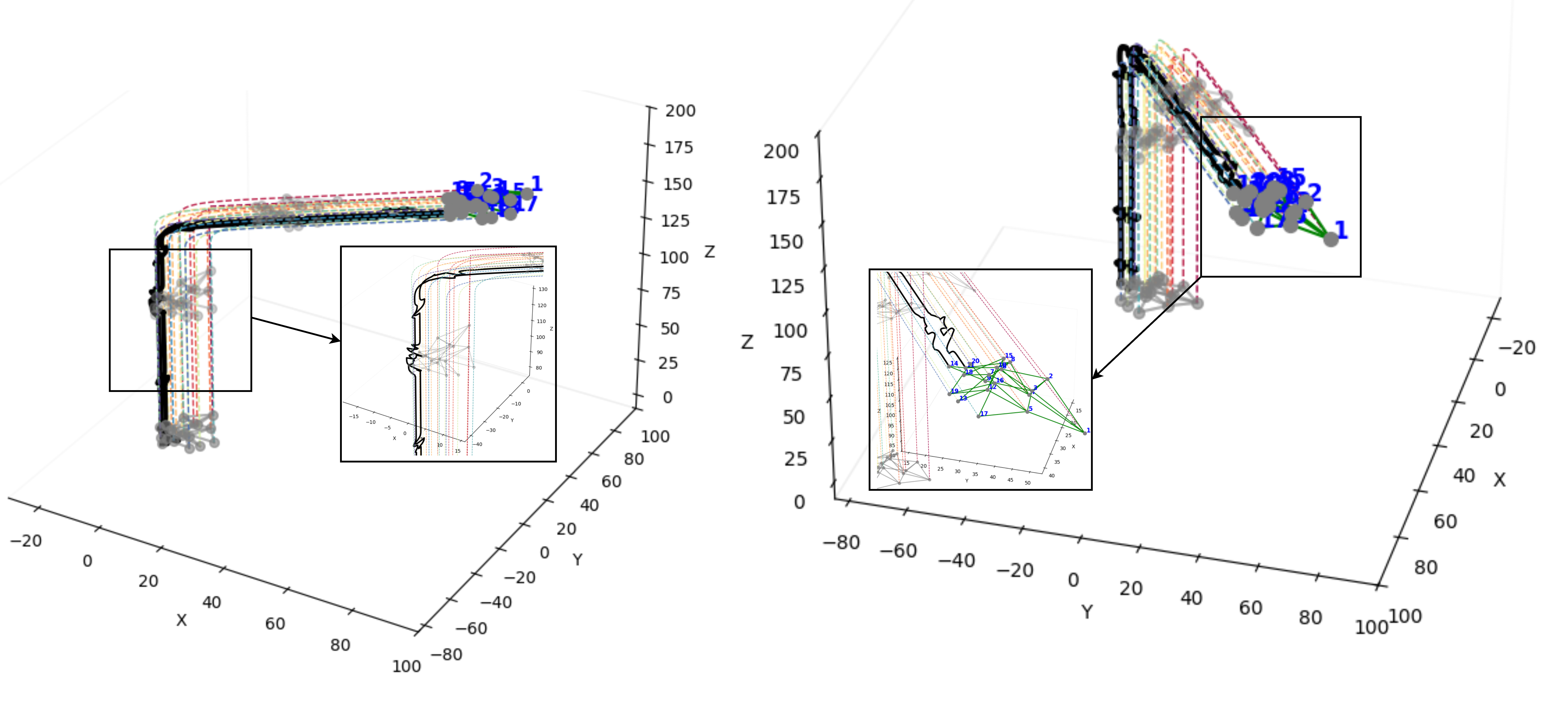}
\caption{{\bf The effect of intermittent faults without the use of a fault tolerance mechanism.} The effect of intermittent faults on a 3D navigation scenario, using the example formation given in Fig.~\ref{fig:Swarm20}. In this example, \( P_f = 0.5 \) and \(P_e = 0.02\).
Robots 11 and 14 are highlighted as illustrative examples, with their trajectories shown in bold black: robot 11 experiences intermittent offset faults in the relative‑position data from its parent, robot 8. The inset highlights the moments when the error affects the trajectory of robots 11 and 14. 
}
\label{fig:Formation_nav}
\end{figure*}

\subsection{Intermittent faults in an example swarm}
\label{sec:Illustrative numerical example_2}

For an example of intermittent faults, we use a setup like that in Sec.~\ref{sec:Illustrative numerical example_1}, but with a formation operating in 3D space while being subjected to intermittent faults on the \(x\) and \(y\) axes. The goal is to maintain the formation while navigating, despite transient erroneous data.

The leader robot begins its motion along the \(z\)-axis at a constant speed. 
As the leader robot moves, the rest of the formation follows, maintaining relative positions as per the prescribed formation accompanying the HHC-constructed graph. During navigation, certain robots are exposed to intermittently faulty relative position data with the fault occurrence probability \(P_f\), combined with communication noise modeled by the channel bit error probability \(P_e\) (see Appendix~\ref{APP:Data} for details of the fault and noise model).

In Fig.~\ref{fig:Formation_nav}, the simulation result is presented for the robot at node 11 in the formation given in Fig.~\ref{fig:Swarm20}, as a representative example. 
The robot at node 11 is affected by faulty relative position data intermittently reported by one of its primary parents (robot at node 8).
Initially, they follow a stable trajectory, moving along with the formation. Then, as intermittent faults affect the relative position data received from node 8, deviations are observed in their movement. These deviations, although minor initially, can accumulate over time, leading to noticeable discrepancies in the formation.

\begin{figure*}[ht]
\centering
\subfigure[]{
\includegraphics[width=0.45\textwidth]{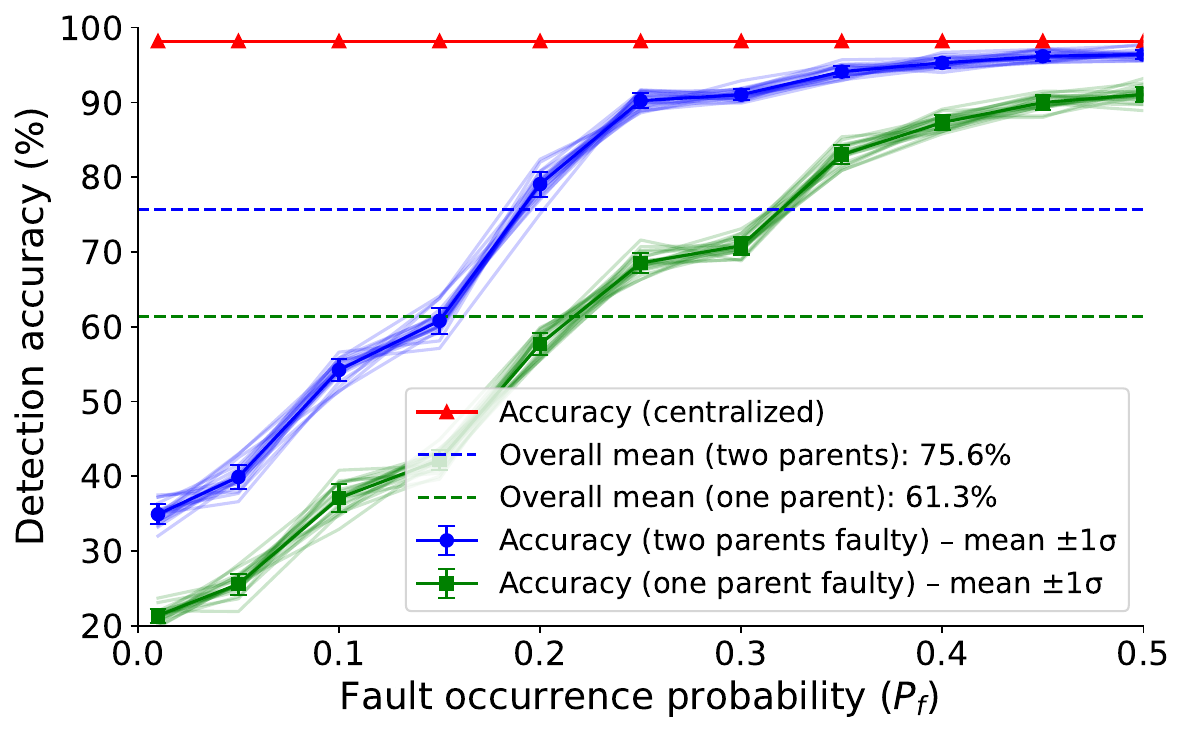}
}
\subfigure[]{
\includegraphics[width=0.45\textwidth]{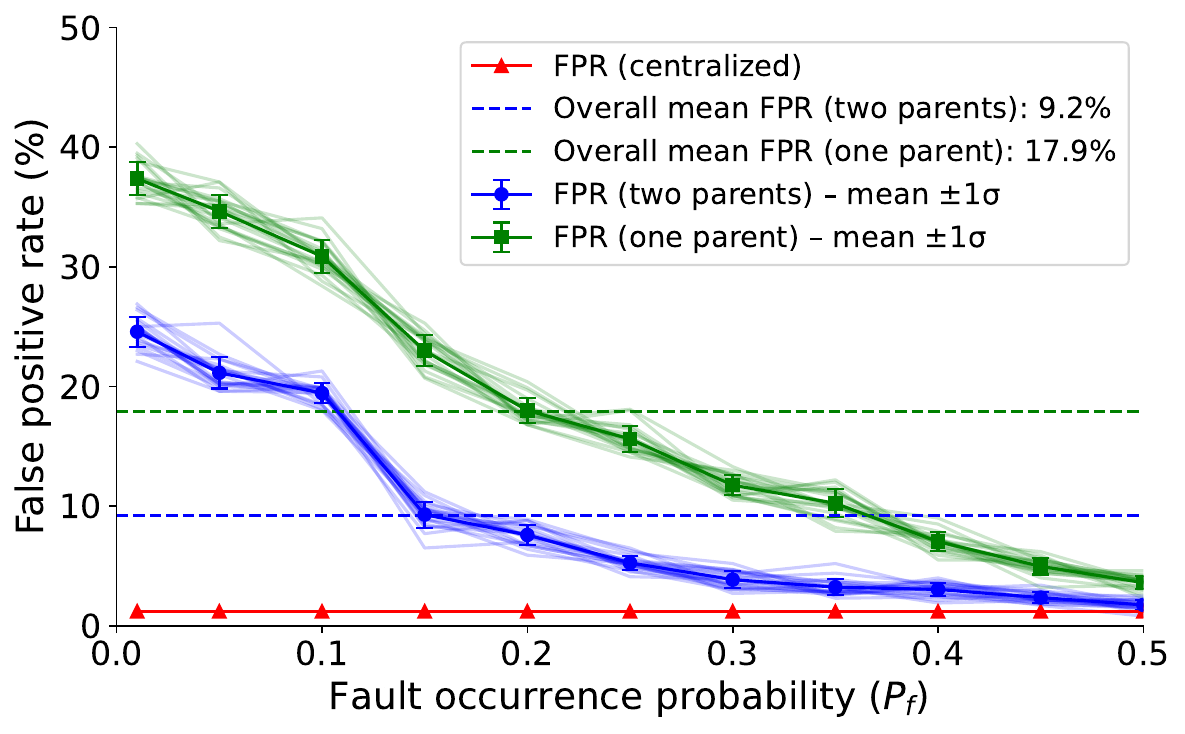}
}
\caption{{\bf Detection of simple intermittent faults.} Performance of the fault detection strategy under varying fault‐occurrence probabilities for the formation control scenario given in Fig.~\ref{fig:Formation_nav}, at fixed channel bit error probability $P_e = 0.02$.
(a) Detection accuracy: 20 independent simulation runs for each case (two faulty parents shown in blue, one faulty parent shown in green).
Thin lines show individual trials; bold lines show the mean \(\pm1\sigma\) across runs, with vertical error bars. Dashed horizontal lines show the overall average accuracy across all fault probabilities for each case (centralized benchmark is shown in red). 
(b) False positive rate, in the same plot style as (a).}
\label{fig:Fault_prob}
\end{figure*}

\subsection{Detection of simple and complex intermittent faults}

To assess the detection accuracy and false positives rate of our {\it proactive-reactive} method, we compare its performance to a benchmark case with a centralized detection network.
In the centralized benchmark, the leader robot acts as an information hub: it disseminates the correct reference values to all robots, gathers all measurements returned by all robots (each corrupted only by channel bit error $P_e$), and then decides whether the sender is faulty or not. 
The samples the leader receives are randomly drawn from two overlapping distributions: \(f_0\) if the sender is not faulty and \(f_1\) (shifted by the fault offset) if the sender is faulty. To separate these noisy and partially overlapping distributions, the leader applies the \emph{Bayes-optimal likelihood-ratio test} by computing \(\Lambda=f_1/f_0\) and comparing this ratio with a fixed threshold that maximizes \(\tfrac12(\mathrm{TPR}+\mathrm{TNR})\) for the given noise level \(P_e\).

In this section, we compare the performance of this centralized benchmark to that of our {\it proactive-reactive} method, in the formation control scenario of Sec.~\ref{sec:Illustrative numerical example_2} (see Fig.~\ref{fig:Formation_nav}). We test our {\it proactive-reactive} method in the cases of one faulty parent and two faulty parents.

Fig.~\ref{fig:Fault_prob} presents the results of 20 independent Monte Carlo trials per fault probability~\(P_f\), with the channel bit error probability $P_e$ held constant (at $P_e = 0.02$). In Fig.~\ref{fig:Fault_prob}(a), the \(1\sigma\) bands remain very narrow (under 3\%) at low fault rates (\(P_f \le 0.10\)), indicating highly consistent behavior across trials. Variability grows modestly at the mid fault rates (\(P_f \approx 0.15\)--\(0.25\)), especially for the case of one faulty parent. For the case of two faulty parents, only a few outlier runs remain before variability reaches nearly zero at the high fault rates (\(P_f \ge 0.40\)). For the case of one faulty parent, the accuracy rises gradually from \(\approx 22\%\) to \(\approx 90\%\). With two faulty parents, accuracy climbs from \(\approx 34\%\) at \(P_f=0.01\) to over 90\% by \(P_f=0.30\), approaching the centralized bound.
Similarly to the accuracy, the false-positive rate variability, see Fig.~\ref{fig:Fault_prob}(b), reaches nearly 4\% for the case of two faulty parents by \(P_f=0.35\).

\begin{figure*}[ht]
\centering
\includegraphics[width=0.9\textwidth]{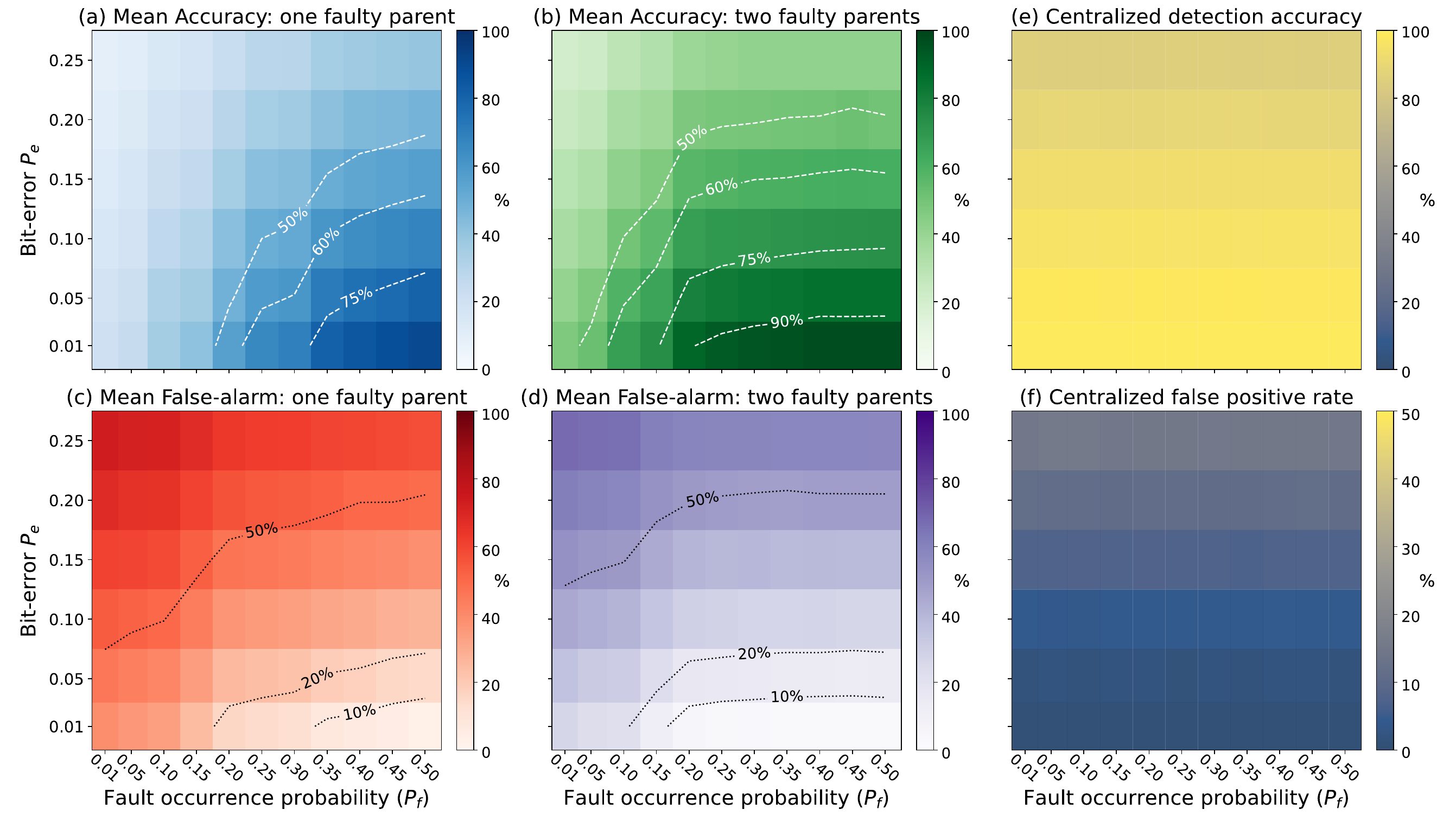}
  \caption{
    {\bf Detection of complex intermittent faults.} 
    (a-d) Detection performance of our {\it proactive-reactive} method under joint offset faults and channel noise, in the cases of (a,c) one faulty parent and (b,d) two faulty parents. Heatmaps (20 trials per grid‐cell) showing the (a-b) mean detection accuracy and (c-d) false positives rate as functions of fault occurrence probability \(P_f\) and bit‑error probability per hop \(P_e\). (e) Mean detection accuracy, and (f) mean false positive rate of the centralized benchmark, as a function of the fault occurrence probability \(P_f\), for varying channel bit‐error probabilities \(P_e\).
  }
\label{fig:Fault_prob2}
\end{figure*}

Fig.~\ref{fig:Fault_prob2} presents the mean detection accuracy (sensitivity) and false positive rate (specificity) of our {\it proactive-reactive} method, computed over 20 Monte Carlo runs per grid cell, for varying fault occurrence probabilities \(P_f \in [0.01, 0.50]\) and channel bit error probabilities \(P_e \in [0.01, 0.25]\). 
For the case of one faulty parent, Fig.~\ref{fig:Fault_prob2}(a) demonstrates that detection accuracy climbs steeply with $P_f$ even under light noise. At $P_e=0.01$, accuracy rises from about 27\% at $P_f=0.01$ to cross the 50\% contour at $P_f = 0.20$, and the 75\% contour once $P_f\gtrsim0.35$. Accuracy only falls below 50\% in the extreme low‐$P_f$, high‐$P_e$ corner, and remains above 10\% across most of the $(P_f,P_e)$ plane. Fig.~\ref{fig:Fault_prob2}(c) shows false positives beginning at $33\%$ for $(P_f=0.01,P_e=0.01)$, and falling below the 10\% contour for $P_f\gtrsim0.35$. The 60\% accuracy and 10\% false positives contours delineate a safe operating envelope of roughly $P_f\gtrsim0.35, P_e\lesssim0.10$, within which the sensitivity versus specificity of detection is balanced. 

For the case of two faulty parents, Fig.~\ref{fig:Fault_prob2}(b) exhibits an even sharper rise in accuracy: under minimal noise ($P_e=0.01$), accuracy rises from $\lesssim 40\%$ at $P_f = 0.01$ to cross the 50\% accuracy contour by $P_f = 0.05$, and the 90\% contour appears near $P_f = 0.20$. As \(P_e\) increases these breakpoints shift rightward. In Fig.~\ref{fig:Fault_prob2}(d), false‐positives start at about $21\%$ and fall under the 10\% contour by $P_f=0.15$ at $P_e=0.01$. The 75\% accuracy and 10\% false positive contours delineate a safe envelope around $P_f\gtrsim0.25, P_e\lesssim0.10$, showing that even when both parents are faulty, detection sensitivity versus specificity can be balanced effectively. 

Compared to the centralized benchmark, see Fig.~\ref{fig:Fault_prob2}(e,f), our {\it proactive-reactive} method sacrifices 10--30\,\% accuracy at very low fault rates or under heavy noise, yet it narrows the gap to within a few percentage points of the upper bound established by the centralized benchmark once moderate fault activity appears. It also achieves lower false positive rates compared to the centralized benchmark in the high-\(P_f\), low-noise regimes that pose the greatest risk during bursts.

\begin{figure}[ht]
\centering
\subfigure[]{
\includegraphics[width=0.45\textwidth]{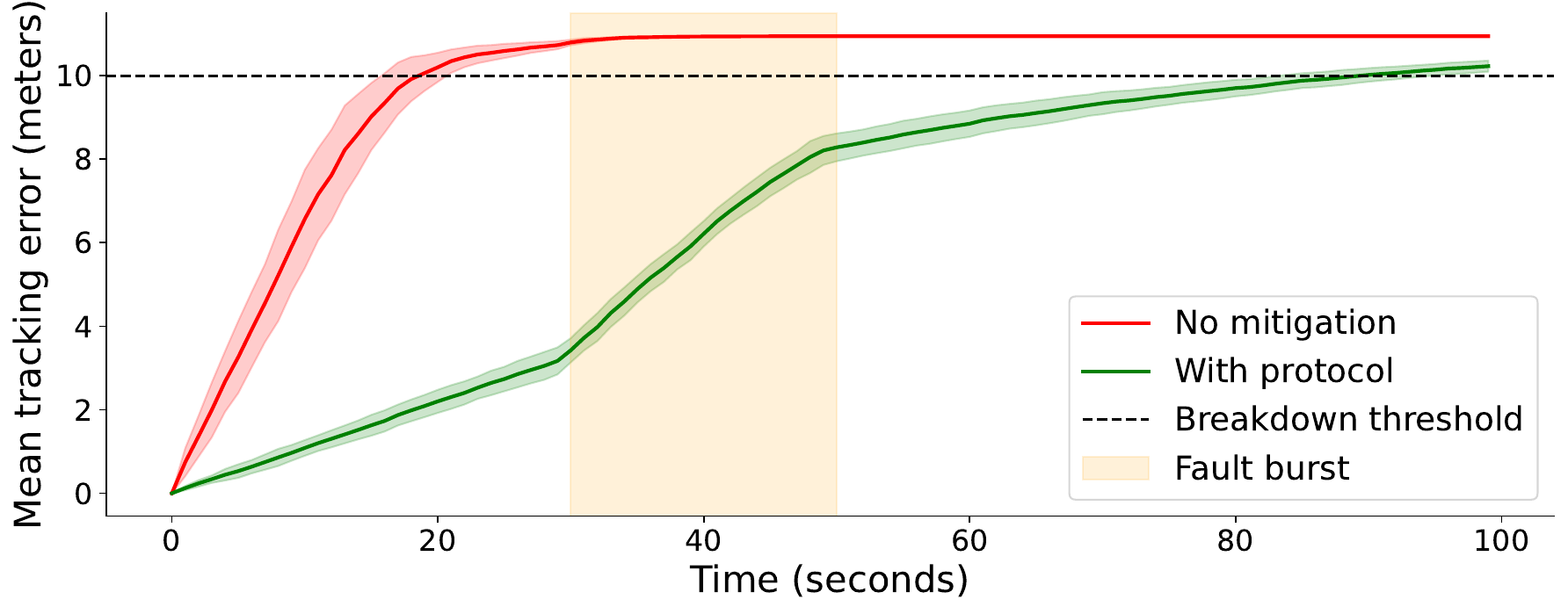}
}
\subfigure[]{
\includegraphics[width=0.45\textwidth]{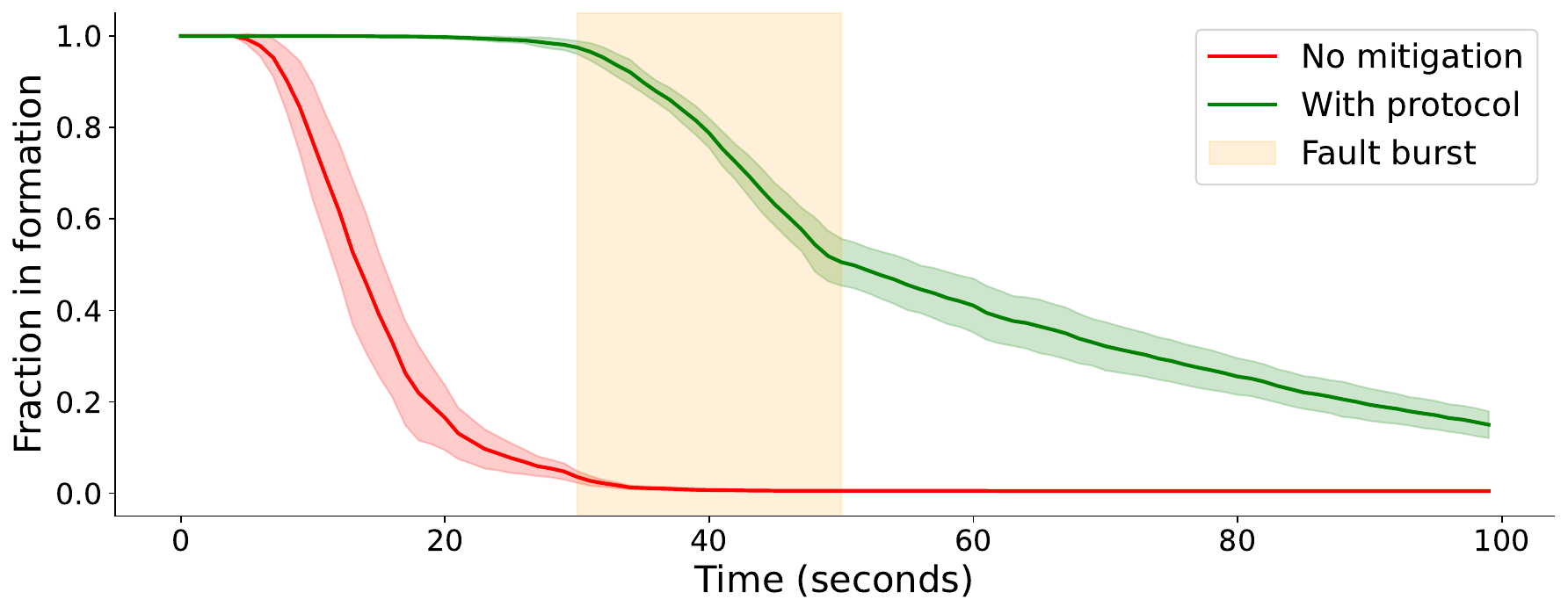}
}
\caption{{\bf Simulation results with a swarm of 200 robots.} Tracking performance of a 200-robot hierarchical swarm under time-varying IFs, with a burst period ($P_f = 0.35$) from $t = 30$ to $t = 50$~s and a low background fault rate ($P_f = 0.1$) otherwise. Results are averaged over 20 independent trials, with shaded regions indicating $\pm 1\sigma$ across trials. The orange band marks the fault burst interval. The impact of the burst phase is governed by the rate of undetected faults, which is determined by: \emph{undetected fault rate} $= P_f \times (1 - \text{detection accuracy})$. (a) Mean absolute tracking error $\overline{e}(t)$ in meters. Tracking errors are cumulative, and robots that exceed the breakdown threshold (shown as dashed black line) are counted as irrecoverable for the remainder of the trial. 
(b) Proportion of robots maintaining the formation as a function of time.}
\label{fig:large_swarm_proof_burst}
\end{figure}

\subsection{Demonstration in a swarm of 200 robots}

This section presents a proof-of-concept demonstration of a 200-robot swarm under time-varying fault conditions, in the scenario of two faulty parents. We conducted 20 independent Monte Carlo simulation trials of a 200-robot HHC-constructed formation exposed to IFs, both with our {\it proactive-reactive} method and in a baseline swarm with no IF-tolerance strategy. In the simulation scenario, we first expose both swarms to a low background probability ($P_f = 0.1$) of IFs occurring randomly in robots at each second. Our {\it proactive-reactive} method performs almost perfectly during this period, while the baseline swarm experiences full breakdown. Therefore, to also study the limitations of our {\it proactive-reactive} method, we secondly expose both swarms to a high-intensity fault burst ($P_f = 0.35$) modeled as a fixed time interval ($t=30$ to $t=50$ seconds). 

Fig.~\ref{fig:large_swarm_proof_burst}(a) presents the mean absolute tracking error $\overline{e}(t)$, averaged across all robots and trials. Without any fault tolerance or mitigation strategy (red), errors accumulate rapidly and exceed the breakdown threshold (dashed black line) well before the burst, after which they remain saturated. 
With our {\it proactive-reactive} method (green), error growth is well-contained during the background phase, but increases more rapidly during the burst interval.
The increase during the burst can be attributed to the fact that, although per‐fault detection accuracy improves as the fault occurrence rate increases (see Fig.~\ref{fig:Fault_prob2}), the total fault occurrence is so high that the absolute rate of undetected faults can stay constant or rise (i.e., multiplying by a larger $P_f$ can outweigh the accuracy gain). For example, moving from $P_f=0.10$ at 60\% detection to $P_f=0.35$ at 90\% detection changes only from $0.10\times0.40=0.04$ to $0.35\times0.10=0.035$. Because each undetected fault contributes permanently to cumulative error, the green curve in Fig.~\ref{fig:large_swarm_proof_burst}(a) still climbs during the burst—though it remains below the threshold far longer than in the unmitigated case. After the burst, error growth reverts to its slower background rate.

Fig.~\ref{fig:large_swarm_proof_burst}(b) shows the proportion of robots maintaining the formation with a tracking error below the breakdown threshold, over time. Without any fault tolerance or mitigation strategy (red), the formation fraction approaches zero (i.e., all robots exceeding the allowable error) within 25\,s and remains there. With our {\it proactive-reactive} method (green), nearly all robots in the swarm hold the formation before the burst, and during the burst they experience a gradual decline rather than a catastrophic collapse. After the burst, the loss rate reduces substantially, but around half of the robots have already been lost and the swarm cannot recover its pre-burst rate (which was nearly no loss). Still, at the end of the trial at $t=100$\,s, a substantial core of robots ($>30$) is still active in the formation.

Overall, the method is extremely effective at detecting and mitigating IFs in low background probability ($P_f = 0.1$) in a 200-robot swarm, especially compared to the baseline swarm with no IF-tolerance method, which degrades completely even under this low IF condition.
During a high-intensity burst of IFs, although the per‐fault detection accuracy improves during the burst, the much higher volume of faults results in the \emph{absolute rate} of undetected faults still growing, so the total number of undetected events remains substantial. In short, in large swarms experiencing substantial high-intensity bursts of IFs, a fault saturation threshold does exist, after which \emph{formation quality} will still degrade when using our {\it proactive-reactive} method, in its current form. However, under this fault saturation threshold, our {\it proactive-reactive} method is highly effective, dramatically outperforming the unmitigated baseline and enabling robust operation even in challenging conditions. 

\section{Conclusions and future work}
\label{sec:Conclusions}

\subsection{Conclusions}
In this paper, we presented a novel {\it proactive--reactive} fault-tolerance strategy designed to address the challenges posed by intermittent faults (IFs) in robot swarms, particularly for establishing reliable multi-hop communication paths and maintaining desired formations.
Proactively, the strategy uses \textit{adaptive biased minimum consensus} (ABMC) for robots to self-organize dynamic backup communication paths before faults occur. The ABMC protocol extends the existing biased minimum consensus (BMC) to dynamic networks of mobile robots. ABMC incorporates dynamic network characteristics, dynamic robot relative positions, and dynamic bias terms, and also allows the establishment of new edges not already present in the original network. This protocol ensures that each robot can continually adapt its backup path to the leader robot, enhancing the swarm's resilience to communication disruptions.
Reactively, the strategy uses one-shot likelihood ratio tests for fault detection and mitigation. By comparing information received via primary and backup paths, robots can detect statistically significant deviations indicative of IFs. Upon IF detection along a primary path, communication is temporarily rerouted through a backup path, until the IF resolves. Thus, reliable flow of positional information is maintained and the effect of IFs on the swarm's formation control performance is minimized.
Overall, this work provides a proof-of-concept of proactive--reactive fault tolerance against IFs in self-organized hierarchical robot swarms, a step towards addressing the broad challenges posed by IFs in general.

\subsection{Future work}

The present work delivers a proof-of-concept of detecting and mitigating data-centric IFs in a self-organized hierarchical robot swarm, but it would be important for future work to expand the types of data-centric IFs that are considered.
Firstly, we have specifically addressed offset faults. While offset faults are indeed a common issue across various sensor modalities, actuator types, and communication channels, other data-centric intermittent faults, such as stuck-at and spike faults, also warrant attention.
Secondly, the present work assumes that some fault-free path exists among the constructed multiplex network (primary paths and backup paths). However, in real-world scenarios, there could be IFs that affect all possible paths in the multiplex network from a robot to its leader. Similarly, the present work assumes the leader itself is fault-free, but this will not always be the case in real-world scenarios. Future research could investigate fault tolerance methods for IFs that synchronously affect many robots (e.g., a large proportion of the robots with few hops to the leader, or a large proportion of the swarm overall) as well as IFs that effect the leader robot.
Finally, future work should also validate the approach with larger swarms exposed to IFs in complex environments, as well as validate the approach with real robots exposed to IFs in field experiments.

\backmatter

%\bmhead{Supplementary information}
%If your article has accompanying supplementary file/s please state so here. 
%Authors reporting data from electrophoretic gels and blots should supply the full unprocessed scans for key as part of their Supplementary information. This may be requested by the editorial team/s if it is missing.
%Please refer to Journal-level guidance for any specific requirements.

%\bmhead{Acknowledgements}

\section*{Declarations}

\bmhead{Funding} 

This work is supported by the Program of Concerted Research Actions (ARC) of the Universit{\'e} libre de Bruxelles and by the Office of Naval Research Global (Award N62909-19-1-2024).
Marco Dorigo and Mary Katherine Heinrich acknowledge support from the Belgian F.R.S.-FNRS, of which they are a Research Director and a Research Associate, respectively.

\bmhead{Competing interests}

The authors declare they have no competing interests.

\bmhead{Data availability}

The dataset for this study has been deposited at Zenodo: https://doi.org/10.5281/zenodo.20427761.

\bmhead{Code availability}

The software for this study has been deposited at Zenodo: https://doi.org/10.5281/zenodo.20427761.

%Some journals require declarations to be submitted in a standardised format. Please check the Instructions for Authors of the journal to which you are submitting to see if you need to complete this section. If yes, your manuscript must contain the following sections under the heading `Declarations':

%\begin{itemize}
%\item Conflict of interest/Competing interests (check journal-specific guidelines for which heading to use)
%\item Ethics approval and consent to participate
%\item Consent for publication
%\item Data availability 
%\end{itemize}

%\noindent
%If any of the sections are not relevant to your manuscript, please include the heading and write `Not applicable' for that section. 


\clearpage
\begin{appendices}

%\section{Section title of first appendix}\label{secA1}
\section{HHC-constructed Graphs and SoNS}
\label{App:HHC}
\setcounter{equation}{0}
\renewcommand{\theequation}{\Alph{section}.\arabic{equation}}

In Hierarchical Henneberg Construction (HHC)~\citep{zhang2023self}, the interaction graph \(\mathcal{G}\) is a rooted directed graph constructed from a starting kernel of two special nodes (the leader $v_1$ and first follower $v_2$). All other nodes besides the starting kernel are standard followers and have exactly two parents, with two different levels of hierarchy \(\mathcal{H}\)) (i.e. the hop count of the longest directed path to the leader \(v_1\)). Details of graph construction and reconstruction (robot addition, framework merging, robot departure, and framework splitting) using HHC can be found in~\cite{zhang2023self}.

%An
We assume an HHC-constructed graph is related to the {\it self-organizing nervous systems (SoNS)}~\citep{zhu2024self} framework in the following way.
A SoNS is a rooted tree used for high-level coordination in which each child has exactly one parent. It can be used in combination with different formation control approaches, in this case with an HHC-constructed graph to support formation control based on relative bearing. For any robot \(v_i\) that is both a follower in an HHC-constructed graph and a child in a SoNS, its SoNS parent will be the same robot as its higher-hierarchy HHC parent (that is, its HHC parent further from the HHC leader \(v_1\) and thus with a higher hierarchy level \(\mathcal{H}\)).

\color{black}

\section{Proofs of Lemmas and Theorems}
\label{App:Proofs}

\subsection*{Proof of Lemma 1}
\label{Proof_L1}
\setcounter{equation}{0}
\renewcommand{\theequation}{\Alph{section}.\arabic{equation}}

Recall $\beta_i(t)$ and the definition of $\mathcal{P}_i^{\mathrm{back}}$ in Eq.~\eqref{eq:minimazers}. For $v_i\in\mathcal{V}_1$, $\dot{\beta}_i(t)=0$. For $v_i\in\mathcal{V}_2$, set $m_i(t)\;=\;\min_{v_j\in\mathcal{P}^{\mathrm{cand}}_i(t)}\bigl\{\,s_j(t)+a_{ij}(t)\,\bigr\}$. Because the $\min$ operator may be non-differentiable, we use the upper right Dini derivative $D^+$ and Clarke’s chain rule~\citep{clarke1990optimization}. It yields $D^+ m_i(t)\ \in\ \operatorname{conv}\Bigl\{\,\dot s_j(t)+\dot a_{ij}(t):~v_j\in\mathcal{P}_i^{\mathrm{back}}\,\Bigr\}$. By Remark~\ref{rm:5}, $a_{ij}(t)$ is piecewise constant on $[t_k,t_{k+1})$, hence $\dot a_{ij}(t)=0$ there. Therefore $D^+\beta_i(t)\ \in\ -\,\dot s_i(t)\;+\;\operatorname{conv}\Bigl\{\,\dot s_j(t):~v_j\in\mathcal{P}_i^{\mathrm{back}}\,\Bigr\}$.

Thus, there exist weights $w_j\ge 0$ with $\sum_{v_j\in\mathcal{P}_i^{\mathrm{back}}}w_j=1$ such that
\begin{equation}
D^+\beta_i(t)\;=\;-\,\dot s_i(t)\;+\;\sum_{v_j\in\mathcal{P}_i^{\mathrm{back}}}w_j\,\dot s_j(t)
\label{dot_beta_refined}
\end{equation}
Using the update rule $\eta\,\dot s_\ell(t)=\beta_\ell(t)$, we have $\dot s_\ell(t)=\beta_\ell(t)/\eta$ for all $\ell$. Substituting into \eqref{dot_beta_refined} gives
\begin{equation}
D^+\beta_i(t)\;=\;\frac{1}{\eta}\sum_{v_j\in\mathcal{P}_i^{\mathrm{back}}}w_j\bigl(\beta_j(t)-\beta_i(t)\bigr)
\label{eq:beta_update}
\end{equation}

Now consider $\bar\beta(t)=\max_{v_i\in\mathcal V}\beta_i(t)$. By Clarke’s max rule,
\begin{equation}
D^+\bar\beta(t)\ \le\ \max_{v_i\in\overline{\mathcal S}(t)} D^+\beta_i(t)\ \le\ \sum_{v_i\in\overline{\mathcal S}(t)}\delta_i\,D^+\beta_i(t)
\end{equation}
for some $\delta_i\ge 0$ with $\sum_{v_i\in\overline{\mathcal S}(t)}\delta_i=1$, where $\overline{\mathcal S}(t)=\{v_i:\beta_i(t)=\bar\beta(t)\}$. Substituting \eqref{eq:beta_update} yields
\begin{equation}
D^+\bar\beta(t)\ \le\ \sum_{v_i\in\overline{\mathcal S}(t)}\sum_{v_j\in\mathcal{P}_i^{\mathrm{back}}}\frac{\delta_i w_j}{\eta}\,\bigl(\beta_j(t)-\beta_i(t)\bigr)
\label{eq:derivative_barBeta}
\end{equation}
For $v_i\in\overline{\mathcal S}(t)$ we have $\beta_i(t)=\bar\beta(t)\ge \beta_j(t)$ for all $v_j$, hence each difference $\beta_j(t)-\beta_i(t)\le 0$. Since $\delta_i\ge 0$, $w_j\ge 0$, and $\eta>0$, every term in \eqref{eq:derivative_barBeta} is $\le 0$, so $D^+\bar\beta(t)\le 0$. Thus $\bar\beta(t)$ is monotonically non-increasing.

For the lower envelope $\underline\beta(t)=\min_{v_i}\beta_i(t)$, note that $\underline\beta(t)=-\max_{v_i}(-\beta_i(t))$. Applying the same argument to $-\beta_i(t)$ gives $D^+\underline\beta(t)\ge 0$, i.e., $\underline\beta(t)$ is monotonically non-decreasing. This completes the proof. \hfill$\blacksquare$

\subsection*{Proof of Lemma 2}
\label{Proof_L2}

From Lemma 1, we have $\underline{\beta}(0) \leq \underline{\beta}(t) \leq \beta_i(t) \leq \bar{\beta}(t) \leq \bar{\beta}(0), ~ \forall\, v_i \in \mathcal{V}, \; t \geq 0$. Since \(\bar{\beta}(t)\) is monotonically non-increasing and bounded below, the Monotone Convergence Theorem~\citep{royden1988real} implies $\lim_{t\to+\infty}\bar{\beta}(t) = c_1, ~ \text{with} ~ \underline{\beta}(0) \leq c_1 \leq \bar{\beta}(0)$. By definition, for every \(v_i\) in $\overline{\mathcal{S}} = \{ v_i \in \mathcal{V} \mid \beta_i(t) = \bar{\beta}(t) \}$, we have $\lim_{t\to+\infty}\beta_i(t) = c_1$. For any \(v_j\in\mathcal{P}_i^{\text{back}}\), noting that \(\beta_j(t) \leq \bar{\beta}(t)\), we obtain $\lim_{t\to+\infty}\beta_j(t) \leq c_1$. As \(t\to+\infty\) (with \(\bar{\beta}(t)\to c_1\)), using the upper right Dini derivative and Eq.~\eqref{eq:derivative_barBeta} gives
\begin{equation}
\begin{aligned}
    0 \;\ge\; &\limsup_{t\to+\infty} D^+\bar{\beta}(t)
\;\ge\; \\
& \limsup_{t\to+\infty}
\sum_{v_i\in\overline{\mathcal{S}}\cap\mathcal{V}_2}\ \sum_{v_j\in\mathcal{P}_i^{\text{back}}} \frac{\delta_i w_j}{\eta}\, \Bigl( \beta_j(t) - c_1 \Bigr)
\end{aligned}
\end{equation}
where $\delta_i\ge 0$, $\sum_{v_i\in\overline{\mathcal S}}\delta_i=1$, and $w_j\ge 0$, $\sum_{v_j\in\mathcal P_i^{\text{back}}}w_j=1$. Since each coefficient is nonnegative, the only way for the limit superior of this weighted sum of nonpositive terms to be zero is that each active term satisfies $\beta_j(t)-c_1\to 0$. Hence
\begin{equation}
\lim_{t\to+\infty}\beta_j(t) = c_1, \quad \forall\, v_j\in\mathcal{P}_i^{\text{back}}
\end{equation}

By LaSalle's Invariance Principle~\citep{isidori2013nonlinear}, the system trajectories converge to the largest invariant set where $D^+\bar{\beta}(t)=0$. In this invariant set, we must have
$ \mathcal{P}_i^{\text{back}} \subset \overline{\mathcal{S}}, \quad \forall\, v_i\in\overline{\mathcal{S}}$, i.e., as \(t\to+\infty\), robots with maximum \(\beta_i(t)\) interact only with robots that also achieve \(\beta_i(t)=c_1\). This completes the proof. \hfill \(\blacksquare\)

\subsection*{Proof of Lemma 3}
\label{Proof_L3}

Since the candidate parent set \(\mathcal{P}^{\mathrm{cand}}_i(t)\) is finite, there exists at least one robot \( v_n \in \mathcal{P}_i^{\text{back}}\) such that
\(\min_{v_j \in \mathcal{P}^{\mathrm{cand}}_i(t)} \{ s_j(t) + a_{ij}(t) \} = s_n(t) + a_{in}(t)\).
Therefore, by the update rule we have \(s_i(t) = s_n(t) + a_{in}(t) - \beta_i(t)\). Since \(\beta_i(t) \ge \underline{\beta}(0)\), we have 
\begin{equation}
    s_i(t) \le s_n(t) + a_{in}(t) - \underline{\beta}(0)
    \label{eq:localIneq}
\end{equation}
More generally, for any \(v_j\in\mathcal{P}^{\mathrm{cand}}_i(t)\),
\begin{equation}
s_i(t)=\min_{v_\ell\in\mathcal P^{\mathrm{cand}}_i(t)}\{s_\ell(t)+a_{i\ell}(t)\}-\beta_i(t)
\;\le\; s_j(t)+a_{ij}(t)-\underline{\beta}(0)
\end{equation}
so \eqref{eq:localIneq} applies to each backup parent.

To extend this local bound to a global one, we use the \emph{reachability via backup parents} assumption: for every robot \(v_i\in\mathcal V_2\) there is a finite directed backup path connecting it to the leader robot \(v_1 \in \mathcal{V}_1\), say
\(\{v_i, \dots, v_n,\, \dots,\, v_1\}\), where for each consecutive pair \((v_{n}, v_{n+1})\) along the path we have \(v_{n+1}\in {\cal P}^{\mathrm{cand}}_n(t)\).
Then, for each link along the path,
\(s_n(t) \le s_{n+1}(t) + a_{n,n+1}(t) - \underline{\beta}(0)\).
If we define \(C = a_{\max} - \underline{\beta}(0)\), where \(a_{\max}\) is an upper bound on all link costs, then for each link we have \(s_n(t) \le s_{n+1}(t) + C\).
By chaining these inequalities from the robot \(v_i\) to the leader robot \(v_1\) (across, say, \(n-1\) links) gives
\(s_i(t) \leq s_{1}(0) + (n-1)C\).
Let \(n_{\max}\) be the maximum number of hops needed for any robot in \(\mathcal V_2\) to reach the leader robot along backup parents.
Then, by taking the worst-case path we define \(S_{\max} = s_{1}(0) + (n_{\max}-1)C\).
Hence, for every \(v_i \in \mathcal{V}_2\) and all \(t > 0\), \(s_i(t) \le S_{\max}\).
This completes the proof. \hfill \(\blacksquare\)

\subsection*{Proof of Lemma 4}
\label{Proof_L4}

Recall that $\beta_1(t)=0$ for all $t$. Suppose, for contradiction, that $\underline{\mathcal{S}}(t) \cap \mathcal{V}_1 = \emptyset$ for arbitrarily large $t$. Then, since $\underline{\beta}(t)=\min_i\beta_i(t)$ and $\beta_1(t)=0$, we must have $\underline{\beta}(t)<0$ for those $t$.

From the definition of $\mathcal{P}^{\text{back}}_i$, we have $\mathcal{P}^{\text{back}}_i \neq \emptyset$ for every $v_i \in \underline{\mathcal{S}}(t)$. Moreover, by Lemma~2 (lower-bound case),  $\mathcal{P}^{\text{back}}_i \subset \underline{\mathcal{S}}(t)\quad \text{for all}\quad v_i \in \underline{\mathcal{S}}(t)$ for sufficiently large $t$. For any such $t$, pick $v_i\in\underline{\mathcal{S}}(t)$ and any $v_n\in\mathcal{P}^{\text{back}}_i$. Using
\begin{equation}
\begin{split}
\beta_i(t) & \;=\; -\,s_i(t) \;+\; \min_{v_j \in \mathcal{P}^{\mathrm{cand}}_i(t)} \bigl\{s_j(t)+a_{ij}(t)\bigr\} \\
&\;=\; -\,s_i(t) \;+\; s_n(t)+a_{in}(t)
\end{split}
\end{equation}

we obtain
\begin{equation}
s_i(t) \;=\; s_n(t)+a_{in}(t)-\beta_i(t)
\label{eq:si-expansion}
\end{equation}
Since $\beta_i(t)=\underline{\beta}(t)<0$ and $a_{in}(t)\ge \gamma>0$, \eqref{eq:si-expansion} gives the strict inequality
\begin{equation}
s_i(t) \;>\; s_n(t)+\gamma \;>\; s_n(t)
\label{eq:strict}
\end{equation}
for all $v_n\in\mathcal{P}^{\text{back}}_i \subset \underline{\mathcal{S}}(t)$.

Now define $s_{m}(t):=\min_{v_\ell \in \underline{\mathcal{S}}(t)} s_\ell(t)$ and choose $v_i\in\underline{\mathcal{S}}(t)$ attaining this minimum. Then $s_i(t)=s_m(t)\le s_n(t)$ for every $v_n\in\underline{\mathcal{S}}(t)$, which contradicts \eqref{eq:strict}. Therefore, our supposition must be false, and $\underline{\mathcal{S}}(t) \cap \mathcal{V}_1 \neq \emptyset$ for all sufficiently large $t$. This completes the proof. \hfill$\blacksquare$

\subsection*{Proof of Theorem 1}
\label{Proof_T1}

Building on Lemma~1, there exist constants $c_1,c_2\in\mathbb{R}$ such that 
$\lim_{t\to\infty}\bar{\beta}(t)=c_1$ and 
$\lim_{t\to\infty}\underline{\beta}(t)=c_2$. 
By Lemma~4, there exists an increasing, unbounded sequence 
$\{t_k\}_{k\in\mathbb{N}}$ with $t_k\to\infty$ such that 
$\underline{\mathcal{S}}(t_k)\cap \mathcal{V}_1 \neq \emptyset$ for all 
$k\in\mathbb{N}$. 
For each $k$, pick $v_i \in \underline{\mathcal{S}}(t_k)\cap \mathcal{V}_1$. 
Since $\beta_j(t)=0$ for all $v_j\in\mathcal{V}_1$ and all $t$, we have 
$\underline{\beta}(t_k)=0$ for every $k$. 
Hence $\liminf_{t\to\infty}\underline{\beta}(t)=0$. 
Because $\underline{\beta}(t)$ has a (finite) limit $c_2$, it follows that 
$c_2=0$; that is, $\lim_{t\to\infty}\underline{\beta}(t)=0$.

Now, consider the definition of \(\bar{\beta}(t)\), which implies \(\bar{\beta}(t) \geq \underline{\beta}(t)\). For any time $t$ and any robot \(v_i \in \overline{{\cal S}}(t)\), the ABMC protocol dictates that $\eta \dot{s}_i(t) = \beta_i(t) = \bar{\beta}(t)$. Summing over all robots and using that every non-maximizer has $\beta_i(t)\ge \underline{\beta}(t)$, we obtain the pointwise bound
\begin{equation}
\eta \sum_{v_i \in \mathcal{V}} \dot{s}_i(t)
=\sum_{v_i\in\mathcal V}\beta_i(t)
\ \ge\ |\overline{{\cal S}}(t)|\,\bar{\beta}(t)\;+\;\big(|\mathcal V|-|\overline{\cal S}(t)|\big)\,\underline{\beta}(t)
\label{eq:ptwise-sum-bound}
\end{equation}

Fix $\varepsilon>0$. Since $\underline{\beta}(t)\to 0$, there exists $T_\varepsilon$ such that $\underline{\beta}(t)\ge -\varepsilon$ for all $t\ge T_\varepsilon$. Integrating \eqref{eq:ptwise-sum-bound} from $T_\varepsilon$ to $\tau$ and using Lemma 3 (boundedness of $s_i(t)$) yields that the left-hand side is uniformly bounded in $\tau$, whereas the right-hand side is
\begin{equation}
\int_{T_\varepsilon}^{\tau} |\overline{{\cal S}}(t)|\,\bar{\beta}(t)\,dt \;-\; (|\mathcal V|-1)\,\varepsilon\,(\tau-T_\varepsilon)
\end{equation}
Since $|\overline{{\cal S}}(t)|\ge 1$, if $\limsup_{t\to\infty}\bar{\beta}(t)>0$ the integral would diverge to $+\infty$ (for $\varepsilon$ arbitrarily small), a contradiction. Hence $\lim_{t\to\infty}\bar{\beta}(t)=0$. Combined with $\lim_{t\to\infty}\underline{\beta}(t)=0$, we have $\max_{v_i\in\mathcal V}|\beta_i(t)|\to 0$, which implies $\dot{s}_i(t)\to 0$ for all $v_i\in\mathcal V$.

Let $s^\infty$ be any limit point of $s(t)$. Passing to the limit in the ABMC update gives, for each $v_i\in\mathcal V$,
\[
0=\lim_{t\to\infty}\beta_i(t)=
\begin{cases}
0, & v_i\in\mathcal V_1,\\[2pt]
-\,s_i^\infty+\min_{v_j\in {\cal P}^{\mathrm{cand}}_i}\{s_j^\infty+a_{ij}\}, & v_i\in\mathcal V_2
\end{cases}
\]
so $s^\infty$ satisfies Eq.~\eqref{eq:ABMC_equil}. By uniqueness of the equilibrium, $s^\infty=s^*$ and therefore $s(t)\to s^*$ globally. This establishes global asymptotic stability of the equilibrium. \hfill $\blacksquare$

\subsection*{Proof of Theorem 2}
\label{Proof_T2}

In the classical shortest‐path problem, the goal is to determine a path that minimizes the cumulative (nonnegative) edge cost from any node \(v_i\) to a designated source, with the corresponding optimal cost \(s_i\) satisfying the recursive relation $s_i \;=\; \min_{v_j \in \mathcal{P}^{\mathrm{cand}}_i} \{\, s_j + c_{ij} \,\}, \qquad s_{\text{source}} = 0$, where \(c_{ij}\ge 0\) is the cost of \((v_i, v_j)\). This is Bellman’s equation~\citep{bellman1958routing}.

On the fixed interval \([t_k,t_{k+1})\), the sets \(\mathcal P^{\mathrm{cand}}_i\) and biases \(a_{ij}\) are constant. In our formulation, the state \(s_i\) represents the minimum cumulative bias along a directed backup path from \(v_i\) to the leader \(v_1\), and the classical edge cost \(c_{ij}\) is replaced by the bias \(a_{ij}\ge\gamma>0\). Hence the recursion becomes
\begin{equation}
s_i \;=\; \min_{v_j \in \mathcal{P}^{\mathrm{cand}}_i} \{\, s_j + a_{ij} \,\}, \qquad s_1 = 0
\end{equation}
This is exactly Bellman’s equation with edge costs \(a_{ij}\).

Define the (time‐invariant on \([t_k,t_{k+1})\)) Bellman operator
\begin{equation}
    (\mathfrak{T}s)_i \;\coloneqq\; \min_{v_j\in\mathcal P^{\mathrm{cand}}_i}\{\, s_j + a_{ij} \,\}, \quad i\in\mathcal V_2,
\qquad (\mathfrak{T}s)_1 \coloneqq 0
\end{equation}
The operator \(\mathfrak{T}\) is monotone and nonexpansive in the max norm: 
\(\|\mathfrak{T}s - \mathfrak{T}\tilde s\|_\infty \le \|s-\tilde s\|_\infty\).
Moreover, by construction of \(\mathcal P^{\mathrm{cand}}_i\) with \(\mathcal H_j<\mathcal H_i\), every admissible hop strictly decreases hierarchy, so directed backup paths are acyclic and finite, and \(v_1\) is reachable from every \(v_i\in\mathcal V_2\). On a finite acyclic graph with nonnegative edge costs and boundary \(s_1=0\), the system \(s=\mathfrak{T}s\) has a unique solution equal to the vector of minimum cumulative costs to \(v_1\) (standard dynamic programming/shortest-path argument, e.g., induction over increasing hierarchy).

By Theorem~1, the ABMC state converges to a unique equilibrium \(s^*\) satisfying the same fixed-point equation \(s^*=\mathfrak{T}s^*\). Hence, for each \(v_i\in\mathcal V\),
\begin{equation}
s_i^* \;=\; \min_{\pi:\,v_i\rightsquigarrow v_1}\ \sum_{(v_p\to v_q)\in \pi} a_{pq}
\end{equation}
i.e., \(s_i^*\) equals the minimum cumulative bias along any directed backup path from \(v_i\) to the leader \(v_1\) (and \(s_1^*=0\)). \hfill\(\blacksquare\)

\section{The Reconfiguration Setup of HHC}
\label{APP:Reconfig}

For the reconfiguration process, we assume that the robot dynamics adhere to a single integrator model and utilize an illustrative distance-based formation control law as provided in \cite{oh2015survey}.

\begin{figure}[ht]
\centering
\includegraphics[width=0.7\textwidth]{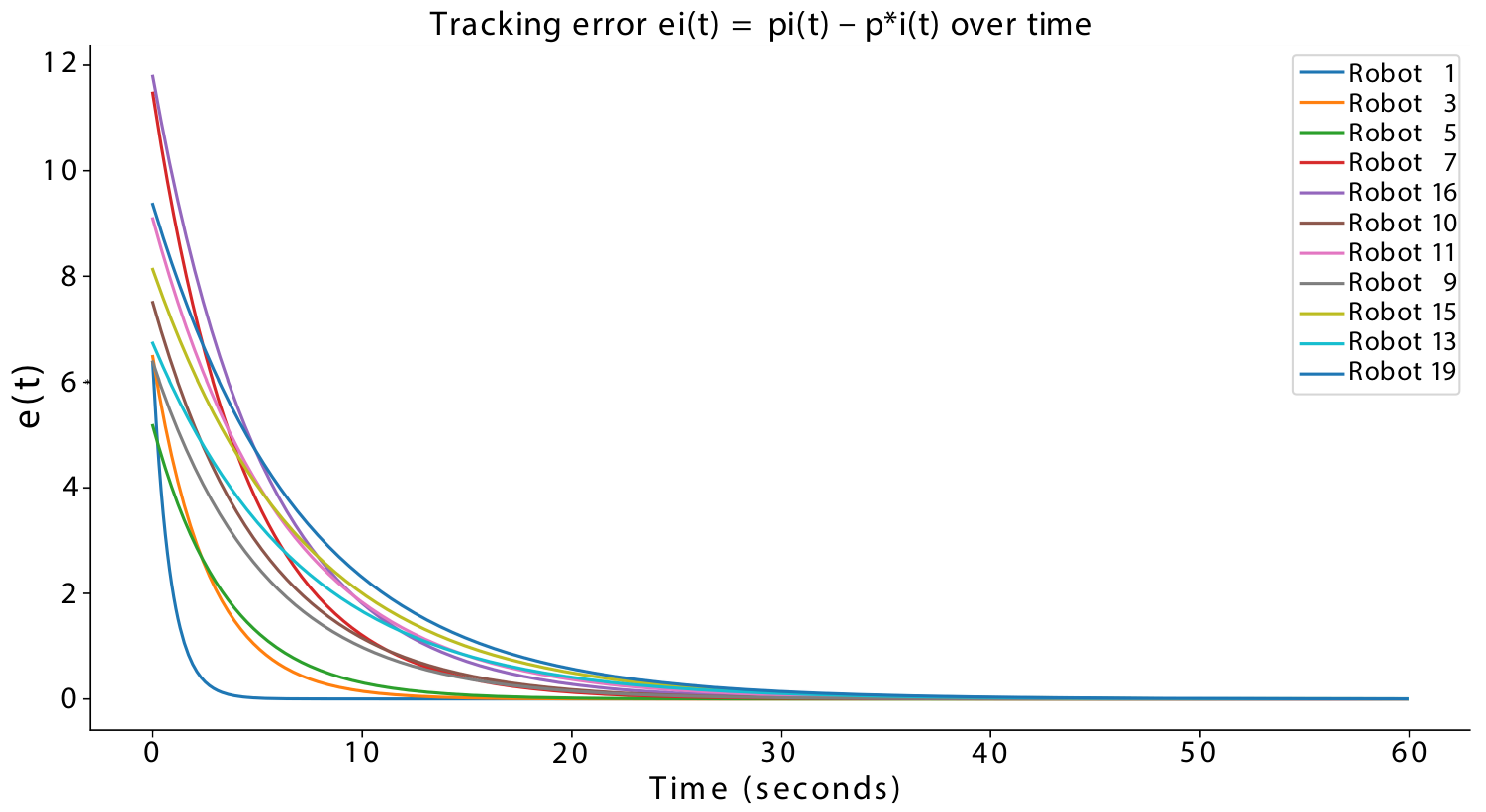}
\caption{Formation tracking error for each robot following formation reconfiguration.}
\label{fig:ABMC6}
\end{figure}

The reconfiguration follows a process where each surviving robot assesses its proximity to the desired formation configuration and assumes the role of the nearest removed robot, as dictated by the reconfiguration algorithm.  For instance, robot 3 takes over the role of robot 2, robot 5 replaces robot 4, etc. The formation tracking error for each robot post-reconfiguration is depicted in Fig.~\ref{fig:ABMC6} for the reconfiguration example given in Fig.~\ref{fig:ABMC4}. The errors are presented as a function of time, showing the asymptotic convergence to zero for all robots, with varying rates depending on their hierarchical level. The tracking errors for robots at lower hierarchical levels diminish rapidly, while those at higher levels converge more slowly. This hierarchical convergence behavior ensures that the overall formation stability is prioritized, with the root and critical robots in the formation tree achieving stability first.

\section{Fault and Noise Modeling}
\label{APP:Data}

% Reset the equation counter and change the numbering format
\setcounter{equation}{0}
\renewcommand{\theequation}{\Alph{section}.\arabic{equation}}

At each time \(t\), robot \(v_i\) in \(\mathbb{R}^2\) receives relative position data from: (1) a path of direct parent robot \(v_j \in \mathcal{P}_i\) that may exhibit offset faults, and  (2) a fault-free backup path \(\mathcal{B}_i[b] \in \bm{\mathcal{B}}_i\) from a leader robot.

We model offset faults as:
\begin{equation}
\label{eq:offsets}
f_{x_{ij}}(t) = o_{x_{ij}} \, s(t)\, w(t), 
\quad
f_{y_{ij}}(t) = o_{y_{ij}} \, s(t)\, w(t)
\end{equation}
where \(o_{x_{ij}}, ~o_{y_{ij}}\) are constant offsets, \(s\) is a Bernoulli random variable with probability \(P_f\), and \(w\) is a function indicating fault duration.

The potentially faulty data 
\(\mathbf{\tilde{q}}_{ij}(t) = [\,x_{ij}(t) + f_{x_{ij}}(t),\, y_{ij}(t) + f_{y_{ij}}(t)\,]^\mathsf{T}\)
follows a Gaussian distribution:
\begin{equation}
\label{eq:tildep}
\begin{aligned}
\mathbf{\tilde{q}}_{ij}(t) &\sim \mathfrak{N}\Biggl(
\begin{bmatrix}
x_{ij}(t) + P_f \times o_{x_{ij}}\\
y_{ij}(t) + P_f \times o_{y_{ij}}
\end{bmatrix},
\\
&\hspace{0.5cm}
\begin{bmatrix}
P_f \times (1-P_f) \times o_{x_{ij}}^2 & 0 \\
0 & P_f \times (1-P_f) \times o_{y_{ij}}^2
\end{bmatrix}
\Biggr)
\end{aligned}
\end{equation}
where $\mathfrak{N}$ denotes the Gaussian (normal) distribution.

Conversely, the fault-free backup data is:
\begin{equation}
\label{eq:pb}
\mathbf{q}_{ij}(t)
= \begin{bmatrix}
x_{ij}(t)\\
y_{ij}(t)
\end{bmatrix}
\sim
\mathfrak{N}\!\Bigl(
\begin{bmatrix}
x_{ij}(t)\\
y_{ij}(t)
\end{bmatrix},
\mathbf{0}\Bigr)
\end{equation}

Each relative‐position packet transmitted from robot \(v_j\) to \(v_i\) traverses a binary‐symmetric channel with bit‐error probability \(P_e\).  Equivalently, we approximate the effect of bit errors as an additional additive noise term,
\begin{equation}
\label{eq:Pe_noise}
\mathbf{q}_{ij}^{\mathrm{rx}}(t)
= \mathbf{q}_{ij}(t) + \boldsymbol\eta_{ij}(t), 
\quad
\boldsymbol\eta_{ij}(t)\sim\mathfrak{N}\bigl(\mathbf{0},\,\sigma_e^2(P_e)\mathbf{I}\bigr)
\end{equation}
where the noise variance \(\sigma_e^2\) is chosen proportional to \(P_e\) to reflect the expected distortion caused by random bit flips.

\section{Calculation of Likelihood Ratio (LR) Statistics}
\label{APP:LR}

\setcounter{equation}{0}
\renewcommand{\theequation}{\Alph{section}.\arabic{equation}}

To calculate the LR, let $\mathbf{\tilde{Q}}_{ij} \triangleq \{\mathbf{\tilde{q}}_{ij}[k] : k = 1, 2, \dots, N\}$ denote the \(N\) samples from the potentially faulty parent robot, and  $\mathbf{Q}_{ij}[b] \triangleq \{\mathbf{q}_{ij}[k] : k = 1, 2, \dots, N\}$ the \(N\) samples from the fault-free backup path $\mathcal{B}_i[b] \in \bm{\mathcal{B}}_i$.

We first compute sample means. For the parent robot path:
\begin{equation}
\label{eq:sample_means_l1}
\boldsymbol{\mu}_{\mathbf{\tilde{Q}}_{ij}}
= \frac{1}{N} \sum_{k=1}^{N} \mathbf{\tilde{q}}_{ij}[k]
= \bigl[\mu_{\tilde{\mathbf{X}}_{ij}},\, \mu_{\tilde{\mathbf{Y}}_{ij}}\bigr]   
\end{equation}
where
\[
\mu_{\tilde{\mathbf{X}}_{ij}} = \frac{1}{N} \sum_{k=1}^{N} \tilde{x}_{ij}[k], 
\quad
\mu_{\tilde{\mathbf{Y}}_{ij}} = \frac{1}{N} \sum_{k=1}^{N} \tilde{y}_{ij}[k]
\]
Similarly, for the fault-free backup path:
\begin{equation}
\label{eq:sample_means_l2}  
\boldsymbol{\mu}_{\mathbf{Q}_{ij}[b]}
= \frac{1}{N} \sum_{k=1}^{N} \mathbf{q}_{ij}[k]
= \bigl[\mu_{\mathbf{X}_{ij}[b]},\, \mu_{\mathbf{Y}_{ij}[b]}\bigr]
\end{equation}
where
\[
\mu_{\mathbf{X}_{ij}[b]} = \frac{1}{N} \sum_{k=1}^{N} x_{ij}[k], 
\quad
\mu_{\mathbf{Y}_{ij}[b]} = \frac{1}{N} \sum_{k=1}^{N} y_{ij}[k]
\]

Next, the sample covariance for the parent robot path is
\begin{equation}
\label{eq:covariance_matrice_parent}   
\boldsymbol{\Sigma}_{\mathbf{\tilde{Q}}_{ij}}
= \frac{1}{N-1} \sum_{k=1}^{N}
\bigl(\mathbf{\tilde{q}}_{ij}[k] - \boldsymbol{\mu}_{\mathbf{\tilde{Q}}_{ij}}\bigr)
\bigl(\mathbf{\tilde{q}}_{ij}[k] - \boldsymbol{\mu}_{\mathbf{\tilde{Q}}_{ij}}\bigr)^T
\end{equation}
For the fault-free backup path, we assume zero covariance:
\begin{equation}
\label{eq:covariance_matrice_backup}  
\boldsymbol{\Sigma}_{\mathbf{Q}_{ij}[b]}
= \begin{bmatrix}
0 & 0 \\
0 & 0
\end{bmatrix}
\end{equation}

The LR compares the probability of the observed parent robot data \(\mathbf{\tilde{Q}}_{ij}\) under the fault hypothesis \(H_1\) versus the no-fault hypothesis \(H_0\). Under \(H_1\), the Gaussian PDF parameters (mean and covariance) are estimated directly from the parent data \(\mathbf{\tilde{Q}}_{ij}\). Under \(H_0\), the Gaussian PDF is defined by the parameters estimated from the fault-free backup path data \(\mathbf{Q}_{ij}[b]\). In this formulation, the backup data do not appear explicitly in the likelihood product; rather, they determine the nominal model parameters.
\begin{equation}
LR_{ij}[b] = \frac{\Pr\{\mathbf{\tilde{Q}}_{ij} \mid H_1\}}{\Pr\{\mathbf{\tilde{Q}}_{ij} \mid H_0\}}
\end{equation}
where
\[
\Pr\{\mathbf{\tilde{Q}}_{ij} \mid H_1\} = \prod_{k=1}^{N} f\bigl(\mathbf{\tilde{q}}_{ij}[k] ;\, \boldsymbol{\mu}_{\mathbf{\tilde{Q}}_{ij}},\, \boldsymbol{\Sigma}_{\mathbf{\tilde{Q}}_{ij}}\bigr)
\]
and
\[
\Pr\{\mathbf{\tilde{Q}}_{ij} \mid H_0\} = \prod_{k=1}^{N} f\bigl(\mathbf{\tilde{q}}_{ij}[k] ;\, \boldsymbol{\mu}_{\mathbf{Q}_{ij}[b]},\, \boldsymbol{\Sigma}_{\mathbf{Q}_{ij}[b]}\bigr)
\]
Here, \(f(\cdot; \boldsymbol{\mu}, \boldsymbol{\Sigma})\) denotes the Gaussian PDF with mean \(\boldsymbol{\mu}\) and covariance \(\boldsymbol{\Sigma}\).

Substituting these PDFs, the closed-form LR is
\begin{align}
LR_{ij}[b] &= \left(\frac{|\boldsymbol{\Sigma}_{\mathbf{Q}_{ij}[b]}|}{|\boldsymbol{\Sigma}_{\mathbf{\tilde{Q}}_{ij}}|}\right)^{N/2} \exp\Bigg(-\frac{1}{2} \sum_{k=1}^{N} \Big[ \nonumber \\
& \quad \quad (\mathbf{\tilde{q}}_{ij}[k] - \boldsymbol{\mu}_{\mathbf{\tilde{Q}}_{ij}})^T 
\boldsymbol{\Sigma}^{-1}_{\mathbf{\tilde{Q}}_{ij}} (\mathbf{\tilde{q}}_{ij}[k] - \boldsymbol{\mu}_{\mathbf{\tilde{Q}}_{ij}}) \nonumber \\
& \quad \quad - (\mathbf{q}_{ij}[k] - \boldsymbol{\mu}_{\mathbf{Q}_{ij}[b]})^T 
\boldsymbol{\Sigma}^{-1}_{\mathbf{Q}_{ij}[b]} (\mathbf{q}_{ij}[k] - \boldsymbol{\mu}_{\mathbf{Q}_{ij}[b]}) \Big] \Bigg)
\label{eq:LRforRoute_b}
\end{align}

The use of a zero covariance matrix for the fault-free backup path may lead to numerical issues. In particular, a zero covariance results in a zero determinant and a non-invertible matrix, which renders the Gaussian likelihood calculations undefined. To avoid these singularities and account for minimal inherent noise, it is common practice to introduce a small positive constant  to the diagonal entries of the covariance matrix.

The log-likelihood ratio (LLR) is:
\begin{align}
\text{LLR}_{ij}[b] &= \frac{N}{2} \ln \left(\frac{|\boldsymbol{\Sigma}_{\mathbf{Q}_{ij}[b]}|}{|\boldsymbol{\Sigma}_{\mathbf{\tilde{Q}}_{ij}}|}\right) -\frac{1}{2} \sum_{k=1}^{N} \Big[ \nonumber \\
& \quad \quad (\mathbf{\tilde{q}}_{ij}[k] - \boldsymbol{\mu}_{\mathbf{\tilde{Q}}_{ij}})^T 
\boldsymbol{\Sigma}^{-1}_{\mathbf{\tilde{Q}}_{ij}} (\mathbf{\tilde{q}}_{ij}[k] - \boldsymbol{\mu}_{\mathbf{\tilde{Q}}_{ij}}) \nonumber \\
& \quad \quad - (\mathbf{q}_{ij}[k] - \boldsymbol{\mu}_{\mathbf{Q}_{ij}[b]})^T 
\boldsymbol{\Sigma}^{-1}_{\mathbf{Q}_{ij}[b]} (\mathbf{q}_{ij}[k] - \boldsymbol{\mu}_{\mathbf{Q}_{ij}[b]}) \Big]
\label{eq:LLR}
\end{align}

Because \(\ln\) of small determinant values can cause further numerical issues, one can impose a lower bound on \(\lvert \boldsymbol{\Sigma}_{\mathbf{Q}_{ij}[b]} \rvert\) to avoid excessive underflow in practical implementations.




\end{appendices}


\bibliography{reference}% common bib file

@article{mo2021global,
  title={Global asymptotic stability of a general biased min-consensus protocol},
  author={Mo, Yuanqiu and Yu, Lanlin and Yu, Changbin},
  journal={IET Control Theory \& Applications},
  volume={15},
  number={8},
  pages={1148--1156},
  year={2021},
  publisher={Wiley Online Library}
}

@article{lin2009stability,
  title={Stability and stabilizability of switched linear systems: a survey of recent results},
  author={Lin, Hai and Antsaklis, Panos J},
  journal={IEEE Transactions on Automatic control},
  volume={54},
  number={2},
  pages={308--322},
  year={2009},
  publisher={IEEE}
}

@article{jam2025sweep,
  author={Aryo Jamshidpey and Wahby, Mostafa and Allwright, Michael and Zhu, Weixu and Dorigo, Marco and Heinrich, Mary Katherine},
  title={Centralization vs. decentralization in multi-robot sweep
coverage with ground robots and {UAV}s},
  journal={Artificial Life and Robotics},
  year={2025},
  publisher={Springer},
  note={DOI:~{10.1007/s10015-025-01049-7}}
}

@article{hespanha2003hysteresis,
  title={Hysteresis-based switching algorithms for supervisory control of uncertain systems},
  author={Hespanha, Joao P and Liberzon, Daniel and Morse, A Stephen},
  journal={Automatica},
  volume={39},
  number={2},
  pages={263--272},
  year={2003},
  publisher={Elsevier}
}

@article{varadharajan2020swarm,
  title={Swarm relays: Distributed self-healing ground-and-air connectivity chains},
  author={Varadharajan, Vivek Shankar and St-Onge, David and Adams, Bram and Beltrame, Giovanni},
  journal={IEEE Robotics and Automation Letters},
  volume={5},
  number={4},
  pages={5347--5354},
  year={2020},
  publisher={IEEE}
}

@book{clarke1990optimization,
  title={Optimization and nonsmooth analysis},
  author={Clarke, Frank H},
  year={1990},
  publisher={SIAM},
  address={Philadelphia}
}

@book{bullo2018lectures,
  title={Lectures on network systems},
  author={Bullo, Francesco and Cort{\'e}s, Jorge and D{\"o}rfler, Florian and Mart{\'\i}nez, Sonia},
  volume={1},
  year={2018},
  publisher={CreateSpace},
  address={North Charleston, SC}
}

@article{thrun2002probabilistic,
  title={Probabilistic robotics},
  author={Thrun, Sebastian},
  journal={Communications of the ACM},
  volume={45},
  number={3},
  pages={52--57},
  year={2002},
  publisher={ACM New York, NY, USA}
}

@article{oh2015survey,
  title={A survey of multi-agent formation control},
  author={Oh, Kwang-Kyo and Park, Myoung-Chul and Ahn, Hyo-Sung},
  journal={Automatica},
  volume={53},
  pages={424--440},
  year={2015},
  publisher={Elsevier}
}

@article{muhammed2017analysis,
  title={An analysis of fault detection strategies in wireless sensor networks},
  author={Muhammed, Thaha and Shaikh, Riaz Ahmed},
  journal={Journal of Network and Computer Applications},
  volume={78},
  pages={267--287},
  year={2017},
  publisher={Elsevier}
}

@article{khaldi2017monitoring,
  title={Monitoring a robot swarm using a data-driven fault detection approach},
  author={Khaldi, Belkacem and Harrou, Fouzi and Cherif, Foudil and Sun, Ying},
  journal={Robotics and Autonomous Systems},
  volume={97},
  pages={193--203},
  year={2017},
  publisher={Elsevier}
}

@article{zhang2021intermittent,
  title={Intermittent fault detection for delayed stochastic systems over sensor networks},
  author={Zhang, Sen and Sheng, Li and Gao, Ming},
  journal={Journal of the Franklin Institute},
  volume={358},
  number={13},
  pages={6878--6896},
  year={2021},
  publisher={Elsevier}
}

@article{yaramasu2015aircraft,
  title={Aircraft electric system intermittent arc fault detection and location},
  author={Yaramasu, Anil and Cao, Yinni and Liu, Guangjun and Wu, Bin},
  journal={IEEE Transactions on aerospace and electronic systems},
  volume={51},
  number={1},
  pages={40--51},
  year={2015},
  publisher={IEEE}
}

@article{sheng2021intermittent,
  title={Intermittent fault detection for linear discrete-time stochastic multi-agent systems},
  author={Sheng, Li and Zhang, Sen and Gao, Ming},
  journal={Applied Mathematics and Computation},
  volume={410},
  pages={126480},
  year={2021},
  publisher={Elsevier}
}

@phdthesis{millard2016exogenous,
  title={Exogenous fault detection in swarm robotic systems},
  author={Millard, Alan},
  year={2016},
  school={University of York}
}

@phdthesis{oladiran2019fault,
  title={Fault Recovery in Swarm Robotics Systems using Learning Algorithms},
  author={Oladiran, Oyinlola Ojuolape},
  year={2019},
  school={University of York}
}

@article{yan2018detection,
  title={Detection, isolation and diagnosability analysis of intermittent faults in stochastic systems},
  author={Yan, Rongyi and He, Xiao and Wang, Zidong and Zhou, DH},
  journal={International Journal of Control},
  volume={91},
  number={2},
  pages={480--494},
  year={2018},
  publisher={Taylor \& Francis}
}

@article{zhang2020intermittent,
  title={Intermittent fault detection for discrete-time linear stochastic systems with time delay},
  author={Zhang, Sen and Sheng, Li and Gao, Ming and Zhou, Donghua},
  journal={IET Control Theory \& Applications},
  volume={14},
  number={3},
  pages={511--518},
  year={2020},
  publisher={Wiley Online Library}
}

@article{zhang2018event,
  title={Event-triggered filtering and intermittent fault detection for time-varying systems with stochastic parameter uncertainty and sensor saturation},
  author={Zhang, Junfeng and Christofides, Panagiotis D and He, Xiao and Wu, Zhe and Zhang, Zhihao and Zhou, Donghua},
  journal={International journal of robust and nonlinear control},
  volume={28},
  number={16},
  pages={4666--4680},
  year={2018},
  publisher={Wiley Online Library}
}

@article{abdelwahed2008practical,
  title={Practical implementation of diagnosis systems using timed failure propagation graph models},
  author={Abdelwahed, Sherif and Karsai, Gabor and Mahadevan, Nagabhushan and Ofsthun, Stanley C},
  journal={IEEE Transactions on instrumentation and measurement},
  volume={58},
  number={2},
  pages={240--247},
  year={2008},
  publisher={IEEE}
}

@article{olfati2004consensus,
  title={Consensus problems in networks of agents with switching topology and time-delays},
  author={Olfati-Saber, Reza and Murray, Richard M},
  journal={IEEE Transactions on automatic control},
  volume={49},
  number={9},
  pages={1520--1533},
  year={2004},
  publisher={IEEE}
}

@article{ren2005consensus,
  title={Consensus seeking in multiagent systems under dynamically changing interaction topologies},
  author={Ren, Wei and Beard, Randal W},
  journal={IEEE Transactions on automatic control},
  volume={50},
  number={5},
  pages={655--661},
  year={2005},
  publisher={IEEE}
}

@article{carvalho2017diagnosability,
  title={Diagnosability of intermittent sensor faults in discrete event systems},
  author={Carvalho, Lilian K and Moreira, Marcos V and Basilio, Jo{\~a}o Carlos},
  journal={Automatica},
  volume={79},
  pages={315--325},
  year={2017},
  publisher={Elsevier}
}

@article{syed2016novel,
  title={A novel intermittent fault detection algorithm and health monitoring for electronic interconnections},
  author={Syed, Wakil Ahmad and Perinpanayagam, Suresh and Samie, Mohammad and Jennions, Ian},
  journal={IEEE Transactions on Components, Packaging and Manufacturing Technology},
  volume={6},
  number={3},
  pages={400--406},
  year={2016},
  publisher={IEEE}
}

@article{khadidos2015exogenous,
  title={Exogenous fault detection and recovery for swarm robotics},
  author={Khadidos, Adil and Crowder, Richard M and Chappell, Paul H},
  journal={IFAC-PapersOnLine},
  volume={48},
  number={3},
  pages={2405--2410},
  year={2015},
  publisher={Elsevier}
}

@inproceedings{strobel2018managing,
	address = {Richland, SC},
	author = {V. Strobel and E. {Castell{\'o} Ferrer} and M. Dorigo},
	booktitle = {Proceedings of the 17th International Conference on Autonomous Agents and Multiagent Systems},
	doi = {http://dl.acm.org/citation.cfm?id=3237383.3237464},
	editor = {M. Dastani and G. Sukthankar and E. Andr{\'e} and S. Koenig},
	pages = {541--549},
	publisher = {International Foundation for Autonomous Agents and Multiagent Systems},
	series = {AAMAS '18},
	title = {Managing Byzantine Robots via Blockchain Technology in a Swarm Robotics Collective Decision Making Scenario},
	year = {2018}}

@article{o2023predictive,
  title={Predictive Fault Tolerance for Autonomous Robot Swarms},
  author={O'Keeffe, James and Millard, Alan Gregory},
  journal={arXiv preprint arXiv:2309.09309},
  year={2023}
}

@article{tarapore2017generic,
  title={Generic, scalable and decentralized fault detection for robot swarms},
  author={Tarapore, Danesh and Christensen, Anders Lyhne and Timmis, Jon},
  journal={PloS one},
  volume={12},
  number={8},
  pages={e0182058},
  year={2017},
  publisher={Public Library of Science San Francisco, CA USA}
}

@article{dorigo2020reflections,
  title={Reflections on the future of swarm robotics},
  author={Dorigo, Marco and Theraulaz, Guy and Trianni, Vito},
  journal={Science Robotics},
  volume={5},
  number={49},
  pages={eabe4385},
  year={2020},
  publisher={American Association for the Advancement of Science}
}

@article{jamshidpey2023reducing,
  title={Reducing Uncertainty in Collective Perception Using Self-Organizing Hierarchy},
  author={Jamshidpey, Aryo and Dorigo, Marco and Heinrich, Mary Katherine},
  journal={Intelligent Computing},
  volume={2},
  pages={0044},
  year={2023},
  publisher={AAAS}
}

@article{dorigo2021swarm,
  title={Swarm robotics: Past, present, and future [point of view]},
  author={Dorigo, Marco and Theraulaz, Guy and Trianni, Vito},
  journal={Proceedings of the IEEE},
  volume={109},
  number={7},
  pages={1152--1165},
  year={2021},
  publisher={IEEE}
}

@article{niu2021distributed,
  title={Distributed intermittent fault detection for linear stochastic systems over sensor network},
  author={Niu, Yichun and Sheng, Li and Gao, Ming and Zhou, Donghua},
  journal={IEEE Transactions on Cybernetics},
  volume={52},
  number={9},
  pages={9208--9218},
  year={2021},
  publisher={IEEE}
}

@article{breitfelder2000ieee,
  title={{IEEE 100}: the authoritative dictionary of IEEE standards terms},
  author={Breitfelder, Kim and Messina, Don},
  journal={Standards Information Network IEEE Press. v879},
  pages={1249},
  year={2000}
}

@misc{IEC192-04-04,
  author = {{International Electrotechnical Commission}},
  title = {192-04-04. Permanent Fault. International Electrotechnical Vocabulary online database},
  howpublished = {\url{http://std.iec.ch/iec60050}},
  note = {Accessed on: Mar. 2024},
  year = 2024
}

@inproceedings{valentini2015efficient,
  title={Efficient decision-making in a self-organizing robot swarm: On the speed versus accuracy trade-off},
  author={Valentini, Gabriele and Hamann, Heiko and Dorigo, Marco},
  booktitle={Proceedings of the 2015 International Conference on Autonomous Agents and Multiagent Systems},
  pages={1305--1314},
  year={2015}
}

@article{tarapore2019fault,
  title={Fault detection in a swarm of physical robots based on behavioral outlier detection},
  author={Tarapore, Danesh and Timmis, Jon and Christensen, Anders Lyhne},
  journal={IEEE Transactions on Robotics},
  volume={35},
  number={6},
  pages={1516--1522},
  year={2019},
  publisher={IEEE}
}

@inproceedings{millard2014towards,
  title={Towards exogenous fault detection in swarm robotic systems},
  author={Millard, Alan G and Timmis, Jon and Winfield, Alan FT},
  booktitle={Towards Autonomous Robotic Systems: 14th Annual Conference, TAROS 2013, Oxford, UK, August 28--30, 2013, Revised Selected Papers 14},
  pages={429--430},
  year={2014},
  organization={Springer}
}

@article{zhou2019review,
  title={Review on diagnosis techniques for intermittent faults in dynamic systems},
  author={Zhou, Donghua and Zhao, Yinghong and Wang, Zidong and He, Xiao and Gao, Ming},
  journal={IEEE Transactions on Industrial Electronics},
  volume={67},
  number={3},
  pages={2337--2347},
  year={2019},
  publisher={IEEE}
}

@article{winfield2006safety,
  title={Safety in numbers: fault-tolerance in robot swarms},
  author={Winfield, Alan FT and Nembrini, Julien},
  journal={International Journal of Modelling, Identification and Control},
  volume={1},
  number={1},
  pages={30--37},
  year={2006},
  publisher={Inderscience Publishers}
}

@incollection{heinrich2022swarm,
  title={Swarm robotics},
  author={Heinrich, Mary Katherine and Wahby, Mostafa and Dorigo, Marco and Hamann, Heiko},
  booktitle={Cognitive robotics},
  editor={Cangelosi, Angelo and Asada, Minoru},
  year={2022},
  pages = {77--98},
  publisher={MIT Press},
  address={Cambridge, MA}
}

@article{olfati2007consensus,
  title={Consensus and cooperation in networked multi-agent systems},
  author={Olfati-Saber, Reza and Fax, J Alex and Murray, Richard M},
  journal={Proceedings of the IEEE},
  volume={95},
  number={1},
  pages={215--233},
  year={2007},
  publisher={IEEE}
}

@inproceedings{ren2005survey,
  title={A survey of consensus problems in multi-agent coordination},
  author={Ren, Wei and Beard, Randal W and Atkins, Ella M},
  booktitle={Proceedings of the 2005, American Control Conference, 2005.},
  pages={1859--1864},
  year={2005},
  organization={IEEE}
}

@article{rubenstein2014programmable, 
  title={Programmable self-assembly in a thousand-robot swarm},
  author={Rubenstein, Michael and Cornejo, Alejandro and Nagpal, Radhika},
  journal={Science},
  volume={345},
  number={6198},
  pages={795--799},
  year={2014},
  publisher={American Association for the Advancement of Science}
}

@article{mathews2017mergeable,
  title={Mergeable nervous systems for robots},
  author={Mathews, Nithin and Christensen, Anders Lyhne and O’Grady, Rehan and Mondada, Francesco and Dorigo, Marco},
  journal={Nat. Commun.},
  volume={8},
  number={1},
  pages={1--7},
  year={2017},
  publisher={Nature Publishing Group}
}

@inproceedings{zhu2020formation,
  title={Formation Control of {UAVs} and Mobile Robots using Self-organized Communication Topologies},
  author={Zhu, Weixu and Allwright, Michael and Heinrich, Mary Katherine and O\u{g}uz, Sinan and Christensen, Anders Lyhne and Dorigo, Marco},
  booktitle={Proc. 12th Ants Conf.},
  address = {Berlin, Germany},
  pages={306--314},
  publisher = {Springer},
  year={2020},
}

@article{cortes2008distributed,
  title={Distributed algorithms for reaching consensus on general functions},
  author={Cort{\'e}s, Jorge},
  journal={Automatica},
  volume={44},
  number={3},
  pages={726--737},
  year={2008},
  publisher={Elsevier}
}

@inproceedings{de2006decentralized,
  title={Decentralized control of connectivity for multi-agent systems},
  author={De Gennaro, Maria Carmela and Jadbabaie, Ali},
  booktitle={Proceedings of the 45th IEEE Conference on Decision and Control},
  pages={3628--3633},
  year={2006},
  organization={IEEE}
}

@article{moreau2005stability,
  title={Stability of multiagent systems with time-dependent communication links},
  author={Moreau, Luc},
  journal={IEEE Transactions on automatic control},
  volume={50},
  number={2},
  pages={169--182},
  year={2005},
  publisher={IEEE}
}

@article{zavlanos2011graph,
  title={Graph-theoretic connectivity control of mobile robot networks},
  author={Zavlanos, Michael M and Egerstedt, Magnus B and Pappas, George J},
  journal={Proceedings of the IEEE},
  volume={99},
  number={9},
  pages={1525--1540},
  year={2011},
  publisher={IEEE}
}

@article{kirst2016dynamic,
  title={Dynamic information routing in complex networks},
  author={Kirst, Christoph and Timme, Marc and Battaglia, Demian},
  journal={Nature communications},
  volume={7},
  number={1},
  pages={11061},
  year={2016},
  publisher={Nature Publishing Group UK London}
}

@inproceedings{pini2011task,
  title={Task partitioning in swarms of robots: reducing performance losses due to interference at shared resources},
  author={Pini, Giovanni and Brutschy, Arne and Birattari, Mauro and Dorigo, Marco},
  booktitle={Informatics in Control Automation and Robotics: Revised and Selected Papers from the International Conference on Informatics in Control Automation and Robotics 2009},
  pages={217--228},
  year={2011},
  organization={Springer}
}

@book{hamann2018swarm,
  title={Swarm robotics: A formal approach},
  author={Hamann, Heiko},
  year={2018},
  publisher={Springer},
  address={Cham}
}

@inproceedings{jamshidpey2020multi,
  title={Multi-robot Coverage Using Self-organized Networks for Central Coordination},
  author={Jamshidpey, Aryo and Zhu, Weixu and Wahby, Mostafa and Allwright, Michael and Heinrich, Mary Katherine and Dorigo, Marco},
  booktitle={Proc. 12th Ants Conf.},
  address = {Berlin, Germany},
  pages={306--314},
  publisher = {Springer},
  year={2020},
}

@article{zhang2023self,
  title={Self-reconfigurable hierarchical frameworks for formation control of robot swarms},
  author={Zhang, Yuwei and O{\u{g}}uz, Sinan and Wang, Shaoping and Garone, Emanuele and Wang, Xingjian and Dorigo, Marco and Heinrich, Mary Katherine},
  journal={IEEE Transactions on Cybernetics},
  year={2023},
  publisher={IEEE}
}

@article{zhang2017distributed,
  title={Distributed biased min-consensus with applications to shortest path planning},
  author={Zhang, Yinyan and Li, Shuai},
  journal={IEEE Transactions on Automatic Control},
  volume={62},
  number={10},
  pages={5429--5436},
  year={2017},
  publisher={IEEE}
}

@article{bellman1958routing,
  title={On a routing problem},
  author={Bellman, Richard},
  journal={Quarterly of applied mathematics},
  volume={16},
  number={1},
  pages={87--90},
  year={1958}
}

@book{royden1988real,
  title={Real analysis},
  author={Royden, Halsey Lawrence and Fitzpatrick, Patrick},
  volume={32},
  year={1988},
  publisher={Macmillan},
  address={New York}
}

@book{isidori2013nonlinear,
  title={Nonlinear control systems II},
  author={Isidori, Alberto},
  year={2013},
  publisher={Springer},
  address={Cham}
}

@article{christensen2009fireflies,
  title={From fireflies to fault-tolerant swarms of robots},
  author={Christensen, Anders Lyhne and OGrady, Rehan and Dorigo, Marco},
  journal={IEEE Transactions on Evolutionary Computation},
  volume={13},
  number={4},
  pages={754--766},
  year={2009},
  publisher={IEEE}
}

@inproceedings{bjerknes2013fault,
  title={On fault tolerance and scalability of swarm robotic systems},
  author={Bjerknes, Jan Dyre and Winfield, Alan FT},
  booktitle={Distributed Autonomous Robotic Systems: The 10th International Symposium},
  pages={431--444},
  year={2013},
  organization={Springer}
}

@article{ghedini2017toward,
  title={Toward fault-tolerant multi-robot networks},
  author={Ghedini, Cinara and Ribeiro, Carlos and Sabattini, Lorenzo},
  journal={Networks},
  volume={70},
  number={4},
  pages={388--400},
  year={2017},
  publisher={Wiley Online Library}
}

@article{zhu2024self,
author = {Weixu Zhu  and Sinan Oğuz  and Mary Katherine Heinrich  and Michael Allwright  and Mostafa Wahby  and Anders Lyhne Christensen  and Emanuele Garone  and Marco Dorigo },
title = {Self-organizing nervous systems for robot swarms},
journal = {Science Robotics},
volume = {9},
number = {96},
pages = {eadl5161},
year = {2024},
doi = {10.1126/scirobotics.adl5161},
URL = {https://www.science.org/doi/abs/10.1126/scirobotics.adl5161},
eprint = {https://www.science.org/doi/pdf/10.1126/scirobotics.adl5161}
}

@article{DorBirBra2014:sch-sr,
	author = {M. Dorigo and M. Birattari and M. Brambilla},
	doi = {http://dx.doi.org/10.4249/scholarpedia.1463},
	journal = {Scholarpedia},
	number = {1},
	optlabel = {IJ.101},
	pages = {1463},
	title = {Swarm Robotics},
	volume = {9},
	year = {2014}}

\end{document}